\documentclass[final,3p,times,authoryear]{elsarticle}

\usepackage{hyperref}
\usepackage{url}
\usepackage{amsmath}
\usepackage{amssymb}
\usepackage{mathtools}
\usepackage{upgreek}
\usepackage{soul}
\usepackage{graphicx}
\usepackage{subcaption}
\usepackage{lineno}
\usepackage{subcaption}
\usepackage{tikz}
\usepackage{graphicx}
\newcommand{\Czero}[1]{C_0([0,T];{\mathbb{R}^#1})}

\newcommand{\sigmagamma}[1][]{\sigma_{\it{\Gamma}}\left(#1\right)}
\newcommand{\sigmapi}[1][]{\sigma_{\it{\Pi}}\left(#1\right)}

\newcommand{\Phigamma}[1]{\Phi_{\it{\Gamma}}\left[#1\right]}

\newcommand{\PhiGamma}{{\Phi}_{{\it{\Gamma}}}}
\newcommand{\PiPhiGamma}{{\it{\Pi}}\circ{\Phi}_{{\it{\Gamma}}}}
\newcommand{\hatPiPhiGamma}{{\hat{\it{\Pi}}}\circ{\Phi}_{{{\hat{\it{\Gamma}}}}}}

\newcommand{\PiPhiGammaone}{{\it{\Pi}}_1\circ{\Phi}_{{\it{\Gamma}}_1}}
\newcommand{\PiPhiGammatwo}{{\it{\Pi}}_2\circ{\Phi}_{{\it{\Gamma}}_2}}

\newcommand{\norminf}[1]{{\left\lVert #1 \right\rVert}_{L^{\infty}}}

\newcommand{\normtwo}[1]{{\left\lVert #1 \right\rVert}_{L^{2}}}

\newcommand{\wGammain}{w_{{\it{\Gamma},{\mathrm{in}}}}}
\newcommand{\wGamma}{w_{\it{\Gamma}}}
\newcommand{\wGammaout}{w_{{\it{\Gamma},{\mathrm{out}}}}}

\newcommand{\wPiin}{w_{{\it{\Pi},{\mathrm{in}}}}}
\newcommand{\wPi}{w_{\it{\Pi}}}
\newcommand{\wPiout}{w_{{\it{\Pi},{\mathrm{out}}}}}
\newcommand{\hPi}{h_{\it{\Pi}}}

\newcommand{\itGamma}{{\it{\Gamma}}}
\newcommand{\itPi}{{\it{\Pi}}}

\newcommand{\FGamma}{\mathcal{F}_{\it{\Gamma}}}
\newcommand{\FPi}{\mathcal{F}_{\it{\Pi}}}

\newcommand{\FPhiGamma}{{\mathcal{F}}_{{\Phi}_{\it{\Gamma}}}}
\newcommand{\FPiPhiGamma}{ {\mathcal{F}}_{{\it{\Pi}}\circ{\Phi}_{{\it{\Gamma}}}} }

\newcommand{\BW}{B_{\mathbf{W}}}
\newcommand{\Bb}{B_{\mathbf{b}}}
\newcommand{\BtildeW}{B_{\tilde{\mathbf{W}}}}
\newcommand{\Btildeb}{B_{\tilde{\mathbf{b}}}}

\newcommand{\BPi}{B_{\it{\Pi}}}

\newcommand{\loss}{{\ell}}
\newcommand{\hatloss}{{\hat{{\ell}}}_{{\mathbf{U}}_N, {\mathbf{V}}_N}}

\newcommand{\Bloss}{B_{\mathrm{loss}}}

\DeclarePairedDelimiter{\floor}{\lfloor}{\rfloor}
\DeclarePairedDelimiter{\ceil}{\lceil}{\rceil}

\definecolor{mybackground}{RGB}{255, 255, 255}
\pagecolor{mybackground}

\begin{document}

\begin{frontmatter}



\title{Upper Generalization Bounds for Neural Oscillators} 

\author[label1]{Zifeng Huang\corref{cor1}}
\ead{zifeng.huang@irz.uni-hannover.de}

\author[label2]{Konstantin M. Zuev}
\ead{kostia@caltech.edu}

\author[label3]{Yong Xia}
\ead{ceyxia@polyu.edu.hk}

\author[label1,label4,label5]{Michael Beer}
\ead{beer@irz.uni-hannover.de}

\cortext[cor1]{Corresponding author.}

\affiliation[label1]{organization={Institute for Risk and Reliability, Leibniz University Hannover},
            addressline={Callinstraße 34},
            city={Hannover},
            postcode={30167},
            country={Germany}}

\affiliation[label2]{organization={Department of Computing and Mathematical Sciences, California Institute of Technology},
            city={Pasadena},
            state={California},
            country={United States}}
            
\affiliation[label3]{organization={Joint Research Centre for Marine Infrastructure, Department of Civil and Environmental Engineering, The Hong Kong Polytechnic University},
            addressline={Kowloon},
            city={Hong Kong},
            country={China}}

\affiliation[label4]{organization={Department of Civil and Environmental Engineering, University of Liverpool},
            city={Liverpool},
            postcode={L69 3GH},
            country={United Kingdom}}

\affiliation[label5]{organization={International Joint Research Center for Resilient Infrastructure \& International Joint Research Center for Engineering Reliability and Stochastic Mechanics, Tongji University},
            city={Shanghai},
            postcode={200092},
            country={China}}

\begin{abstract}
Neural oscillators that originate from second-order ordinary differential equations (ODEs) have shown competitive performance in learning mappings between dynamic loads and responses of complex nonlinear structural systems. Despite this empirical success, theoretically quantifying the generalization capacities of their neural network architectures remains undeveloped. In this study, the neural oscillator consisting of a second-order ODE followed by a multilayer perceptron (MLP) is considered. Its upper probably approximately correct (PAC) generalization bound for approximating causal and uniformly continuous operators between continuous temporal function spaces and that for approximating the uniformly asymptotically incrementally stable second-order dynamical systems are derived by leveraging the Rademacher complexity framework. {\color{black} These bounds are further extended to the squared Wasserstein-1 distances between the probability measures of quantities of interest calculated from target causal operators and the corresponding learned neural oscillators.} The theoretical results show that the estimation errors grow polynomially with respect to both {\color{black} MLP sizes} and the time length, thereby avoiding the curse of parametric complexity. Furthermore, the derived error bounds demonstrate that constraining the Lipschitz constants of the MLPs via loss function regularization can improve the generalization ability of the neural oscillator. {\color{black} Numerical studies} considering a Bouc-Wen nonlinear system under stochastic seismic excitation validates the theoretically predicted power laws of the estimation errors with respect to the sample size and time length, and confirms the effectiveness of constraining MLPs' matrix and vector norms in enhancing the performance of the neural oscillator  under limited training data. 
\end{abstract}






\begin{keyword}
Neural oscillator \sep
PAC generalization bound \sep
Causal continuous operator \sep
Rademacher complexity \sep 
{\color{black} Wasserstein-1 distance}
\sep
Lipschitz regularization


\end{keyword}

\end{frontmatter}



\section{Introduction}
\label{sec1}
Accurately modeling mappings among long sequences or continuous temporal functions is a significant challenge in machine learning, yet it is critical for a wide range of applications in science and engineering. To address this challenge, numerous neural network architectures have been proposed for sequential and functional learning tasks, which {\color{black} include} recurrent neural networks (RNNs) \citep{rumelhart1986learning} and their enhanced variants \citep{hochreiter1997long,cho2014learning}, attention-based architectures \citep{vaswani2017attention}, state-space (SS) models \citep{gu2020hippo,gu2021efficiently,gu2022parameterization}, and neural oscillators \citep{rusch2020coupled,rusch2021unicornn,lanthaler2023neural,rusch2024oscillatory,zifeng2025}. Originating from ordinary differential equations (ODEs), SS models and neural oscillators combine the fast inference capability of RNNs with the competitive long-range sequence learning performance of attention-based architectures \citep{gu2023mamba,rusch2024oscillatory}. Neural oscillators have been theoretically demonstrated to alleviate vanishing and exploding gradient problems \citep{rusch2020coupled,rusch2021unicornn} and to ensure stable dynamics in learning long-range dependencies \citep{rusch2024oscillatory}. {\color{black} Regarding their applications, neural oscillators have already been utilized in heart-rate prediction from wearable-device data \citep{rusch2024oscillatory}, latent-dynamics modeling of soft-robot control from raw pixel images \citep{stolzle2024input}, generalization enhancement of physics-informed machine learning for time-dependent nonlinear partial differential equations (PDEs) \citep{kapoor2024neural}, and direct learning of nonlinear derivative terms governing complex ODE systems \citep{zifeng2025}.}

In addition to the empirical success of SS models and neural oscillators in practical applications, advancing their theoretical understanding, especially in quantifying the approximation and generalization bounds of their network architectures, is also a significant issue. Several studies have theoretically investigated the approximation and generalization bounds of SS models. \citet{JMLR:v23:21-0368} proved an approximation bound for linear RNNs in universally approximating linear, causal, continuous, regular, and time-homogeneous operators between continuous temporal function spaces. \citet{orvieto2024universality} derived an approximation bound for the SS models consisting of a {\color{black} discrete-time} linear RNN followed by an multilayer perceptron (MLP) in modeling the mappings between finite-length sequences. \citet{liu2024generalization} derived a probably approximately correct (PAC) upper generalization bound for a linear SS model using the Rademacher complexity. In this bound, the Rademacher complexity constant incorporates the model parameters and the expectation and variance of sub-Gaussian input function processes. Under the constraint that the spectral abscissa of the state matrix is smaller than zero, \citet{honarpisheh2025generalization} established a length-independent norm-based generalization bound for the selective SS model. Both studies only considered the mappings from input sequences or continuous temporal functions to vector outputs at a final time instant. By leveraging a layer-by-layer Rademacher contraction approach, \citet{racz2024length} derived a PAC upper generalization bound for a deep SS model consisting of a {\color{black} discrete-time} linear RNN followed by an MLP. This bound is still limited to the mappings from input sequences to scalar outputs, and the estimation error (difference between the generalization error and the approximation error) in this generalization bound grows exponentially with the depth of the employed MLP. In addition to these results, the approximation and generalization properties of the traditional RNNs for approximating continuous mappings between continuous temporal function spaces \citep{hanson2020universal,hanson2021learning,fermanian2021framing,hanson2024rademacher} and those for approximating fading memory causal operators of {\color{black} discrete-time} sequences \citep{grigoryeva2018echo,grigoryeva2018universal,gonon2020risk,gonon2023approximation, yasumoto2025universality} have been studied. In contrast to SS models, theoretical analyses of the approximation and generalization properties of neural oscillators remain limited. For a neural oscillator architecture consisting of a second-order ODE followed by an MLP, \citet{huang2025upper} established its upper approximation bounds for approximating causal continuous operators between continuous temporal function spaces and for approximating stable second-order dynamical systems. The generalization bounds of neural oscillators, however, remain unexplored.


In this study, the neural oscillator consisting of a second-order ODE followed by an MLP is considered. Its upper PAC generalization bound for approximating causal and uniformly continuous operators between continuous temporal function spaces and that for approximating uniformly asymptotically incrementally stable second-order dynamical systems are established. In Section \ref{sec2}, the neural oscillator architecture is introduced, some necessary assumptions are outlined, and the established theoretical approximation bounds of the neural oscillator in \citet{huang2025upper} are briefly reviewed. The newly derived theoretical generalization bounds of the neural oscillator are also informally provided in Section \ref{sec2} to facilitate understanding. In Section \ref{sec3}, the formal statements of two theorems that establish the generalization bounds for the neural oscillator, accompanied by the supporting lemmas essential for their proofs, are provided. {\color{black} In addition, by leveraging the Kantorovich-Rubinstein duality, the established generalization bounds are extended to the squared Wasserstein-1 distances between the probability measures of quantities of interest calculated from the temporal output functions of the target causal operators and the corresponding learned neural oscillators.} These generalization bounds reveal that constraining the Lipschitz constants of the employed MLPs in the loss function can improve the generalization capability of the neural oscillator. In Section \ref{sec4}, {\color{black} numerical studies based on a Bouc-Wen nonlinear system under stochastic seismic excitation validate} the power laws of the established estimation errors with respect to the sample size and time length, and confirms that constraining the MLPs' matrix and vector norms can improve performance of the neural oscillator under limited training data.

{\bf Notations:} For an \textit{n}-dimensional vector $\mathbf{a} = [a_1, a_2,..., a_n]^\top$, its $\it{L^r}$-norm is $|\mathbf{a}|_r = [\sum_{i=1}^n |a_i|^r]^{1/r}$ for $1 \leq r < \infty$ and $|\mathbf{a}|_r = \max_{1 \leq i \leq n}|a_i|$ for $r = +\infty$. $|\mathbf{a}|$ is short for $|\mathbf{a}|_1$. For a matrix $\mathbf{M} = [m_{ij}]_{p \times q}$, its $L^{r,s}$-norm is $\lvert\mathbf{M}\rvert_{r,s} = \big|\left[\lvert\mathbf{m}_{1}\rvert_{r}, \lvert\mathbf{m}_{2}\rvert_{r},\dots,\lvert\mathbf{m}_{q}\rvert_{r}\right]^\top\big|_{s}$, where $1 \leq r,s \leq +\infty$ and $\mathbf{m}_{j}$ is the $j^\text{th}$ column vector of $\mathbf{M}$. The $L^r$-norm of $\mathbf{M}$ is $|\mathbf{M}|_r = \sup_{|\mathbf{a}|_r = 1}|\mathbf{Ma}|_r/|\mathbf{a}|_r$. $\mathbb{Z}$ and $\mathbb{R}$ respectively represent the integer and real number spaces. $\floor{\cdot}$ and $\ceil{\cdot}$ respectively represent the floor and ceiling functions. Re($\cdot$) and Im($\cdot$) respectively represent the real and imaginary parts of a complex number. $\mathbb{R}^{p{\times}q}$ represents the $p \times q$-dimensional real matrix space. $f'(t) = \mathrm{d}f(t)/\mathrm{d}t$ and $f''(t) = \mathrm{d}^{2}f(t)/\mathrm{d}t^{2}$. $C_0([0, T]; \mathbb{R}^p)$ represents a \textit{p}-dimensional real-valued continuous function space defined over $[0, T]$ with respect to {\color{black} the} $L^{\infty}$-norm ${\lVert \cdot \rVert}_{L^{\infty}}$. For $\mathbf{u}(t) = [u_1(t), u_2(t),..., u_p(t)]^\top \in \Czero{p}$, its $L^{\infty}$-norm is ${\lVert\mathbf{u}(t) \rVert}_{L^{\infty}}$ $= \sup_{t\in[0, T]}|\mathbf{u}(t)|$ and $L^2$-norm ${\lVert \cdot \rVert}_{L^2}$ is ${\lVert \mathbf{u}(t) \rVert}_{L^2} = \sqrt{\int_{0}^{T}|\mathbf{u}(t)|_{2}^{2}\mathrm{d}t}$. Big $O$ notation: $f(n) = O\left[g(n)\right]$ if there exist positive constants $c$ and $n_0$ such that $f(n) \leq c \cdot g(n)$ for all $n \geq n_0$.

\section{Neural Oscillator}
\label{sec2}
A neural oscillator architecture mapping $\mathbf{u}(t) = [u_1(t), u_2(t),..., u_p(t)]^{\top} \in \Czero{p}$ to $\mathbf{y}(t) = [y_1(t), y_2(t),..., y_q(t)]^{\top} \in \Czero{q}$ is \citep{huang2025upper}
\begin{equation}
    \left\{
    \begin{aligned}
    &\mathbf{x}''(t) = {\it{\Gamma}}[\mathbf{x}(t), \mathbf{x}'(t), \mathbf{u}(t)] \\
    &\mathbf{y}(t) = {\it{\Pi}}[\mathbf{x}(t), \mathbf{u}(0), t]
    \end{aligned}
     \right.
    \label{eq:2.1}
\end{equation} with $\mathbf{x}(0) = \mathbf{x}'(0) = \mathbf0$, where $\mathbf{x}(t) = [x_1(t), x_2(t),..., x_r(t)]^\top$ is an intermediate \textit{r}-dimensional real-valued state function, ${\it{\Gamma}}(\cdot)$ and ${\it{\Pi}}(\cdot)$ are two MLPs
\begin{equation}
   {\it{\Gamma}}\left[ {{\bf{x}}(t),{\bf{x'}}(t),{\bf{u}}(t)} \right] = {{\bf{W}}_2}{\sigma _{\it{\Gamma}}}\left\{ {{{\bf{W}}_1}{{\left[ {{{\bf{x}}^{\top}}(t),{{{\bf{x'}}}^{\top}}(t),{{\bf{u}}^{\top}}(t)} \right]}^{\top}} + {{\bf{b}}_1}} \right\} + {{\bf{b}}_2},
   \label{eq:2.2}
\end{equation}
\begin{equation}
   {\it{\Pi}}\left[ {{\bf{x}}(t),{\bf{u}}(0),t} \right] = {{\tilde{\bf W}}_{{h_{\it{\Pi}} }}}{\sigma _{\it{\Pi}}}\left( { \ldots {\sigma _{\it{\Pi}} }\left\{ {{{{\tilde{\bf W}}}_1}{{\left[ {{{\bf{x}}^{\top}}(t),{{\bf{u}}^{\top}}(0),t} \right]}^{\top}} + {{{\tilde{\bf b}}}_1}} \right\}} \right) + {{\tilde{\bf b}}_{{h_{\it{\Pi}}}}},
   \label{eq:2.3}
\end{equation}
$\mathbf{W}_i$, \textit{i} = 1 and 2, $\tilde{\mathbf{W}}_j$, \textit{j} = 1, 2,..., $h_{\it{\Pi}}$, are real-valued {\color{black} trainable} weight matrices, $\mathbf{b}_i$ and $\tilde{\mathbf{b}}_j$ are real-valued {\color{black} trainable} bias vectors, $h_{\it{\Pi}}$ is the depth of ${\it{\Pi}}(\cdot)$, and $\sigmagamma[\cdot]$ and $\sigmapi[\cdot]$ are respectively the activation functions of ${\it{\Gamma}}(\cdot)$ and ${\it{\Pi}}(\cdot)$. Unless otherwise specified, these two activation functions are set to the Rectified Linear Unit (ReLU) \citep{nair2010rectified}, defined as $\sigmagamma[x] = \sigmapi[x] = \sigma_{\mathrm{ReLU}}(x)=\mathrm{max}(0,x)$, in this study. The input layer widths of ${\it{\Gamma}}(\cdot)$ and ${\it{\Pi}}(\cdot)$ are denoted as $\wGammain$ and $\wPiin$, respectively. The output layer widths of ${\it{\Gamma}}(\cdot)$ and ${\it{\Pi}}(\cdot)$ are denoted as $\wGammaout$ and $\wPiout$, respectively. The largest hidden layer widths of ${\it{\Gamma}}(\cdot)$ and ${\it{\Pi}}(\cdot)$ are denoted as $\wGamma$ and $\wPi$, respectively. The neural oscillator in Eq.~\eqref{eq:2.1} can be written as $\mathbf{y}(t) = {\it{\Pi}}\circ{\Phi}_{\it{\Gamma}}[\mathbf{u}(\tau)](t) = {\it{\Pi}}\left\{{{\Phi}_{\it{\Gamma}}[{\mathbf{u}(\tau)}](t),\mathbf{u}(0),t}\right\}$, where $\circ$ is the composition operator and $\mathbf{x}(t) = \Phigamma{\mathbf{u}(\tau)}(t)$ represents the mapping governed by the second-order ODE in Eq.~\eqref{eq:2.1}. 

Two theoretical approximation bounds of the neural oscillator in Eq.~\eqref{eq:2.1}, corresponding respectively to the approximation of causal and uniformly continuous operators and that of uniformly asymptotically incrementally stable second-order dynamical systems, have been established in \citet{huang2025upper}. These two approximation bounds, together with {\color{black} their necessary assumptions} and the definition of uniformly asymptotically incrementally stable systems, are summarized below. {\color{black} The detailed relationships between each assumption and the two approximation error bounds are provided in \citet{huang2025upper}. The reader is referred to that work for the complete proofs and intermediate results.}

{\bf Assumption 1:} $C_0([0,T];{\mathbb{R}}^p)$ with $T \geq 1$ is a Banach space equipped with the $L^\infty$-norm. Its metric is defined as ${\lVert{\mathbf{u}_1(t)-\mathbf{u}_2(t)}\rVert}_{L^{\infty}}$ for $\mathbf{u}_1(t)$ and $\mathbf{u}_2(t)$ $\in$ $C_0([0,T];\mathbb{R}^p)$.

{\bf Assumption 2:} \textit{K} is a compact subset of ${C_0([0,T];\mathbb{R}^p)}$ and it is closed. From the Arzelà-Ascoli theorem \citep{beattie2013convergence}, all functions belonging to \textit{K} are uniformly bounded and there exists a constant $B_K$ such that ${\lVert{u_i(t)}\rVert}_{L^\infty} \leq B_K$ for all $\mathbf{u}(t) \in K$ and $1 \leq i \leq p$, where $u_i(t)$ is the $i^\mathrm{th}$ element of $\mathbf{u}(t)$. It is assumed $B_K \geq 1$. In addition, all functions belonging to $K$ are equicontinuous and there exists a continuous modulus of continuity {\color{black} $\phi_K(t) = \sum_{i=1}^{p}\phi_{K,i}(t)$} such that for arbitrary $0 \leq t_1 \leq t_2 \leq T$ and all $\mathbf{u}(t) \in K$, it is satisfied that $|u_i(t_1) - u_i(t_2)|\leq\phi_{K,i}(t_2-t_1)$ {\color{black} for $1 \leq i \leq p$}, thereby {\color{black} $|\mathbf{u}(t_1) - \mathbf{u}(t_2)|\leq \phi_K(t_2-t_1)$}, where $\phi_{K,i}(t): [0,+\infty) \to [0,+\infty)$ is monotonic with $\phi_{K,i}(t) \to 0$ as $t \to 0$.

{\bf Assumption 3:} ${{\Phi}} = [\Phi_1,\Phi_2...., \Phi_q]: C_0([0,T];{\mathbb{R}}^p) \to {\color{black} C_0([0,T];{\mathbb{R}}^q)}$ is a causal and uniformly continuous operator. For $\forall{t} \in [0,T]$, $\Phi[\mathbf{u}(\tau)](t)$ only depends on $\mathbf{u}(\tau)$ over $\tau \in [0,t]$. For $\forall{\varepsilon} > 0$, there exists a distance $d_{\varepsilon} > 0$, such that ${\lVert{\mathbf{u}_1(t)-\mathbf{u}_2(t)}\rVert}_{L^{\infty}} < d_{\varepsilon} \Rightarrow {\lVert{\Phi[\mathbf{u}_1(\tau)](t) - \Phi[\mathbf{u}_2(\tau)](t)}\rVert}_{L^{\infty}} < \varepsilon$ for $\forall \mathbf{u}_1(t), \mathbf{u}_2(t)$ $\in {\color{black} C_0([0,T];{\mathbb{R}}^p)}$.

{\bf Assumption 4:} Since ${\Phi}: C_0([0,T];{\mathbb{R}}^p) \to {\color{black} C_0([0,T];{\mathbb{R}}^q)}$ is a causal and uniformly continuous operator and $K \subseteq {\color{black} C_0([0,T];{\mathbb{R}}^p)} $ is compact, the image set $\Phi(K) \subseteq C_0([0,T];{\mathbb{R}}^q)$ of $K$ by $\Phi$ is also compact. There exists a constant $B_{\Phi(K)}$ such that all functions belonging to $\Phi(K)$ are bounded by $B_{\Phi(K)}$, that is ${\lVert{v_j(t)}\rVert}_{L^\infty} \leq B_{\Phi(K)}$ for all $\mathbf{v}(t) = [v_1(t), v_2(t),..., v_q(t)]^\top \in \Phi(K)$ and $1 \leq j \leq q$.

{\bf Assumption 5:} For each element $\Phi_j$ of the causal and uniformly continuous operator ${{\Phi}} = [\Phi_1,\Phi_2...., \Phi_q]$, $1 \leq j \leq q$, it is assumed that the first-order derivatives of all temporal functions in the image set $\Phi_j\left\{ \Czero{p} \right\}$ by $\Phi_j$ are bounded by a finite constant $D_{\Phi_j}$. 

{\bf Assumption 6:} Since ${\Phi}: C_0([0,T];{\mathbb{R}}^p) \to {\color{black} C_0([0,T];{\mathbb{R}}^q)}$ is a causal and uniformly continuous operator, its $j^{\mathrm{th}}$ element $\Phi_j$ has a monotonic continuous modulus of continuity $\phi_{\Phi_j}(\cdot): \left[0,+\infty\right) \to \left[0,+\infty\right)$ with $\phi_{\Phi_j}\left({\color{black} \norminf{\Delta\mathbf{u}}}\right) \to 0$ as ${\color{black} \norminf{\Delta\mathbf{u}}} \to 0$, such that $\norminf{\Phi_j[\mathbf{u}_1(\tau)](t) - \Phi_j[\mathbf{u}_2(\tau)](t)}$ $\leq \phi_{\Phi_j}\left[\norminf{\mathbf{u}_1(t) - \mathbf{u}_2(t)}\right]$ for $\forall\mathbf{u}_1(t), \mathbf{u}_2(t)$ $\in \Czero{p}$. $\phi_{\Phi}(\cdot)$ $=\sum_{j=1}^{q}{\phi_{\Phi_j}(\cdot)}: \left[0,+\infty\right) \to \left[0,+\infty\right)$ with $\phi_{\Phi}\left({\color{black} \norminf{\Delta\mathbf{u}}}\right) \to 0$ as ${\color{black} \norminf{\Delta\mathbf{u}}} \to 0$ is a monotonic modulus of continuity of $\Phi$, such that $\norminf{\Phi[\mathbf{u}_1(\tau)](t) - \Phi[\mathbf{u}_2(\tau)](t)} \leq \sum_{j=1}^{q}\norminf{\Phi_j[\mathbf{u}_1(\tau)](t) - \Phi_j[\mathbf{u}_2(\tau)](t)} \leq \phi_{\Phi}\left[\norminf{\mathbf{u}_1(t) - \mathbf{u}_2(t)}\right]$ for $\forall\mathbf{u}_1(t), \mathbf{u}_2(t)$ $\in \Czero{p}$.

{\bf Definition 1 \citep{hanson2020universal}:} Given a second-order ODE ${\hat{\bf x}}''(t) = {\color{black} \mathbf{g}}\left[ {{\hat{\bf x}}(t),{\hat{\bf x}}'(t),{\bf{u}}(t)} \right]$ with its state function ${\mathbf{z}}(t) = \left[\hat{{\mathbf{x}}}^{\top}(t),\hat{{\mathbf{x}}}'^{\top}(t)\right]^{\top}$, where $\hat{{\mathbf{x}}}(t) = [\hat{x}_{1}(t),\hat{x}_{2}(t),\dots,\hat{x}_{r}(t)]^{\top}$, it is uniformly asymptotically incrementally stable for the inputs in a subset $K \subset \Czero{p}$ on a domain $\Omega_{\mathbf{z}} \subset \mathbb{R}^{2r}$ if there exists a set of non-negative stability functions $\beta_k({\color{black} z},t)$ over ${\color{black} z} \geq 0$ and $t \geq 0$, $k = 1, 2,..., 2r$, where $\beta_k(0,t) = 0$,$\lim\limits_{t\to +\infty}\beta_k({\color{black} z},t) = 0$, $\beta_k({\color{black} z},t)$ is continuous and strictly increasing with respect to ${\color{black} z}$ and is continuous and strictly decreasing with respect to \textit{t}, such that $\left| {{z_k}(t,{{\bf{z}}_\tau }) - {z_k}(t,{{{\hat{\bf z}}}_\tau })} \right| \le {\beta _k}\left( {{{\left| {{{\bf{z}}_\tau } - {{{\hat{\bf z}}}_\tau }} \right|}_2},t - \tau } \right)$ holds for all $\mathbf{u}(t) \in K$, all ${\mathbf{z}}_\tau$ and ${\hat{\bf z}}_\tau$ $\in \Omega_{\mathbf{z}}$, all $t \in [0,T]$, and all $0 \leq \tau \leq t$, where $z_k\left(t, {\mathbf{z}}_\tau\right)$ is the $k^{\text{th}}$ element of $\mathbf{z}\left(t, {\mathbf{z}}_\tau\right)$ and $\mathbf{z}\left(t, {\mathbf{z}}_\tau\right)$ represents the solution $\mathbf{z}(t)$ at \textit{t} driven by $\mathbf{u}(t)$ with an initial condition ${\mathbf{z}}_{\tau}$ at an initial time instant $\tau$. The corresponding stability bound $B_\beta$ is calculated as
\begin{equation}
    {B_\beta } = \int_0^{ + \infty } {{{\left. {\frac{\partial }{{\partial {\color{black} z}}}\sum\limits_{k = 1}^{2r} {{\beta _k}\left({\color{black} z},\tau \right)} } \right|}_{{\color{black} z} = {0^+}}}{\rm{d}}\tau}.
    \label{eq:2.4}
\end{equation}

{\bf Lemma 1 (Theorem 1 of \citealt{huang2025upper}):} {\it Given a compact subset $K \subset \Czero{p}$ and a causal and uniformly continuous operator $\Phi: \Czero{p}$ $\to \Czero{q}$ satisfying Assumptions 1 to 6 with {\color{black} ${\phi}_K(t) = pL_Kt$} and $L_K > 0$, for every integer $M_{\it{\Gamma}}$ {\color{black} implicitly} larger than an independent deterministic threshold and every positive integer $H_{\it{\Pi}}$, there exist two MLPs $\it{\Gamma}(\cdot)$ and $\it{\Pi}(\cdot)$ employing $\sigma_{\mathrm{ReLU}}(\cdot)$, such that for $\forall \mathbf{u}(t) \in K$, the corresponding solution $\mathbf{y}(t) = [y_1(t), y_2(t),...,y_q(t)]^{\top}$ from Eq.~\eqref{eq:2.1} subject to $\mathbf{u}(t)$ with the initial conditions $\mathbf{x}(0) = {\mathbf{x}}'(0) = \mathbf{0}$ satisfies
\begin{equation}
    \left| {\Phi \left[ {{\bf{u}}(\tau )} \right](t) - {\bf{y}}(t)} \right| \le {\varepsilon _{\bf{y}}}
    \label{eq:2.5}
\end{equation}
for all $t \in [0,T]$, where
{\color{black} \begin{equation}
    {\varepsilon _{\bf{y}}} = {\phi _\Phi}\left\{O\left[ \frac{p{L_K}T\left(\ln{M_\itGamma}\right)^{2}}{M_{\it{\Gamma}}}\right]\right\} + O\left\{\frac{qp^{3}M_{\it{\Gamma}}^5}{H_{\it{\Pi}}^{{1 \mathord{\left/
    {\vphantom {1 {\left[ {p\left( {{M_{\it{\Gamma}} } + 1} \right) + 1} \right]}}} \right.
    \kern-\nulldelimiterspace} {\left[ {p\left( {{M_{\it{\Gamma}} } + 1} \right) + 1} \right]}}}}\right\}
    \label{eq:2.6}
\end{equation}}
and ${\phi_{\Phi}}(\cdot)$ is the modulus of continuity of $\Phi$ in Assumption 6. $\it{\Gamma}(\cdot)$ has one hidden layer and the widths of its input, hidden, and output layers are $\wGammain = p\left(2M_{\itGamma}+1\right)$, $\wGamma = 2pM_{\itGamma}$, and $\wGammaout = pM_{\itGamma}$, respectively. For ${\it{\Pi}}(\cdot)$, the widths of its input, hidden, and output layers are $\wPiin = p\left(M_{\itGamma}+1\right)+1$, {\color{black} $\wPi = q\left[p\left(M_{\itGamma}+1\right)+4\right]$}, and $\wPiout = q$, respectively, and its depth is $\hPi = H_{\it{\Pi}} + 1$.} 

{\bf Lemma 2 (Theorem 2 of \citealt{huang2025upper}):} {\it Let K be a compact subset of $\Czero{p}$, a second-order dynamical system
\begin{equation}
    \left\{
    \begin{aligned}
        &{\hat{\bf x}}''(t) = {\color{black} \mathbf{g}}\left[ {{\hat{\bf x}}(t),{\hat{\bf x}}'(t),{\bf{u}}(t)} \right]\\
        &{\hat{\bf y}}(t) = {\color{black} \mathbf{h}}\left[ {{\hat{\bf x}}(t)} \right]
    \end{aligned} 
    \right.,
    \label{eq:2.7}
\end{equation}
with ${\hat{\mathbf{x}}}(0) = {{\hat{\mathbf{x}}}}'(0) = \mathbf{0}$, is uniformly asymptotically incrementally stable for all $\mathbf{u}(t) \in K$ on a domain $\left[-B_{\mathbf{x}}, B_{\mathbf{x}}\right]^{r} \times \left[-B_{{\mathbf{x}}'}, B_{{\mathbf{x}}'}\right]^{r} \times \left[-B_K, B_K\right]^{p} \subset \mathbb{R}^{2r+p}$, where ${\hat{\mathbf{x}}}(t) = \left[{\hat{x}}_1(t),{\hat{x}}_2(t),...,{\hat{x}}_r(t)\right]^{\top}$, ${\hat{\mathbf{y}}}(t) = \left[{\hat{y}}_1(t),{\hat{y}}_2(t),...,{\hat{y}}_q(t)\right]^{\top}$, $B_{{\mathbf{x}}} = \alpha B_{\beta{g}} + B_{\hat{\mathbf{x}}}$, $B_{{\mathbf{x}
}'} = \alpha B_{\beta{g}} + B_{\hat{\mathbf{x}}'}$, $\alpha$ is a positive coefficient, $B_{\hat{\mathbf{x}}}$ and $B_{\hat{\mathbf{x}}'}$ are the bounds of ${\hat{\mathbf{x}}}(t)$ and ${\hat{\mathbf{x}}}'(t)$, respectively, $B_K$ is the bound of all $\mathbf{u}(t) \in K$ in Assumption 2, and {\color{black} $B_{\beta_{\mathbf{g}}}$} is the stability bound of the system calculated by Eq.~\eqref{eq:2.4}. {\color{black} $\mathbf{g}(\cdot) = \left[g_1(\cdot), g_2(\cdot),\dots,g_r(\cdot)\right]^{\top}$ satisfies that for every compact domain $\Omega \subset $ ${\mathbb{R}}^{2r+p}$, there exists a Barron function extension $\tilde{\mathbf{g}}_{\Omega}(\cdot) = \mathbf{g}(\cdot)$ on $\Omega$ and the spectral Barron norms of all extensions are uniformly bounded. $\mathbf{h}(\cdot) = \left[h_1(\cdot), h_2(\cdot),\dots,h_q(\cdot)\right]^{\top}$ satisfies that for every compact domain $\Omega \subset $ ${\mathbb{R}}^{r}$, there exists a Barron function extension $\tilde{\mathbf{h}}_{\Omega}(\cdot) = \mathbf{h}(\cdot)$ on $\Omega$ and the spectral Barron norms of all extensions are uniformly bounded.} Then, for arbitrary positive errors $\varepsilon_1 \leq \alpha/r$ and $\varepsilon_2$, there exist two one-hidden-layer MLPs ${\it{\Gamma}}(\cdot)$ and ${\it{\Pi}}(\cdot)$ employing $\sigma_{\mathrm{ReLU}}(\cdot)$, such that for $\forall\mathbf{u}(t) \in K$, the corresponding solutions $\mathbf{y}(t)$ to Eq.~\eqref{eq:2.1} and ${\hat{\mathbf{y}}}(t)$ to Eq.~\eqref{eq:2.7} satisfy
\begin{equation}
    \left|{{\bf{y}}(t) - {\hat{\bf y}}(t)} \right| \le {\color{black} L_{\mathbf{h}} B_{\beta_{\mathbf{g}}}}\sqrt r {\varepsilon _1} + q{\varepsilon _2}
    \label{eq:2.8}
\end{equation}
for all $t \in [0,T]$, where ${\color{black} L_{\mathbf{h}}}$ is the Lipschitz constant of {\color{black} $\mathbf{h}(\cdot)$}. For ${\it{\Gamma}}(\cdot)$, the widths of its input, hidden, and output layers are $\wGammain = 2r + p$, $\wGamma = 8r\ceil{C_{\itGamma} \varepsilon_{1}^{-2}}$, and $\wGammaout = r$, respectively. For ${\it{\Pi}}(\cdot)$, the widths of its input, hidden, and output layers are $\wPiin = r + p + 1$, $\wPi = 8q\ceil{C_{\itPi} \varepsilon_{2}^{-2}}$, and $\wPiout = q$, respectively.
$C_{\itGamma}$ is a constant only depending on {\color{black} $\mathbf{g}(\cdot)$}, $\sigma_{\mathrm{ReLU}}(\cdot)$, $r$, $p$, $B_{{\mathbf{x}}}$, $B_{{\mathbf{x}}'}$, and $B_K$. $C_{\itPi}$ is a constant only depending on {\color{black} $\mathbf{h}(\cdot)$}, $\sigma_{\mathrm{ReLU}}(\cdot)$, $r$, and $B_K$.}

{\bf Remark 1:} Lemma 1 proves that the neural oscillator in Eq.~\eqref{eq:2.1} can function as a universal approximator of causal and uniformly continuous operators by first encoding an input function $\mathbf{u}(t)$ into the intermediate function $\mathbf{x}(t)$ through the second-order ODE in Eq.~\eqref{eq:2.1}, and then mapping $\mathbf{x}(t)$, together with the initial value $\mathbf{u}(0)$ of $\mathbf{u}(t)$ and time \textit{t}, to the output function $\mathbf{y}(t)$. This capability is similar to the universal approximation property of neural operators \citep{kovachki2023neural}. Moreover, for the second-order dynamical systems in science and engineering \citep{strogatz2024nonlinear}, Lemma 2 demonstrates that the neural oscillator can directly approximate these systems using its ODE in Eq.~\eqref{eq:2.1} with an efficient neural network size.

In the following content of this section, the established theoretical generalization bounds of the neural oscillator in Eq.~\eqref{eq:2.1} are presented in an informal manner. The formal statements of the derived theoretical results are presented in Theorems 1 and 2 in Section \ref{sec3}, together with {\color{black} several additional necessary assumptions and} supporting lemmas required for their proofs.

{\bf Informal Statement of Theorem 1:} {\it {\color{black} Under the conditions in Lemma 1 and some additional assumptions}, for the target causal and uniformly continuous operator $\Phi$, with the specified sizes of $\itGamma(\cdot)$ and $\itPi(\cdot)$ in Lemma 1, the {\color{black} mean-square} generalization error $\loss\left(\hatPiPhiGamma\right)$ of a neural oscillator $\hatPiPhiGamma$ {\color{black} learned from $N$ independent and identically distributed (i.i.d.) sample pairs} is bounded by
\begin{equation}
    \ell\left( {\hat \itPi  \circ {\Phi _{\hat \itGamma }}} \right) \le T\varepsilon _{\bf{y}}^2 + \frac{C_1}{{\sqrt N }}\left[C_2+\sqrt{0.5{{\color{black} \ln} \left( {2{\delta ^{ - 1}}} \right)}}\right]
    \label{eq:2.9}
\end{equation}
with probability at least $1-\delta$ for $\forall\delta \in \left(0,1\right)$, where $\varepsilon_{\mathbf{y}}$ is in Eq.~\eqref{eq:2.6} in Lemma 1, {\color{black} $C_1 = O\left(Tq\right)$, $C_2=O\left\{w_{\mathrm{max}}^{1.5}\left(H_{\itPi}+1\right)\left[\left(w_{\mathrm{max}}+1\right)\sqrt{T} + \sqrt{\left(H_{\itPi}+1\right)\ln{w_{\mathrm{max}}}} + \sqrt{H_{\itPi}+1} + \sqrt{\ln{w_{\mathrm{max}}}} + \sqrt{\ln{T}} \right]  \right\}$, and $w_{\mathrm{max}}=\max\left[p\left(2M_{\itGamma}+1\right),q\left(pM_{\itGamma}+p+4 \right)\right]$.}}

{\bf Informal Statement of Theorem 2:} {\it {\color{black} Under the conditions in Lemma 2 and some additional assumptions}, for the target second-order differential system governed by Eq.~\eqref{eq:2.7}, with the specified sizes of $\itGamma(\cdot)$ and $\itPi(\cdot)$ in Lemma 2, the {\color{black} mean-square} generalization error $\ell\left(\hatPiPhiGamma\right)$ of $\hatPiPhiGamma$ {\color{black} learned by from $N$ i.i.d. sample pairs} is bounded by
\begin{equation}
    \ell \left( {\hat{\it\Pi}  \circ {\Phi _{\hat \itGamma }}} \right) \le 16T\left( {\frac{{{\color{black} L_{\mathbf{h}}^2B_{\beta _{\mathbf{g}}}^2}r^2{C_\itGamma }}}{{{w_\itGamma } - 8r}} + \frac{{q^3{C_\itPi }}}{{{w_\itPi } - 8q}}} \right) + \frac{C_1}{{\sqrt N }}\left[C_2+\sqrt{0.5{{\color{black} \ln} \left( {2{\delta ^{ - 1}}} \right)}}\right] 
    \label{eq:2.10}
\end{equation}
with probability at least $1-\delta$ for $\forall\delta \in \left(0,1\right)$, where $\wGamma > 8\alpha^{-2}r^3C_{\itGamma}+8r$, $\wPi > 8q$, $C_{\itGamma}$ and $C_{\itPi}$ are the constants specified in Lemma 2, {\color{black} $C_1 = O\left(Tq\right)$, $C_2=O\left[w_{\mathrm{max}}^{2.5}\left(\sqrt{T} + \sqrt{\ln{w_{\mathrm{max}}}} + \sqrt{\ln{T}}\right)\right]$, and $w_{\mathrm{max}} = \max\left(2r+p,\wPi,\wGamma\right)$.}}

\section{Upper Generalization Bounds of the Neural Oscillator}
\label{sec3}
In this section, corresponding to the results in Lemmas 1 and 2, the PAC upper generalization bounds for the neural oscillator trained with i.i.d. samples are derived using the method based on the Rademacher complexity and the covering number of a candidate neural oscillator class. This method has been applied to derive the upper generalization bounds of MLPs \citep{bartlett2017spectrally}, RNNs \citep{chen2020generalization}, neural ODEs \citep{bleistein2023generalization,marion2023generalization}, and implicit networks \citep{fung2024generalization}. First, {\color{black} several additional necessary} assumptions and {\color{black} the} definitions on the loss function, covering number, and Rademacher complexity are provided. Then, Lemmas 3 and 4 demonstrate that the Rademacher complexity of the loss function can be bounded by the expected supremum of a sub-Gaussian process. Lemmas 5-10 bound the expected supremum of the sub-Gaussian process using the covering number of a neural oscillator class. Finally, the upper generalization bounds for the neural oscillator in approximating causal and uniformly continuous operators between continuous temporal function spaces and in approximating the uniformly asymptotically incrementally stable second-order dynamical systems are established in Theorems 1 and 2, respectively. {\color{black} In addition, by leveraging the Kantorovich-Rubinstein duality, the generalization bounds in Theorems 1 and 2 are extended to the uncertainty quantification of quantities of interest calculated from the temporal output functions of the target causal operators and the corresponding learned neural oscillators. The resulting generalization error bounds for the squared Wasserstein-1 distance between the probability measures of these quantities of interest calculated from the target causal operators and learned neural oscillators are established in Corollaries 1 and 2, respectively. The proofs of Theorems 1 and 2, Corollaries 1 and 2, and all lemmas in this section are provided in \ref{appA}.} Furthermore, based on the established generalization bounds, an additional regularization term constraining the Lipschitz constants of $\itGamma(\cdot)$ and $\itPi(\cdot)$ is added into the loss function to enhance the generalization ability of the neural oscillator. 

{\bf Assumption 7:} $\mu(\cdot)$ is a {\color{black} Borel} probability measure supported on the compact set $K \subset {\color{black} C_0([0,T];{\mathbb{R}}^p)}$.

{\bf Assumption 8:} Suppose that the sizes of ${\it{\Gamma}}(\cdot)$ in Eq.~\eqref{eq:2.2} and ${\it{\Pi}}(\cdot)$ in Eq.~\eqref{eq:2.3} are specified according to Lemma 1 or Lemma 2, the input layer widths of ${\it{\Gamma}}(\cdot)$ and ${\it{\Pi}}(\cdot)$ are $\wGammain = 2r+p$ and $\wPiin = r+p+1$, respectively. The output layer widths of ${\it{\Gamma}}(\cdot)$ and ${\it{\Pi}}(\cdot)$ are $\wGammaout = r$ and $\wPiout = q$, respectively. The largest {\color{black} hidden-layer} widths of ${\it{\Gamma}}(\cdot)$ and ${\it{\Pi}}(\cdot)$ are $\wGamma$ and $\wPi$, respectively, and the depth $\hPi$ of ${\it{\Pi}}(\cdot)$ satisfies $\hPi \geq 2$. {\color{black} $r$ is the dimension of the intermediate state function $\mathbf{x}(t)$ of the neural oscillator in Eq.~\eqref{eq:2.1} and satisfies $r = pM_{\itGamma}$ in Lemma 1.} 

{\bf Assumption 9:} $\FGamma$ represents a class of MLPs $\itGamma(\cdot)$ whose sizes are specified in Assumption 8. The weight matrices $\mathbf{W}_i$ and vectors $\mathbf{b}_i$, \textit{i} = 1 and 2, of $\itGamma(\cdot) \in \FGamma$ are bounded by $\left|\mathbf{W}_i \right|_{\infty,\infty} \leq B_{\mathbf{W}}$ and $\left|\mathbf{b}_i\right|_{\infty} \leq B_{\mathbf{b}}$. $\FPi$ represents a class of MLPs $\itPi(\cdot)$ whose sizes are specified in Assumption 8. The weight matrices ${\tilde{\mathbf{W}}}_j$ and vectors ${\tilde{\mathbf{b}}}_j$, $j = 1, 2,..., \hPi$, of $\itPi(\cdot) \in \FPi$ are bounded by $\left|{\tilde{\mathbf{W}}}_j \right|_{\infty,\infty} \leq B_{\tilde{{\mathbf{W}}}}$ and $\left|{\tilde{\mathbf{b}}}_j\right|_{\infty} \leq B_{\tilde{{\mathbf{b}}}}$. The outputs of all $\itPi(\cdot) \in \FPi$ are bounded by $\norminf{\itPi_j(\mathbf{x})} \le B_{\itPi}$, $j = 1, 2,..., \wPiout$, where $\itPi_j(\mathbf{x})$ is the $j^{\text{th}}$ output of $\itPi(\mathbf{x})$. The bounds $\BW$ and $\Bb$ depend on the size of $\itGamma(\cdot)$ and the bounds $\BtildeW$, $\Btildeb$, and $\BPi$ depend on the size of $\itPi(\cdot)$, ensuring that the results of Lemma 1 or Lemma 2 can be achieved. 

{\bf Assumption 10:} $\FPhiGamma$ represents a second-order ODE class whose elements $\PhiGamma$ are governed by the ODE in Eq.~\eqref{eq:2.1} with $\itGamma(\cdot) \in \FGamma$. $\FPiPhiGamma$ represents a neural oscillator class whose elements $\PiPhiGamma$ are governed by Eq.~\eqref{eq:2.1} with $\itGamma(\cdot) \in \FGamma$ and $\itPi(\cdot) \in \FPi$.    

Consider the causal continuous operator $\Phi: K \subset \Czero{p} \to \Czero{q}$ in Assumption 3, given \textit{N} i.i.d. samples $\mathbf{U}_N(t) = [\mathbf{u}_1(t),\mathbf{u}_2(t),...,\mathbf{u}_N(t)]$ from the probability measure $\mu(\cdot)$ in Assumption 7, and the corresponding \textit{N} output function samples $\mathbf{V}_N(t) = [\mathbf{v}_1(t),\mathbf{v}_2(t),...,\mathbf{v}_N(t)]$ with $\mathbf{v}_i(t) = \Phi\left[{\mathbf{u}}_i(\tau)\right](t)$, for a neural oscillator $\PiPhiGamma \in \FPiPhiGamma$ in Assumption 10, its {\color{black} mean-square} empirical loss $\hatloss\left(\PiPhiGamma\right)$ is defined as
\begin{equation}
    \hatloss\left(\PiPhiGamma\right) = \frac{1}{N}\sum_{i=1}^{N}\normtwo{{\mathbf{v}}_i(t)-\PiPhiGamma\left[\mathbf{u}_i(\tau)\right](t)}^2.
    \label{eq:3.1}
\end{equation}
An estimated neural oscillator $\hatPiPhiGamma$ is learned by the following optimization problem
\begin{equation}
    \hatPiPhiGamma \in \arg\min\limits_{\PiPhiGamma \in \FPiPhiGamma} \hatloss\left(\PiPhiGamma \right).
    \label{eq:3.2}
\end{equation}
{\color{black} A mean-square} generalization error $\loss\left(\hatPiPhiGamma\right)$ of $\hatPiPhiGamma$ is calculated as
\begin{equation}
    \loss\left(\hatPiPhiGamma\right) = \mathbb{E}_{\mathbf{u}\backsim\mu}\vartheta _{\hatPiPhiGamma}\left[\mathbf{u}(\tau)\right] = \int_{K}\vartheta _{\hatPiPhiGamma}\left[\mathbf{u}(\tau)\right]\text{d}\mu\left[\mathbf{u}(\tau)\right],
    \label{eq:3.3}
\end{equation}
where
\begin{equation}
    {\vartheta _{\hat {\it{\Pi}}  \circ {\Phi _{\hat {\it{\Gamma}} }}}}\left[ {{\bf{u}}(\tau )} \right] = \left\| {\Phi \left[ {{\bf{u}}(\tau )} \right](t) - \hat {\it{\Pi}}  \circ {\Phi _{\hat {\it{\Gamma}} }}\left[ {{\bf{u}}(\tau )} \right](t)} \right\|_{{L^2}}^2.
    \label{eq:3.4}
\end{equation}

{\bf Assumption 11:} Since the outputs of all neural oscillators in $\FPiPhiGamma$ are bounded by $\BPi$ from Assumptions 9 and 10, and $\mathbf{v}(t) = \Phi\left[\mathbf{u}(\tau)\right](t)$ with $\mathbf{u}(t) \in K$ is bounded by $B_{\Phi(K)}$ in Assumption 4, the empirical loss $\hatloss$ and generalization error $\loss$ are bounded by $\hatloss, \loss \leq Tq\Bloss^2$, where $\Bloss = \BPi + B_{\Phi(K)}$.

{\bf Definition 2 (Cover):} Let $\mathcal{M}$ be a space with a pseudo-metric $d_{\mathcal{M}}$, for any real $\varepsilon > 0$, a subset $\mathcal{S} \subset \mathcal{M}$ is an $\varepsilon$-cover of $\mathcal{M}$ with respect to $d_{\mathcal{M}}$ if for $\forall f \in \mathcal{M}$, there exists $\tilde{f} \in \mathcal{S}$, such that $d_{\mathcal{M}}\left(f,\tilde{f}\right) < \varepsilon$.

{\bf Definition 3 (Covering number, Definition 2.2.3 of \citealt{van2013weak}):} Let $\mathcal{M}$ be a space with a pseudo-metric $d_{\mathcal{M}}$, for any real $\varepsilon > 0$, the covering number of $\mathcal{M}$ is defined as $N\left(\mathcal{M}, d_{\mathcal{M}}, \varepsilon\right) = \min\left(\#\mathcal{S}: \mathcal{S} \text{ is an } \varepsilon \text{-cover of } \mathcal{M} \right)$, where $\#\mathcal{S}$ is the cardinality of $\mathcal{S}$.

{\bf Definition 4 (Empirical Rademacher Complexity, Definition 3.1 of \citealt{mohri2018foundations}):} Given \textit{N} i.i.d. samples $\mathbf{U}_N(t) = [\mathbf{u}_1(t),\mathbf{u}_2(t),...,\mathbf{u}_N(t)]$ from $\mu(\cdot)$ in Assumption 7, the empirical Rademacher complexity $\Re_{\mathcal{G}_{\vartheta}}$ of the functional class for $\vartheta_{\PiPhiGamma}\left[\mathbf{u}(\tau)\right]$ in Eq.~\eqref{eq:3.4} with $\PiPhiGamma \in \FPiPhiGamma$ is defined as
\begin{equation}
    \Re_{\mathcal{G}_{\vartheta}} = \mathbb{E}_{\boldsymbol{\upsigma}}\sup\limits_{\PiPhiGamma \in \FPiPhiGamma}\frac{1}{N}\sum_{i=1}^{N}\sigma_i\vartheta_{\PiPhiGamma}\left[\mathbf{u}_i(\tau)\right],
    \label{eq:3.5}
\end{equation}
where $\boldsymbol{\upsigma} = [\sigma_1, \sigma_2,...,\sigma_N]$ is a Rademacher vector of i.i.d. Rademacher variables $\sigma_i$ taking values of 1 or –1 with equal probability and $i = 1, 2,..., N$.

{\bf Lemma 3:} {\it Given \textit{N} i.i.d. samples $\mathbf{U}_N(t) = [\mathbf{u}_1(t),\mathbf{u}_2(t),...,\mathbf{u}_N(t)]$ from $\mu(\cdot)$ in Assumption 7, the empirical Rademacher complexity $\Re_{\mathcal{G}_{\vartheta}}$ in Eq.~\eqref{eq:3.5} is bounded by
\begin{equation}
    {\Re _{{\mathcal{G}_\vartheta }}} \le \frac{{2\sqrt {2T{w_{{\it{\Pi}} ,{\rm{out}}}}} {B_{{\rm{loss}}}}}}{N}{\mathbb{E}_{{\boldsymbol{\tilde \upsigma }}}}\mathop {\sup }\limits_{{\it{\Pi}}  \circ {\Phi _{\it{\Gamma}} } \in {\mathcal{F}_{{\it{\Pi}}  \circ {\Phi _{\it{\Gamma}} }}}} \kappa \left( {{\it{\Pi}}  \circ {\Phi _{\it{\Gamma}} },{\boldsymbol{\tilde \upsigma }}} \right),
    \label{eq:3.6}
\end{equation}
where
\begin{equation}
    \kappa \left( {{\it{\Pi}}  \circ {\Phi _{\it{\Gamma}} },{\boldsymbol{\tilde \upsigma }}} \right) = \sum\limits_{i = 1}^N {\sum\limits_{j = 1}^{{w_{{\it{\Pi}} ,{\rm{out}}}}} {\sum\limits_{k = 1}^{ + \infty } {{{\tilde \sigma }_{ijk}}{\alpha _{jk}}\left\{ {{\it{\Pi}}\circ {\Phi _{\it{\Gamma}} }{{\left[ {{{\bf{u}}_i}(\tau )} \right]_j}}} \right\}} } },
    \label{eq:3.7}
\end{equation}
$\Bloss$ is in Assumption 11, $\alpha_{jk}\left\{\PiPhiGamma\left[\mathbf{u}_i(\tau)\right]_j\right\}$ is the coefficient of $\PiPhiGamma\left[\mathbf{u}_i(\tau)\right]_j(t)$ under a countable set of standard orthogonal basis functions $e_k(t)$ over $[0, T]$, $i = 1, 2,..., N$, $j = 1, 2,..., \wPiout$, $k = 1, 2,...$, and $\boldsymbol{\tilde \upsigma }$ is a third-order tensor with elements ${\tilde{\sigma}}_{ijk}$ being i.i.d. Rademacher variables over discrete indices i, j, and k.}

{\bf Lemma 4:} {\it Under the conditions in Lemma 3, $\kappa \left( {{\it{\Pi}}  \circ {\Phi _{\it{\Gamma}} },{\boldsymbol{\tilde \upsigma }}} \right)$ in Eq.~\eqref{eq:3.7} is a zero-mean sub-Gaussian process with respect to $\PiPhiGamma$ over $\FPiPhiGamma$ satisfying
\begin{equation}
    \mathbb{E}_{\tilde{\boldsymbol{\upsigma}}}\exp \left\{ {w\left[ {\kappa \left( {{{\it{\Pi}} _1} \circ {\Phi _{{{\it{\Gamma}}_1}}},{\tilde{\boldsymbol{\upsigma}}}} \right) - \kappa \left( {{{\it{\Pi}}_2} \circ {\Phi _{{{\it{\Gamma}}_2}}},{\tilde{\boldsymbol{\upsigma}}}} \right)} \right]} \right\} \le {\color{black} \exp \left[ {0.5{w^2}d_\kappa ^2\left( {{{\it{\Pi}}_1} \circ {\Phi _{{{\it{\Gamma}}_1}}},{{\it{\Pi}}_2} \circ {\Phi _{{{\it{\Gamma}}_2}}}} \right)} \right]}
    \label{eq:3.8}
\end{equation}
for $\forall w \in \mathbb{R}$ and $\forall\PiPhiGammaone$, $\forall\PiPhiGammatwo \in \FPiPhiGamma$, where the pseudo-metric $d_{\kappa}\left(\PiPhiGammaone, \PiPhiGammatwo\right)$ is
\begin{equation}
    {d_\kappa }\left( {{{\it{\Pi}} _1} \circ {\Phi _{{{\it{\Gamma}}_1}}},{{\it{\Pi}}_2} \circ {\Phi _{{{\it{\Gamma}}_2}}}} \right) = \sqrt {\sum\limits_{i = 1}^N {\left\| {{{\it{\Pi}}_1} \circ {\Phi _{{{\it{\Gamma}}_1}}}\left[ {{{\bf{u}}_i}(\tau )} \right](t) - {{\it{\Pi}}_2} \circ {\Phi _{{{\it{\Gamma}}_2}}}\left[ {{{\bf{u}}_i}(\tau )} \right](t)} \right\|_{{L^2}}^2} }.
    \label{eq:3.9}
\end{equation}
The diameter $\Delta_{\kappa}\left(\FPiPhiGamma\right)$ of $\kappa \left( {{\it{\Pi}}  \circ {\Phi _{\it{\Gamma}} },{\boldsymbol{\tilde \upsigma }}} \right)$ over $\FPiPhiGamma$ is bounded by
\begin{equation}
    {\Delta _\kappa }\left( {{{\cal F}_{{\it{\Pi}}\circ {\Phi _{\it{\Gamma}} }}}} \right) = \mathop {\sup }\limits_{{{\it{\Pi}}_1} \circ {\Phi _{{{\it{\Gamma}}_1}}},{{\it{\Pi}}_2} \circ {\Phi _{{{\it{\Gamma}}_2}}} \in {{\cal F}_{{\it{\Pi}}\circ {\Phi _{\it{\Gamma}}}}}} {d_\kappa }\left( {{{\it{\Pi}}_1} \circ {\Phi _{{{\it{\Gamma}}_1}}},{{\it{\Pi}}_2} \circ {\Phi _{{{\it{\Gamma}}_2}}}} \right) \le 2\sqrt {NT{w_{{\it{\Pi}},{\rm{out}}}}} {B_{\it{\Pi}}},
    \label{eq:3.10}
\end{equation}
where $\BPi$ is in Assumption 9.}

Given a neural oscillator $\hatPiPhiGamma$ learned from \textit{N} i.i.d. sample pairs $\mathbf{U}_N(t)$ and $\mathbf{V}_N(t)$ using Eq.~\eqref{eq:3.2}, from Theorem 11.3 of \citet{mohri2018foundations}, for $\forall\delta \in \left(0,1\right)$, the generalization error $\loss\left(\hatPiPhiGamma\right)$ of $\hatPiPhiGamma$, which is calculated using Eq.~\eqref{eq:3.3}, is bounded by
\begin{equation}
    \loss\left( {\hat \itPi  \circ {\Phi _{\hat \itGamma }}} \right) \le
    \underbrace{\hatloss\left( {\hat \itPi  \circ {\Phi _{\hat \itGamma }}} \right)}_{\text{approximation error}}+\underbrace{2{\Re _{{{\cal G}_\vartheta }}} + 3T{w_{\itPi ,{\rm{out}}}}B_{{\rm{loss}}}^2\sqrt {\frac{{{\color{black} \ln} \left( {2{\delta ^{ - 1}}} \right)}}{{2N}}}}_{\text{estimation error}}
    \label{eq:3.11}
\end{equation}
with probability at least $1 - \delta$. In Eq.~\eqref{eq:3.11}, the empirical loss $\hatloss\left( {\hat \itPi  \circ {\Phi _{\hat \itGamma }}} \right)$ can be bounded using the approximation error bounds in Lemma 1 or Lemma 2. The empirical Rademacher complexity ${\Re _{{{\cal G}_\vartheta }}}$ can be bounded by the expected supremum of the sub-Gaussian process $\kappa \left( {{\it{\Pi}}  \circ {\Phi _{\it{\Gamma}} },{\boldsymbol{\tilde \upsigma }}} \right)$, as demonstrated in Lemma 3. Based on the Dudley entropy integral bound (Theorem 8.23 of \citealp{foucart2013mathematical}), the expected supremum of $\kappa \left( {{\it{\Pi}}  \circ {\Phi _{\it{\Gamma}} },{\boldsymbol{\tilde \upsigma }}} \right)$ over $\FPiPhiGamma$ can be bounded by the covering number $N\left(\FPiPhiGamma, d_{\kappa}, \varepsilon\right)$ of $\FPiPhiGamma$
\begin{equation}
    {\mathbb{E}_{{\tilde{\boldsymbol \upsigma }}}}\mathop {\sup }\limits_{\itPi  \circ {\Phi _\itGamma } \in {{\cal F}_{\itPi  \circ {\Phi _\itGamma }}}} \kappa \left( {\itPi  \circ {\Phi _\itGamma },{\tilde{\boldsymbol \upsigma }}} \right) \le 4\sqrt 2 \int_0^{\sqrt {NT{w_{\itPi ,{\rm{out}}}}} {B_\itPi }} {\sqrt {\ln N\left( {{{\cal F}_{\itPi  \circ {\Phi _\itGamma }}},{d_\kappa },\varepsilon } \right)} {\rm{d}}\varepsilon }.
    \label{eq:3.12}
\end{equation}
For the sake of completeness, the proof of Eq.~\eqref{eq:3.12} is provided in \ref{appB}.

In what follows, the bound of the covering number $N\left(\FPiPhiGamma, d_{\kappa}, \varepsilon\right)$ is derived. Subsequently, the bound of the expected supremum of $\kappa \left( {{\it{\Pi}}  \circ {\Phi _{\it{\Gamma}} },{\boldsymbol{\tilde \upsigma }}} \right)$ is obtained using the bound of $N\left(\FPiPhiGamma, d_{\kappa}, \varepsilon\right)$. Finally, by substituting the bound of the expected supremum of $\kappa \left( {{\it{\Pi}}  \circ {\Phi _{\it{\Gamma}} },{\boldsymbol{\tilde \upsigma }}} \right)$ into Eq.~\eqref{eq:3.6} in Lemma 3, the bound of the generalization error $\loss\left(\hatPiPhiGamma\right)$ can be derived using Eq.~\eqref{eq:3.11}. 

{\bf Lemma 5:} {\it Under Assumptions 8 and 9, all $\itGamma(\cdot) \in \FGamma$ and $\itPi(\cdot) \in \FPi$ are Lipschitz continuous. $L_{\FGamma} = \wGammaout\wGamma\BW^2$ is an $L^1$-norm Lipschitz constant of all $\itGamma(\cdot) \in \FGamma$. $L_{\FPi} = \wPiout\wPi^{\hPi-1}\BtildeW^{\hPi}$ is an $L^1$-norm Lipschitz constant of all $\itPi(\cdot) \in \FPi$.}

{\bf Lemma 6:} {\it Under Assunptions 2, 8, 9, and 10, for all $\Phi_{\itGamma} \in \FPhiGamma$ and all $\mathbf{u}(t) \in K$, $\mathbf{x}(t) = \Phi_{\itGamma}\left[\mathbf{u}(\tau)\right](t)$ with the initial conditions $\mathbf{x}(0) = {\mathbf{x}}'(0) = \mathbf{0}$ satisfies
\begin{equation}
    \mathop {\max}\limits_{t \in [0,T]} \left[ {\left| {{\bf{x}}(t)} \right| + \left| {{\bf{x}}'(t)} \right|} \right] \le T\left[ {p{B_K}L_{\FGamma} + \wGammaout{B_{\bf{b}}}\left( {{w_\itGamma }{B_{\bf{W}}} + 1} \right)} \right]{e^{T\left( {L_{\FGamma} + 1} \right)}},
    \label{eq:3.13}
\end{equation}
where $L_{\FGamma}$ is an arbitrary $L^1$-norm Lipschitz constant for all $\itGamma(\cdot) \in \FGamma$.}

{\bf Lemma 7:} {\it Under Assunptions 1, 2, 8, 9, and 10, given MLPs $\itGamma_1(\cdot)$ and $\itGamma_2(\cdot)$ $\in \FGamma$, the differences between their weight matrices and vectors are bounded by $\varsigma$, that is $\left|{\mathbf{W}}_{2,i} - {\mathbf{W}}_{1,i}\right|_{\infty, \infty} \leq \varsigma$ and $\left|{\mathbf{b}}_{2,i} - {\mathbf{b}}_{1,i}\right|_{\infty} \leq \varsigma$, where ${\mathbf{W}}_{j,i}$ and ${\mathbf{b}}_{j,i}$ are respectively the weight matrix and vector of $\itGamma_j(\cdot)$, and i,j = 1 and 2, then, for all $\mathbf{u}(t) \in K$, the difference between ${\mathbf{x}}_1(t) = \Phi_{\itGamma_1}\left[\mathbf{u}(\tau)\right](t)$ and ${\mathbf{x}}_2(t) = \Phi_{\itGamma_2}\left[\mathbf{u}(\tau)\right](t)$, with the initial conditions ${\mathbf{x}}_1(0) = {\mathbf{x}}_1'(0) = {\mathbf{x}}_2(0) = {\mathbf{x}}_2'(0) = \mathbf{0}$, is bounded by
\begin{equation}
    \mathop {\max }\limits_{t \in [0,T]} \left| {{{\bf{x}}_2}(t) - {{\bf{x}}_1}(t)}\right|\le 2\varsigma {\wGammaout}\wGamma{pT^2}{e^{2T\left( {L_{\FGamma} + 1} \right)}} \left\{{\max\left(B_{\bf{W}},B_{\bf{b}}\right)\left[{{B_K}\left(L_{\FGamma}+1\right) + \wGammaout{B_{\bf{b}}}\left( {{w_\itGamma }{B_{\bf{W}}} + 1} \right)}+1\right]} +1\right\},    
    \label{eq:3.14}
\end{equation}
where $L_{\FGamma}$ is an arbitrary $L^1$-norm Lipschitz constant for all $\itGamma(\cdot) \in \FGamma$.}

{\bf Lemma 8:} {\it Under Assumptions 8 and 9, given MLPs $\itPi_1(\cdot)$ and $\itPi_2(\cdot)$ $\in \FPi$, the differences between their weight matrices and vectors are bounded by $\varsigma$, that is $\left|\tilde{\mathbf{W}}_{2,i} - \tilde{\mathbf{W}}_{1,i}\right|_{\infty,\infty} \leq \varsigma$ and $\left|\tilde{\mathbf{b}}_{2,i} - \tilde{\mathbf{b}}_{1,i}\right|_{\infty} \leq \varsigma$, where $\tilde{\mathbf{W}}_{j,i}$ and $\tilde{\mathbf{b}}_{j,i}$ are respectively the weight matrix and vector of $\itPi_j(\cdot)$, $i = 1, 2,..., \hPi$ and j = 1 and 2, then $\left| {{\itPi _1}({\bf{x}}) - {\itPi _2}({\bf{x}})} \right|$ is bounded by
\begin{equation}
     \left| {{\itPi _1}({\bf{x}}) - {\itPi _2}({\bf{x}})} \right| \leq \varsigma\max\left(\wGammaout,\wGamma\right)h_{\itPi}^2\left(L_{\FPi,\mathrm{layer}}^{h_\itPi - 1}+1\right)\left(\left| {\bf{x}} \right| + \wPi\Btildeb + 1\right),
    \label{eq:3.15}
\end{equation}
where $L_{\FPi,\mathrm{layer}}$ is an arbitrary $L^1$-norm Lipschitz constant that simultaneously bounds the mappings between all adjacent layers of all $\itPi(\cdot) \in \FPi$.}

{\bf Lemma 9:} {\it Under Assumptions 1, 2, 8, 9, and 10, given MLPs $\itGamma_1(\cdot)$ and $\itGamma_2(\cdot)$ $\in \FGamma$ and MLPs $\itPi_1(\cdot)$ and $\itPi_2(\cdot)$ $\in \FPi$, the differences between the weight matrices and vectors of $\itGamma_1(\cdot)$ and $\itGamma_2(\cdot)$ and those of $\itPi_1(\cdot)$ and $\itPi_2(\cdot)$ are bounded by $\varsigma$, as specified in Lemmas 7 and 8, then, for all $\mathbf{u}(t) \in K$, the difference between $\PiPhiGammaone\left[\mathbf{u}(\tau)\right](t)$ and $\PiPhiGammatwo\left[\mathbf{u}(\tau)\right](t)$, with the initial conditions ${\mathbf{x}}_1(0) = {\mathbf{x}}'_1(0) = {\mathbf{x}}_2(0) = {\mathbf{x}}'_2(0) = \mathbf{0}$, is bounded by
\begin{equation}
    \begin{aligned}
        &\mathop {\max }\limits_{t \in \left[ {0,T} \right]} \left| {{\itPi _1} \circ {\Phi _{{\itGamma _1}}}\left[ {{\bf{u}}(\tau )} \right](t) - {\itPi _2} \circ {\Phi _{{\itGamma _2}}}\left[ {{\bf{u}}(\tau )} \right](t)} \right|\\
        &\;\;\;\;\;\;\;\;\;\le 3\varsigma\left(L_{\FPi,\mathrm{layer}}^{h_\itPi}+1\right)\wPi{\wGammaout^2}\wGamma^2{h_{\itPi}^2}{pT^2}{e^{2T\left( {L_{\FGamma} + 1} \right)}}\left(L_{\FGamma} + 1\right){B_K}{\color{black} B\left(\Btildeb,\BW,\Bb\right)},
    \end{aligned}
    \label{eq:3.16}
\end{equation}
where $L_{\FGamma}$ is an arbitrary $L^1$-norm Lipschitz constant for all $\itGamma(\cdot) \in \FGamma$, $L_{\FPi,\mathrm{layer}}$ is an arbitrary inter-layer $L^1$-norm Lipschitz constant for all $\itPi(\cdot) \in \FPi$, and 
\begin{equation}
    {\color{black} B\left(\Btildeb,\BW,\Bb\right)} = \left[\max\left(B_{\bf{W}},B_{\bf{b}}\right)+1\right]\left[B_{\bf{b}}\left(B_{\bf{W}} + 1 \right)+2\right]+\Btildeb +2.
    \label{eq:3.17}
\end{equation}}

{\bf Lemma 10:} {\it Under Assumptions 1, 2, 8, 9, and 10, the expected supremum of $\kappa \left( {\itPi  \circ {\Phi _\itGamma },{\tilde{\boldsymbol \upsigma }}} \right)$ in Eq.~\eqref{eq:3.12} can be bounded by
\begin{equation}
    {\mathbb{E}_{{\tilde{\boldsymbol \upsigma }}}}\mathop {\sup }\limits_{\itPi  \circ {\Phi _\itGamma } \in {{\cal F}_{\itPi  \circ {\Phi _\itGamma }}}} \kappa \left( {\itPi  \circ {\Phi _\itGamma },{\tilde{\boldsymbol \upsigma }}} \right) \le 32\sqrt{2}{B_\itPi }w_{\mathrm{max}}^{1.5}\hPi\sqrt {NT\ln \left( {3 + 6B_{\mathrm{max}}{\Delta _{\itPi  \circ {\Phi _\itGamma }}}} \right)},
    \label{eq:3.18}
\end{equation}
where
\begin{equation}
    {\Delta_{\itPi  \circ {\Phi _\itGamma }}} = 30w_{\mathrm{max}}^{5.5}{h_{\itPi}^2}\left(L_{\FPi,\mathrm{layer}}^{h_\itPi}+1\right)\left(L_{\FGamma} + 1\right){T^2}{e^{2T\left( {L_{\FGamma} + 1} \right)}}\BPi^{-1}{B_K}\left(B_{\mathrm{max}}^3+1\right),
    \label{eq:3.19}
\end{equation}
$w_{\mathrm{max}} = \max\left(\wPiin,\wPi,\wPiout,\wGammain,\wGamma,\wGammaout\right)$, $B_{\mathrm{max}} = \max{\left(\BW,\Bb,\BtildeW,\Btildeb\right)}$, $L_{\FGamma}$ is an arbitrary $L^1$-norm Lipschitz constant for all $\itGamma(\cdot) \in \FGamma$, and $L_{\FPi,\mathrm{layer}}$ is an arbitrary inter-layer $L^1$-norm Lipschitz constant for all $\itPi(\cdot) \in \FPi$.}

{\color{black} Given a neural oscillator $\hatPiPhiGamma$ learned from \textit{N} i.i.d. sample pairs $\mathbf{U}_N(t)$ and $\mathbf{V}_N(t)$ using Eq.~\eqref{eq:3.2}, Eq.~\eqref{eq:3.11} indicates that its the generalization error $\loss\left(\hatPiPhiGamma\right)$ is bounded by the empirical loss $\hatloss\left( {\hat \itPi  \circ {\Phi _{\hat \itGamma }}} \right)$ and the Rademacher complexity ${\Re _{{{\cal G}_\vartheta }}}$. Since ${\hat \itPi  \circ {\Phi _{\hat \itGamma }}}$ minimizes the empirical loss $\hatloss\left( {\itPi  \circ {\Phi _{\itGamma }}} \right)$ over ${\itPi  \circ {\Phi _{\itGamma }}} \in \FPiPhiGamma$, where under Assumption 10 $\FPiPhiGamma$ represents a neural oscillator class whose elements $\PiPhiGamma$ are governed by Eq.~\eqref{eq:2.1} with $\itGamma(\cdot) \in \FGamma$ and $\itPi(\cdot) \in \FPi$, $\hatloss\left( {\hat \itPi  \circ {\Phi _{\hat \itGamma }}} \right)$ can be bounded by the approximation error bounds in either Lemma 1 or Lemma 2. In addition, Lemma 3 demonstrates that the Rademacher complexity ${\Re _{{{\cal G}_\vartheta }}}$ can be bounded by the expected supremum ${\mathbb{E}_{{\tilde{\boldsymbol \upsigma }}}}\mathop {\sup }\limits_{\itPi  \circ {\Phi _\itGamma } \in {{\cal F}_{\itPi  \circ {\Phi _\itGamma }}}} \kappa \left( {\itPi  \circ {\Phi _\itGamma },{\tilde{\boldsymbol \upsigma }}} \right)$ of the sub-Gaussian process $\kappa \left( {{\it{\Pi}}  \circ {\Phi _{\it{\Gamma}} },{\boldsymbol{\tilde \upsigma }}} \right)$. Lemma 10 provides an upper bound of ${\mathbb{E}_{{\tilde{\boldsymbol \upsigma }}}}\mathop {\sup }\limits_{\itPi  \circ {\Phi _\itGamma } \in {{\cal F}_{\itPi  \circ {\Phi _\itGamma }}}} \kappa \left( {\itPi  \circ {\Phi _\itGamma },{\tilde{\boldsymbol \upsigma }}} \right)$ in terms of the widths, depths, and Lipschitz constants of the MLPs $\itGamma(\cdot)$ and $\itPi(\cdot)$, as well as the time length $T$. Lemma 5 demonstrates that the Lipschitz constants $L_{\FGamma}$ and $L_{\FPi,\mathrm{layer}}$ can be calculated as $L_{\FGamma} = w_{\mathrm{max}}^2B_{\mathrm{max}}^2$ and $L_{\FPi,\mathrm{layer}} = w_{\mathrm{max}}^{\hPi}B_{\mathrm{max}}^{\hPi}$, respectively. By synthesizing these results, the generalization error bounds for the cases in Lemmas 1 and 2 are provided in following Theorems 1 and 2, respectively.}

{\bf Theorem 1:} {\it Under Assumptions 1-11 and the conditions in Lemma 1, for the target causal and uniformly continuous operator $\Phi$, with the specified sizes of $\itGamma(\cdot)$ and $\itPi(\cdot)$ in Lemma 1, the generalization error $\loss\left(\hatPiPhiGamma\right)$ of a neural oscillator $\hatPiPhiGamma$, that is learned by Eq.~\eqref{eq:3.2} with N i.i.d. sample pairs $\mathbf{U}_N(t)$ and $\mathbf{V}_N(t)$ from $\mu(\cdot)$, is bounded by
\begin{equation}
    \ell\left( {\hat \itPi  \circ {\Phi _{\hat \itGamma }}} \right) \le T\varepsilon _{\bf{y}}^2 + 3\frac{TqB_{{\rm{loss}}}^2}{{\sqrt N }}\left[86w_{\mathrm{max}}^{1.5}\hPi\sqrt{\ln \left( {3 + 6B_{\mathrm{max}}{\Delta _{\itPi  \circ {\Phi _\itGamma }}}} \right)}+\sqrt{0.5{{\color{black} \ln} \left( {2{\delta ^{ - 1}}} \right)}}\right]
    \label{eq:3.20}
\end{equation}
with probability at least $1-\delta$ for $\forall\delta \in \left(0,1\right)$, where $\varepsilon_{\mathbf{y}}$ is in Eq.~\eqref{eq:2.6} in Lemma 1, $\Delta_{\PiPhiGamma}$ is in Eq.~\eqref{eq:3.19} with $L_{\FGamma} = w_{\mathrm{max}}^2B_{\mathrm{max}}^2$, $L_{\FPi,\mathrm{layer}}^{\hPi} = w_{\mathrm{max}}^{\hPi}B_{\mathrm{max}}^{\hPi}$, {\color{black} $h_{\itPi} = H_{\itPi}+1$, $w_{\mathrm{max}}=\max\left[p\left(2M_{\itGamma}+1\right),q\left(pM_{\itGamma}+p+4 \right)\right]$, and $B_{\mathrm{max}} = \max{\left(\BW,\Bb,\BtildeW,\Btildeb\right)}$.}}

{\bf Theorem 2:} {\it Under Assumptions 1-11 and the conditions in Lemma 2, for the target second-order differential system governed by Eq.~\eqref{eq:2.7}, with the specified sizes of $\itGamma(\cdot)$ and $\itPi(\cdot)$ in Lemma 2, the generalization error $\ell\left(\hatPiPhiGamma\right)$ of $\hatPiPhiGamma$, that is learned by Eq.~\eqref{eq:3.2} with N i.i.d. sample pairs $\mathbf{U}_N(t)$ and $\mathbf{V}_N(t)$ from $\mu(\cdot)$, is bounded by
\begin{equation}
    \ell \left( {\hat{\it\Pi}  \circ {\Phi _{\hat \itGamma }}} \right) \le 16T\left( {\frac{{{\color{black} L^2_{\mathbf{h}}B_{\beta _{\mathbf{g}}}^2}r^2{C_\itGamma }}}{{w_\itGamma } - 8r} + \frac{{q^3{C_\itPi }}}{{w_\itPi } - 8q}} \right) + 3\frac{TqB_{{\rm{loss}}}^2}{{\sqrt N }}\left[172w_{\mathrm{max}}^{1.5}\sqrt{\ln \left( {3 + 6B_{\mathrm{max}}{\Delta _{\itPi  \circ {\Phi _\itGamma }}}} \right)}+\sqrt{0.5{{\color{black} \ln} \left( {2{\delta ^{ - 1}}} \right)}}\right] 
    \label{eq:3.21}
\end{equation}
with probability at least $1-\delta$ for $\forall\delta \in \left(0,1\right)$, where $\wGamma > 8\alpha^{-2}r^3C_{\itGamma}+8r$, $\wPi > 8q$, $\Delta_{\PiPhiGamma}$ is in Eq.~\eqref{eq:3.19} with $\hPi = 2$, $L_{\FGamma} = w_{\mathrm{max}}^2B_{\mathrm{max}}^2$, {\color{black} $L_{\FPi,\mathrm{layer}}^{\hPi} = w_{\mathrm{max}}^{\hPi}B_{\mathrm{max}}^{\hPi}$, $B_{\mathrm{max}} = \max{\left(\BW,\Bb,\BtildeW,\Btildeb\right)}$, $w_{\mathrm{max}} = \max\left(2r+p,\wPi,\wGamma\right)$}, and {\color{black} $B_{\beta _{\mathbf{g}}}$, $C_{\itGamma}$, and $C_{\itPi}$} are specified in Lemma 2.}

{\bf Remark 2:} {\color{black} The PAC generalization bounds for RNNs and their variants have been derived in \citet{chen2020generalization}, but the analysis only considers the mappings from input sequences to the RNN output vectors at the final time instant. Similarly, \citet{marion2023generalization} established a bound of parameterized neural ODEs for mappings from input vectors to output vectors at the final time instant of the ODE solutions.} An upper generalization bound for a deep SS model has been derived by leveraging the Rademacher complexity combined with a layer-by-layer Rademacher contraction approach \citep{racz2024length}. The considered deep SS model produces a scalar output through pooling and its derived estimation error grows exponentially with the depth of a utilized MLP. {\color{black} In contrast to above results,} the outputs of the neural oscillator {\color{black} in this study} are continuous temporal functions. {\color{black} Lemmas 3 and 4 prove that $\kappa \left( {{\it{\Pi}}  \circ {\Phi _{\it{\Gamma}} },{\boldsymbol{\tilde \upsigma }}} \right)$ in Eq.~\eqref{eq:3.7} is a zero-mean sub-Gaussian process with respect to $\PiPhiGamma$ over $\FPiPhiGamma$, thereby enabling the Rademacher complexity framework to be applied for deriving the generalization error bounds of the neural oscillator.} The exponential terms in $\Delta_{\PiPhiGamma}$ of Eq.~\eqref{eq:3.19}, which contain the MLP sizes, parameter bound $B_{\mathrm{max}}$, and time length $T$, are encapsulated within the logarithmic functions in Eqs.~\eqref{eq:3.20} and \eqref{eq:3.21}. Consequently, the estimation errors of the generalization bounds established in Theorems 1 and 2 grow polynomially with the sizes and parameter values of the employed MLPs $\itGamma(\cdot)$ and $\itPi(\cdot)$, as well as the time length $T$. In the {\color{black} numerical studies} conducted by \citet{huang2025upper}, the numerical generalization errors can clearly illustrate the decaying approximation error curves of the neural oscillator with the increasing MLP sizes. These results indicate that increasing the MLP sizes has only a marginal influence on the generalization errors, which is consistent with the theoretical results in Theorems 1 and 2 of this study.

{\bf Remark 3:} \citet{lanthaler2024operator} has proved that decreasing the required number of parameters in neural operators using super-expressive neural networks, which aims to mitigate the curse of parametric complexity,
may come at the expense of requiring larger model parameter values. The generalization bounds in Theorems 1 and 2 increase with the growing bound $B_{\mathrm{max}}$ of the MLP parameters. Therefore, whether employing super-expressive neural networks can finally decrease the generalization error of the neural oscillator remains an open question. Theorem 10. 4 in \citet{foucart2013mathematical} proves that the Gelfand widths of the normed spaces containing a target operator can be employed to establish lower generalization bounds for all models that approximate this target operator. This method has been employed to prove that polynomial-based neural operators attain the optimal rate of sample complexity for the approximation of Lipschitz operators \citep{adcock2025samplecomplexitylearninglipschitz}. Utilizing this method to derive lower generalization bounds for the neural oscillator represents an interesting direction for future research.

{\color{black} The error bounds established in Theorems 1 and 2 can be further extended to the uncertainty quantification of quantities of interest calculated from the temporal output functions of target causal operators and the corresponding learned neural oscillators. In many engineering and scientific applications, the practical objects of statistical inference are not the entire output functions but rather scalar or vector-valued quantities of interest extracted from them, such as integrated energies, peak responses, modal coordinates, and extreme-value statistics. The discrepancy between the probability measures of these quantities of interest calculated from the target causal operators and the corresponding learned neural oscillators directly governs the reliability of risk assessments and predictive analyses. Let $\Psi: \Czero{q} \to \mathbb{R}^s$ be a target Lipschitz continuous functional mapping satisfying $\left|\Psi\left[\mathbf{v}_1(t)\right] - \Psi\left[\mathbf{v}_2(t)\right]\right|_2 \le L_{\Psi}\normtwo{\mathbf{v}_1(t) - \mathbf{v}_2(t)}$ for $\forall \mathbf{v}_1(t), \mathbf{v}_2(t) \in \Czero{q}$, where $L_{\Psi} > 0$ is its $L^2$-norm Lipschitz constant. The probability measures of the random vectors $\Psi\left\{\Phi\left[\mathbf{u}(\tau)\right](t)\right\}$ and $\Psi\left\{\hatPiPhiGamma\left[\mathbf{u}(\tau)\right](t)\right\}$ on $\mathbb{R}^s$, which are induced by $\mathbf{u}(t) \sim \mu(\cdot)$ in Assumption 7, are denoted by $\rho_{\Phi}$ and $\rho_{\hatPiPhiGamma}$, respectively. By leveraging the Kantorovich-Rubinstein duality (Theorem 11.8.2 of \citealt{dudley2018real}) together with Jensen's inequality, the squared Wasserstein-1 distance $W_1^2\left(\rho_{\Phi}, \rho_{\hatPiPhiGamma}\right)$ between $\rho_{\Phi}$ and $\rho_{\hatPiPhiGamma}$ is bounded by $W_1^2\left(\rho_{\Phi}, \rho_{\hatPiPhiGamma}\right) \le L_{\Psi}^2 \loss\left(\hatPiPhiGamma\right)$. Substituting the upper bounds of $\loss\left(\hatPiPhiGamma\right)$ from Theorems 1 and 2 into this inequality yields the generalization error bounds for $W_1^2\left(\rho_{\Phi}, \rho_{\hatPiPhiGamma}\right)$, which take the same form as those in Theorems 1 and 2 up to the multiplicative constant $L_{\Psi}^2$. These two error bounds are formally stated in the following two corollaries.

{\bf Corollary 1:} {\it Under the assumptions and conditions in Theorem 1, for the target causal and uniformly continuous operator $\Phi$, the neural oscillator $\hatPiPhiGamma$ learned from $N$ i.i.d. sample pairs, and any Lipschitz continuous mapping $\Psi: \Czero{q} \to \mathbb{R}^s$ with an $L^2$-norm Lipschitz constant $L_{\Psi}$, the squared Wasserstein-1 distance $W_1^2\left(\rho_{\Phi}, \rho_{\hatPiPhiGamma}\right)$ between the probability measures $\rho_{\Phi}$ of $\Psi\left\{\Phi\left[\mathbf{u}(\tau)\right](t)\right\}$ and $\rho_{\hatPiPhiGamma}$ of $\Psi\left\{\hatPiPhiGamma\left[\mathbf{u}(\tau)\right](t)\right\}$ induced by $\mathbf{u}(t) \sim \mu(\cdot)$ is bounded by
\begin{equation}
    W_1^2\left(\rho_{\Phi}, \rho_{\hatPiPhiGamma}\right) \le L_{\Psi}^2\left\{T\varepsilon_{\mathbf{y}}^2 + 3\frac{TqB_{\rm{loss}}^2}{\sqrt{N}}\left[86w_{\mathrm{max}}^{1.5}\hPi\sqrt{\ln\left(3 + 6B_{\mathrm{max}}\Delta_{\PiPhiGamma}\right)} + \sqrt{0.5\ln\left(2\delta^{-1}\right)}\right]\right\}
    \label{eq:3.22}
\end{equation}
with probability at least $1-\delta$ for $\forall\delta \in \left(0,1\right)$,  where all variables and coefficients are the same as those specified as in Theorem 1.}

{\bf Corollary 2:} {\it Under the assumptions and conditions in Theorem 2, for the target second-order differential system governed by Eq.~\eqref{eq:2.7} denoted as $\hat{\mathbf{y}}(t) = \Phi\left[\mathbf{u}(\tau)\right](t)$, the neural oscillator $\hatPiPhiGamma$ learned from $N$ i.i.d. sample pairs, and any Lipschitz continuous mapping $\Psi: \Czero{q} \to \mathbb{R}^s$ with an $L^2$-norm Lipschitz constant $L_{\Psi}$, the squared Wasserstein-1 distance $W_1^2\left(\rho_{\Phi}, \rho_{\hatPiPhiGamma}\right)$ between the probability measures $\rho_{\Phi}$ of $\Psi\left\{\Phi\left[\mathbf{u}(\tau)\right](t)\right\}$ and $\rho_{\hatPiPhiGamma}$ of $\Psi\left\{\hatPiPhiGamma\left[\mathbf{u}(\tau)\right](t)\right\}$ induced by $\mathbf{u}(t) \sim \mu(\cdot)$ is bounded by
\begin{equation}
    W_1^2\left(\rho_{\Phi}, \rho_{\hatPiPhiGamma}\right) \le L_{\Psi}^2\left\{16T\left(\frac{L_{\mathbf{h}}^2 B_{\beta_{\mathbf{g}}}^2 r^2 C_{\itGamma}}{w_{\itGamma}-8r} + \frac{q^3 C_{\itPi}}{w_{\itPi}-8q}\right) + 3\frac{TqB_{\rm{loss}}^2}{\sqrt{N}}\left[172w_{\mathrm{max}}^{1.5}\sqrt{\ln\left(3+6B_{\mathrm{max}}\Delta_{\PiPhiGamma}\right)} + \sqrt{0.5\ln\left(2\delta^{-1}\right)}\right]\right\}
    \label{eq:3.23}
\end{equation}
with probability at least $1-\delta$ for $\forall\delta \in \left(0,1\right)$, where all variables and coefficients are the same as those specified as in Theorem 2.}}

The estimation errors in Theorems 1 and 2 can be expressed in the form of $N^{-0.5}O\left\{\left[{L_{\FGamma} + \ln{\left(L_{\FGamma}\right)}+\hPi\ln{\left(L_{\FPi,\mathrm{layer}}\right)}}\right]^{0.5}\right\}$, which can be interpreted from a sample-dependent perspective \citep{wei2019data}. During a practical training process, each {\color{black} trainable} parameter in the matrices and vectors of $\itGamma(\cdot)$ and $\itPi(\cdot)$ varies within a bounded range and attains a largest absolute value. Thus, for each pair of $\itGamma(\cdot)$ and $\itPi(\cdot)$ learned from $\hatloss\left(\PiPhiGamma\right)$ in Eq.~\eqref{eq:3.2},
the MLP spaces $\FGamma$ and $\FPi$ in Assumption 9 can be narrowed to the ones determined by the largest absolute values of all {\color{black} trainable} MLP parameters during the training process, under which all above theoretical results remain valid. In this setting, $L_{\FGamma}$ and $L_{\FPi,\mathrm{layer}}$ are the $L^1$-norm Lipschitz constants of these two narrowed MLP spaces and they can be much smaller than $\wGammaout\wGamma\BW^2$ and $\max{\left(\wPiout,\wPi\right)}\BtildeW$, respectilvey. Therefore, constraining the Lipschitz constants of $\itGamma(\cdot)$ and $\itPi(\cdot)$ rather than the maximum parameter bound $B_{\mathrm{max}}$ provides an effective approach to reducing the generalization error of a neural oscillator learned from samples. Motivated by this observation, a loss function containing explicit constraints on the Lipschitz constants of $\itGamma(\cdot)$ and $\itPi(\cdot)$ is introduced
{\color{black} \begin{equation}
    \tilde{\loss}_{{\mathbf{U}}_N, {\mathbf{V}}_N}\left(\PiPhiGamma\right) = \frac{1}{N}\sum_{i=1}^{N}\normtwo{{\mathbf{v}}_i(t)-\PiPhiGamma\left[\mathbf{u}_i(\tau)\right](t)}^2 + 
    \frac{\lambda_L}{\sqrt{N}}\left(L_{\itGamma} +\hPi{L_{\itPi,\mathrm{layer}}}\right),
    \label{eq:3.24}
\end{equation}}where $\lambda_L$ {\color{black} is} a manually determined parameter, $L_{\itGamma}$ is the $L^1$-norm Lipschitz constant of $\itGamma(\cdot)$, and $L_{\itPi,\mathrm{layer}}$ is the inter-layer $L^1$-norm Lipschitz constant of $\itPi(\cdot)$. Existing methods for constraining Lischitz constants of MLPs, including weight matrix norm regularization methods \citep{yoshida2017spectral}, weight matrix normalization methods \citep{gouk2021regularisation,liu2022learning}, and Jacobian regularization methods \citep{hoffman2019robust,wei2019data}, can be directly employed to constrain $L_{\itGamma}$ and $L_{\itPi,\mathrm{layer}}$. Since the ReLU MLPs $\itGamma(\cdot)$ and $\itPi(\cdot)$ are piecewise affine functions \citep{hanin2019universal}, decreasing the number of affine segments can reduce the Lipschitz constants of $\itGamma(\cdot)$ and $\itPi(\cdot)$ learned from samples. In this study, $L_{\itGamma}$ and $\hPi{L_{\itPi,\mathrm{layer}}}$ are respectively replaced by $\sum_{i=1}^{2}\left(\left|\mathbf{W}_i\right|_{1,1}+\left|\mathbf{b}_i\right|_{1} \right)$ and $\sum_{j=1}^{\hPi}\left(\left|\tilde{\mathbf{W}}_j\right|_{1,1}+\left|\tilde{\mathbf{b}}_j\right|_{1} \right)$ to enhance the sparsity of $\itGamma(\cdot)$ and $\itPi(\cdot)$, thereby reducing their Lipschitz constants.

\section{{\color{black} Numerical Studies}}
\label{sec4}
In this section, a Bouc-Wen nonlinear system subjected to stochastic seismic excitation, which exhibits plastic deformation responses \citep{huang2025upper}, is employed to validate the power laws of the estimation errors in the
generalization bounds with respect to the sample size $N$ and time length $T$, and to demonstrate the effectiveness of $\tilde{\loss}_{{\mathbf{U}}_N, {\mathbf{V}}_N}$ in {\color{black} Eq.~\eqref{eq:3.24}} for improving the generalization capability of the neural oscillator under limited training samples. The differential equation of the five-degree of freedom Bouc-Wen system is
{\color{black} \begin{equation}
    \left\{
    \begin{aligned}
        &\mathbf{M}\mathbf{X}''(t) + \mathbf{C}\mathbf{X}'(t) + \lambda{\mathbf{K}\mathbf{X}}(t) + (1 - \lambda ){\tilde{\mathbf{K}}\mathbf{Z}}(t) =  - {\mathbf{M}}{U_{\text{e}}}(t)\\
        &{Z_i'}(t) = \tilde{X}_i'(t) - \beta \left|\tilde{X}_i'(t)\right|{\left|{Z_i}(t)\right|^{s - 1}}{Z_i}(t) - \gamma \tilde{X}_i'(t){\left| {{Z_i}(t)}\right|^s},\;\;i = 1,2, \ldots,5
    \end{aligned}
    \right.,
    \label{eq:4.1}
\end{equation}}
where ${\mathbf{X}} = \left[X_1(t), X_2(t),\dots,X_5(t)\right]^\top$, $\tilde{\bf{X}}(t) = [X_1(t), X_2(t)-X_1(t),$ $X_3(t)-X_2(t),\dots,X_5(t)-X_4(t)]^\top$, ${\mathbf{Z}} = \left[Z_1(t), Z_2(t),\dots,Z_5(t)\right]^\top$, ${\mathbf{X}}(0)={\mathbf{X}}'(0)={\mathbf{Z}}(0)={\mathbf{0}}$,
\begin{equation}
    {\mathbf{M}} = 
    \begin{bmatrix}
        m & 0 & 0 & 0 & 0 \\
        0 & m & 0 & 0 & 0 \\
        0 & 0 & m & 0 & 0 \\
        0 & 0 & 0 & m & 0 \\
        0 & 0 & 0 & 0 & m
    \end{bmatrix},\;\;\;
    {\mathbf{K}} = 
    \begin{bmatrix}
        2k & -k & 0 & 0 & 0 \\
        -k & 2k & -k & 0 & 0 \\
        0 & -k & 2k & -k & 0 \\
        0 & 0 & -k & 2k & -k \\
        0 & 0 & 0 & -k & k
    \end{bmatrix},\;\;\;
    {\tilde{{\mathbf{K}}}} = 
    \begin{bmatrix}
        k & -k & 0 & 0 & 0 \\
        0 & k & -k & 0 & 0 \\
        0 & 0 & k & -k & 0 \\
        0 & 0 & 0 & k & -k \\
        0 & 0 & 0 & 0 & k
    \end{bmatrix},\;\;\;
    \label{eq:4.2}
\end{equation}
$m=1382.4$ kilogram, $k=1.7 \times 10^6$ Newton/meter, $\mathbf{C}$ is determined using $\mathbf{M}$ and $\mathbf{K}$ such that the damping ratios of all response modes are 0.05, $\lambda = 0.01$, $\beta = 2$, $\gamma = 2$, and $s = 3$, and $U_{\text{e}}(t)$ is a stochastic Gaussian harmonizable earthquake ground motion acceleration whose Wigner-Ville spectrum is
\begin{equation}
    {W_{\text{e}}}(t,f) = 2500{t^2}{f^2}\exp \left[ { - 0.3(1 + {f^2})t} \right].
    \label{eq:4.3}
\end{equation}
The ODE system in Eq.~\eqref{eq:4.1} is solved numerically using a fourth-order Runge-Kutta method \citep{griffiths2010numerical} with a time discretization of $\Delta t = 0.01$ seconds. For the numerical implementation of training, the neural oscillator is discretized using a second-order Runge–Kutta scheme. The approach for training the neural oscillator with input-output pair samples is briefly described in \ref{appC}. {\color{black} In this section, the mapping from $U_{\text{e}}(t)$ to the response $X_5(t)$ and that from $U_{\text{e}}(t)$ to the extreme-value process $E_{X_5}(t) = \max_{\tau \in \left[0, t\right]}\left|X_5(\tau)\right|$ of $X_{5}(t)$ are learned using the neural oscillator with several independent sets of simulated sample pairs. These two mappings are investigated in the following Cases 1 and 2, respectively. All code for the numerical experiments in this section is available at: \url{https://github.com/ZifengH22/Upper-Generalization-Bounds-for-Neural-Oscillators}}.
{\color{black} \subsection{Case 1}
\label{sec:case1}
For learning the mapping from $U_{\text{e}}(t)$ to $X_5(t)$, $5 \times 10^4$ independent discrete-time samples $U_{\text{e},l}(t_i)$ of $U_{\text{e}}(t)$ over a time interval $[0, 10]$ seconds are simulated, where $i = 0, 1, 2, \dots, 999$ and $l = 1, 2, \dots, 5 \times 10^4$. Their induced $5\times10^4$ samples $X_{5,l}(t_i)$ of $X_5(t)$ are computed by numerically solving the ODE in Eq.~\eqref{eq:4.1}. $\itGamma(\cdot)$ and $\itPi(\cdot)$ in Eq.~\eqref{eq:2.1} are one-hidden layer MLPs. Their activation functions are $\sigma_{\mathrm{ReLU}}(\cdot)$. The widths of input, hidden, and output layers of $\itGamma(\cdot)$ are 21, 40, and 10, respectively. The widths of input, hidden, and output layers of $\itPi(\cdot)$ are 12, 20, and 1, respectively. Four independent sets of randomly selected training and validation samples are employed to train the neural oscillator with the loss function $\hat\ell_{\lambda_L}$ in Eq.~\eqref{eq:c3}, both with ($\lambda_L = 0.002$) and without ($\lambda_L = 0$) the parameter-norm constraint. Each of the four sets consists of six datasets, which respectively contain (i) 100 training and 20 validation samples, (ii) 200 training and 50 validation samples, (iii) 400 training and 100 validation samples, (iv) 800 training and 200 validation samples, (v) 1600 training and 400 validation samples, and (vi) 3200 training and 800 validation samples. All training phases consist of 2000 epochs, each epoch contains 32 batches with a batch size of 100. For each training phase, the initial learning rate is set to 0.01, with a decay period of 100 epochs and a decay factor of 0.965. A gradient-norm clipping strategy \citep{pascanu2013difficulty} with a threshold of 1 is applied to all MLPs to prevent gradient explosion. $\itGamma(\cdot)$ and $\itPi(\cdot)$ are initialized using identical random seeds to ensure consistent initialization across experiments. In each training phase, the neural oscillator network achieving the smallest training loss is selected as the final model.

For each trained neural oscillator, its $5 \times 10^4$ prediction samples $\tilde{X}_{5,l}(t_i)$ driven by the input samples $U_{\text{e},l}(t_i)$ are simulated. The relative generalization error $\tilde{\varepsilon}_{X,2}$ is computed as
\begin{equation}
    \tilde{\varepsilon}_{X,2} = \frac{\sum\limits_{l = 1}^{50000}{\sum\limits_{i = 0}^{999} \left[{{X}_{5,l}}(t_i) - \tilde{X}_{5,l}(t_i)\right]^2}}{\sum\limits_{l = 1}^{50000}\sum\limits_{i = 0}^{999} X_{5,l}^2(t_i)}.
    \label{eq:4.4}
\end{equation}
The decays of $\tilde{\varepsilon}_{X,2}$ with respect to the increasing sample number $N$ are shown in Figure~\ref{fig:1}, where the markers represent the average values of $\tilde{\varepsilon}_{X,2}$ across the four independent sets of randomly selected training samples, and the upper and lower bars indicate the largest and smallest values, respectively. It can be observed that, for small sample sizes ($N \in \left[100,200\right]$), the convergence rate of approximately $-2.8$ obtained from the numerical results is steeper than the theoretical value of $-0.5$ in Theorem 1. As the sample number increases ($N \in \left[400,3200\right]$), the convergence rate of the numerical errors coincides with the theoretical value of $-0.5$. Since the established theoretical result in Theorem 1 provides an upper generalization bound, the numerical results in Figure~\ref{fig:1} confirm the validity of the convergence rate of $-0.5$. For the small sample sizes ($N \in \left[100,200\right]$), the numerical generalization errors are significantly reduced by constraining the $L^1$-norms of the matrices and vectors of $\itGamma(\cdot)$ and $\itPi(\cdot)$ during training. This validates the effectiveness of $\tilde{\loss}_{{\mathbf{U}}_N, {\mathbf{V}}_N}$ in Eq.~\eqref{eq:3.24} for decreasing the generalization error of the neural oscillator when training data are limited.

In addition to the relative generalization error $\tilde{\varepsilon}_{X,2}$ in Eq.~\eqref{eq:4.4}, the averaged squared Wasserstein-1 distance over the time interval $[0, 10]$ seconds between the probability measures $\rho_{X_5(t_i)}$ and $\rho_{\tilde{X}_5(t_i)}$ of $X_5(t_i)$ and $\tilde{X}_5(t_i)$, respectively, is also evaluated. At each time point $t_i$, the squared Wasserstein-1 distance is computed from the $5\times 10^4$ samples $X_{5,l}(t_i)$ and $\tilde{X}_{5,l}(t_i)$ via the sorted-sample formula
\begin{equation}
    W_{1}^{2}\!\left[\rho_{X_5(t_i)},\,\rho_{\tilde{X}_5(t_i)}\right] = \left[\frac{1}{50000}\sum_{l=1}^{50000}\left|X_{5,(l)}(t_i) - \tilde{X}_{5,(l)}(t_i)\right|\right]^{2},
    \label{eq:4.5}
\end{equation}
where $X_{5,(1)}(t_i) \le X_{5,(2)}(t_i) \le \cdots \le X_{5,(50000)}(t_i)$ and $\tilde{X}_{5,(1)}(t_i) \le \tilde{X}_{5,(2)}(t_i) \le \cdots \le \tilde{X}_{5,(50000)}(t_i)$ denote the order statistics of $\{X_{5,l}(t_i)\}_{l=1}^{50000}$ and $\{\tilde{X}_{5,l}(t_i)\}_{l=1}^{50000}$, respectively. The averaged squared Wasserstein-1 distance over $[0, 10]$ seconds is then computed as
\begin{equation}
    \bar{W}_{1,X}^{2} = \frac{1}{1000}\sum_{i=0}^{999} W_1^{2} \left[\rho_{X_5(t_i)},\,\rho_{\tilde{X}_5(t_i)}\right].
    \label{eq:4.6}
\end{equation}
This quantity is used to numerically validate the generalization bound for the squared Wasserstein-1 distance between the probability measures of quantities of interest established in Corollary 1. The decay of $\bar{W}_{1,X}^{2}$ with respect to the increasing sample number $N$ is shown in Figure~\ref{fig:2}, where the markers represent the average values of $\bar{W}_{1,X}^{2}$ across the four independent sets of randomly selected training samples, and the upper and lower bars indicate the largest and smallest values, respectively. The convergence behavior of $\bar{W}_{1,X}^{2}$ aligns with that of $\tilde{\varepsilon}_{X,2}$ in Figure~\ref{fig:1}. As the sample number increases ($N \in \left[400, 3200\right]$), the convergence rate of $\bar{W}_{1,X}^{2}$ coincides with the theoretical value of $-0.5$ predicted by Corollary 1. Moreover, constraining the $L^1$-norms of the matrices and vectors of $\itGamma(\cdot)$ and $\itPi(\cdot)$ during training (with $\lambda_L = 0.002$) reduces $\bar{W}_{1,X}^{2}$ for small sample sizes, which is in agreement with the observation for $\tilde{\varepsilon}_{X,2}$ in Figure~\ref{fig:1}.
\begin{figure}[t]
    \centering
    \begin{minipage}[t]{0.48\linewidth}
        \centering
        \includegraphics[width=\linewidth]{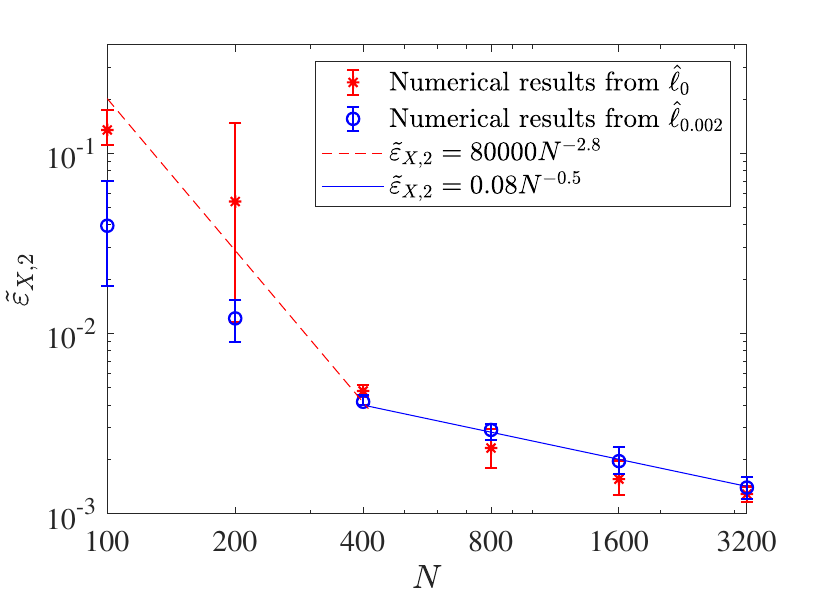}
        \caption{{\color{black} Relative generalization error $\tilde{\varepsilon}_{X,2}$ between $X_{5,l}(t_i)$ and $\tilde{X}_{5,l}(t_i)$ versus the sample number $N$.}}
        \label{fig:1}
    \end{minipage}
    \hfill
    \begin{minipage}[t]{0.48\linewidth}
        \centering
        \includegraphics[width=\linewidth]{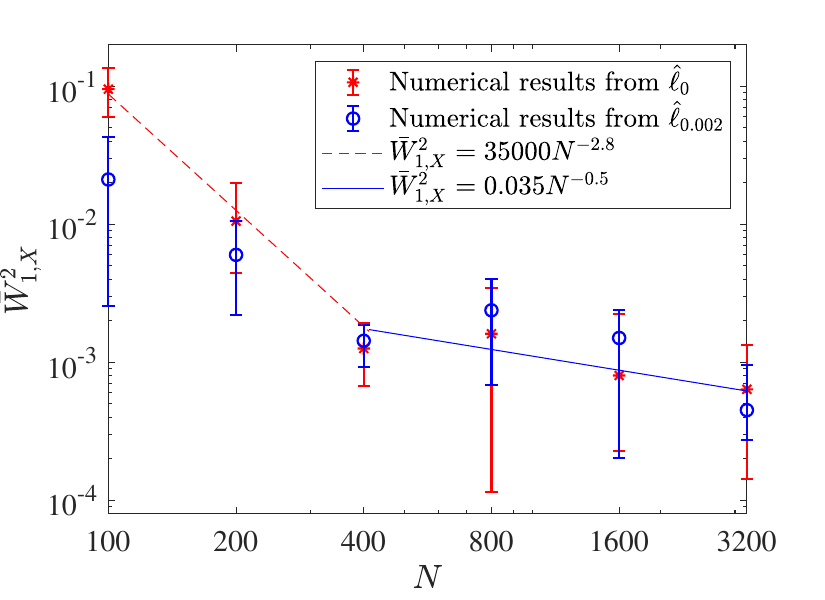}
        \caption{{\color{black} Averaged squared Wasserstein-1 distance $\bar{W}_{1,X}^{2}$ between the probability measures $\rho_{X_5(t_i)}$ of $X_{5,l}(t_i)$ and $\rho_{\tilde{X}_5(t_i)}$ of $\tilde{X}_{5,l}(t_i)$ over the time interval $[0, 10]$ seconds versus the sample number $N$.}}
        \label{fig:2}
    \end{minipage}
\end{figure}}

{\color{black} \subsection{Case 2:}
\label{sec:case2}
For learning the mapping from $U_{\text{e}}(t)$ to $E_{X_5}(t)$, $5 \times 10^4$ independent discrete-time samples $U_{\text{e},l}(t_i)$ of $U_{\text{e}}(t)$ over a time interval $[0, 30]$ seconds, where $i = 0, 1, 2, \dots, 2999$ and $l = 1, 2, \dots, 5 \times 10^4$, are simulated. Their induced $5\times10^4$ samples $E_{X_5,l}(t_i)$ of $E_{X_5}(t)$ are computed by $E_{X_5,l}(t_i) = \max_{\tau \in \left[0, t_i\right]}\left|X_{5,l}(\tau)\right|$, where $X_{5,l}(t_i)$ is the sample of $X_5(t)$ computed by numerically solving the ODE in Eq.~\eqref{eq:4.1}. $\itGamma(\cdot)$ in Eq.~\eqref{eq:2.1} is a one-hidden layer MLP. The widths of input, hidden, and output layers of $\itGamma(\cdot)$ are 11, 20, and 10, respectively. For $\itPi(\cdot)$, its input, hidden, and output layer widths are 12, 15, and 1, respectively, and its hidden-layer number is $2$. The activation functions of $\itGamma(\cdot)$ and $\itPi(\cdot)$ adopt the Parametric Rectified Linear Unit (PReLU) $\sigma_{\text{PReLU}}(x)=\max(0,x) + \alpha\min(0,x)$ to facilitate the training process, where $\alpha$ is a parameter learned during training \citep{he2015delving}. Since $\sigma_{\mathrm{PReLU}}(\cdot)$ is also a piecewise linear activation function with bounded first-order derivatives, the Lipschitz constants $L_{\FGamma}$ and $L_{\FPi,\mathrm{layer}}$ in Theorems 1 and 2, and Corollaries 1 and 2, can be modified by a multiplicative constant. Consequently, the estimation errors in Theorems 1 and 2, and Corollaries 1 and 2, which can be expressed in the form of $N^{-0.5}O\left\{\left[{L_{\FGamma} + \ln{\left(L_{\FGamma}\right)}+\hPi\ln{\left(L_{\FPi,\mathrm{layer}}\right)}}\right]^{0.5}\right\}$, remain valid up to a constant factor.

To analyze the variation of the generalization error of the neural oscillator with respect to the increasing time length $T$, four independent sets of randomly selected training and validation samples are employed to train the neural oscillator with the loss function $\hat\ell_{\lambda_L}$ in Eq.~\eqref{eq:c3} without the parameter-norm constraint ($\lambda_L = 0$). Each of the four samples sets consists of six datasets corresponding to the time lengths of 5 seconds, 10 seconds, 15 seconds, 20 seconds, 25 seconds, and 30 seconds, respectively. Each dataset contains 1600 training and 400 validation samples. All training phases consist of 3000 epochs, each containing 16 batches with a batch size of 100. For each training phase, the learning rate first linearly increases from 0.0005 to 0.02 within the first 40 epochs, and subsequently decreases exponentially with a decay period of 100 epochs and a decay factor of 0.95. A gradient-norm clipping strategy \citep{pascanu2013difficulty} with a threshold of 1 is applied to all MLPs to prevent gradient explosion. $\itGamma(\cdot)$ and $\itPi(\cdot)$ are initialized using identical random seeds to ensure consistent initialization across experiments. The neural oscillator network achieving the smallest training loss is selected as the final model. Among the four independent sets, for each $T \in \{5, 10, 15, 20\}$ seconds, the training run fails to converge for one set, leaving three convergent sets; for each $T \in \left\{25, 30\right\}$ seconds, the training runs fail to converge for three sets, leaving one convergent set. The numerical results reported below are computed from the convergent sets only.

For each trained neural oscillator, its $5 \times 10^4$ prediction samples $\tilde{E}_{X_5,l}(t_i)$ driven by the input samples $U_{\text{e},l}(t_i)$ are simulated. The relative generalization error $\tilde{\varepsilon}_{E,2}$ is computed as
\begin{equation}
    \tilde{\varepsilon}_{E,2} = \frac{\sum\limits_{l = 1}^{50000}{\sum\limits_{i = 0}^{100T-1} \left[{{E}_{X_5,l}}(t_i) - \tilde{E}_{X_5,l}(t_i)\right]^2}}{\sum\limits_{l = 1}^{50000}\sum\limits_{i = 0}^{2999} E_{X_5,l}^2(t_i)}.
    \label{eq:4.7}
\end{equation}
The time length $T$ in Eq.~\eqref{eq:4.7} captures the impact of increasing $T$ on the generalization error $\loss\left(\hatPiPhiGamma\right)$ in Eq.~\eqref{eq:3.3}. As illustrated in Figure~\ref{fig:3}, where the markers represent the average values of $\tilde{\varepsilon}_{E,2}$ across the convergent sets and the upper and lower bars indicate the largest and smallest values, respectively, the variation of $\tilde{\varepsilon}_{E,2}$ with respect to the increasing time length $T$ can be approximated by a power-law function with an exponent of $1.5$, which is consistent with the leading term in the theoretical estimation errors $O\left[T^{1.5} + T\sqrt{\ln{\left(T\right)}}+T\right]$ derived from Eqs.~\eqref{eq:3.19} and \eqref{eq:3.21}. These results confirm that the increase in the estimation error of the neural oscillator with respect to the time length $T$ is moderate.

In addition to the relative generalization error $\tilde{\varepsilon}_{E,2}$ in Eq.~\eqref{eq:4.7}, the averaged squared Wasserstein-1 distance over the time interval $[0, T]$ between the probability measures $\rho_{E_{X_5}(t_i)}$ and $\rho_{\tilde{E}_{X_5}(t_i)}$ of $E_{X_5}(t_i)$ and $\tilde{E}_{X_5}(t_i)$, respectively, is also evaluated. At each time point $t_i$, the squared Wasserstein-1 distance $W_1^{2}\!\left[\rho_{E_{X_5}(t_i)},\rho_{\tilde{E}_{X_5}(t_i)}\right]$ is computed using the empirical cumulative distribution functions (CDFs) of the $5\times 10^4$ samples $E_{X_5,l}(t_i)$ and $\tilde{E}_{X_5,l}(t_i)$ via the sorted-sample formula in Eq.~\eqref{eq:4.5}. The averaged squared Wasserstein-1 distance over $[0, T]$ is then computed as
\begin{equation}
    \bar{W}_{1,E}^{2} = \frac{1}{100T}\sum_{i=0}^{100T-1} W_1^{2}\left[\rho_{E_{X_5}(t_i)},\rho_{\tilde{E}_{X_5}(t_i)}\right].
    \label{eq:4.8}
\end{equation}
The variation of $\bar{W}_{1,E}^{2}$ with respect to the increasing time length $T$ is shown in Figure~\ref{fig:4}, where the markers represent the average values across the convergent sets, and the upper and lower bars indicate the largest and smallest values, respectively. Similar to the trend observed in Figure~\ref{fig:3}, the variation of $\bar{W}_{1,E}^{2}$ with respect to $T$ can be approximated by a power-law function with an exponent of $1.5$, albeit with relatively large upper and lower deviations. This result is consistent with the leading term $O\!\left[T^{1.5} + T\sqrt{\ln{T}}+T\right]$ in the theoretical estimation error predicted by Corollary 2, thereby further confirming that the growth of the estimation error of the neural oscillator with respect to the time length $T$ is moderate and validating the established generalization bound for the squared Wasserstein-1 distance.

\begin{figure}[t]
    \centering
    \begin{minipage}[t]{0.48\linewidth}
        \centering
        \includegraphics[width=\linewidth]{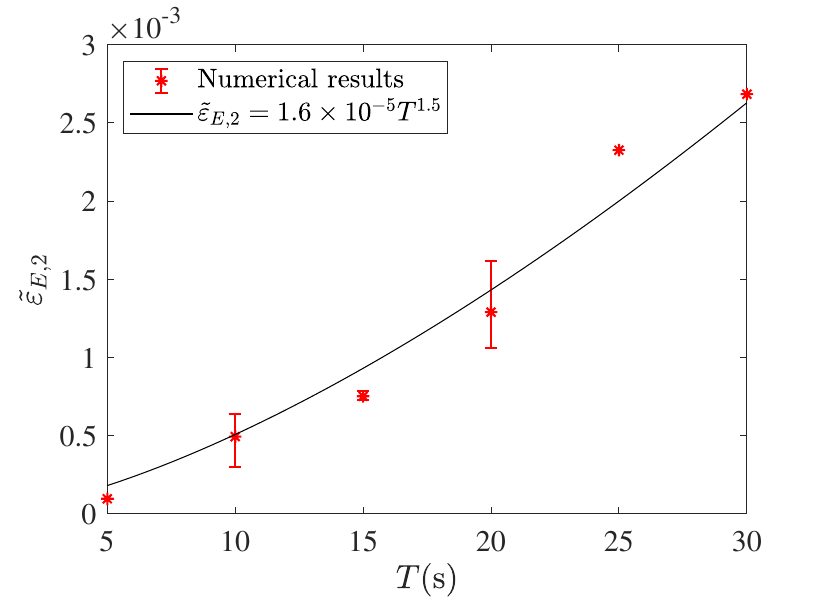}
        \caption{{\color{black} Relative generalization error $\tilde{\varepsilon}_{E,2}$ between $E_{X_5,l}(t_i)$ and $\tilde{E}_{X_5,l}(t_i)$ versus the time length $T$.}}
        \label{fig:3}
    \end{minipage}
    \hfill
    \begin{minipage}[t]{0.48\linewidth}
        \centering
        \includegraphics[width=\linewidth]{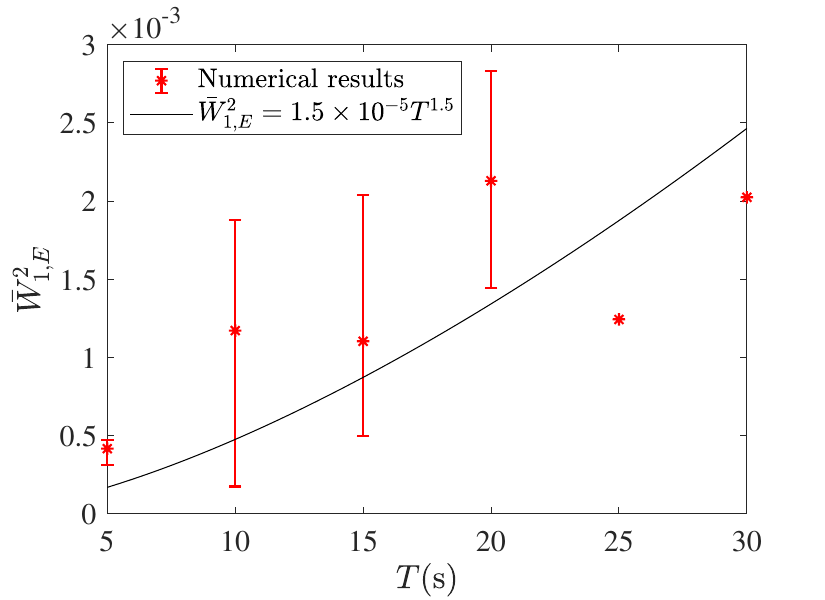}
        \caption{{\color{black} Averaged squared Wasserstein-1 distance $\bar{W}_{1,E}^{2}$ between the probability measures $\rho_{E_{X_5}(t_i)}$ of $E_{X_5,l}(t_i)$ and $\rho_{\tilde{E}_{X_5}(t_i)}$ of $\tilde{E}_{X_5,l}(t_i)$ versus the time length $T$.}}
        \label{fig:4}
    \end{minipage}
\end{figure}

The ODE of the extreme-value process $E_{X_5}(t)$ can be expressed as
\begin{equation}
    E_{X_5}'(t)=\left\{
    \begin{aligned}
        &\left|X_5'(t)\right|,\;\;\; \left|X_5(t)\right|\geq E_{X_5}(t)  \\
        &0,\;\;\;\;\;\;\;\;\;\;\;\;\mathrm{otherwise}
    \end{aligned}
     \right..
    \label{eq:4.9}
\end{equation}
By incorporating Eq.~\eqref{eq:4.9} into Eq.~\eqref{eq:4.1}, it can be shown that the mapping from $U_{\mathrm{e}}(t)$ to $E_{X_5}(t)$ is continuous but not smooth due to the piecewise nature of the derivative term in Eq.~\eqref{eq:4.9}. This nonsmoothness renders the learning of this mapping more difficult than that of the smooth one from $U_{\mathrm{e}}(t)$ to $X_{5}(t)$. {\color{black} To assess the ability of the neural oscillator to approximate this nonsmooth mapping, both qualitative and quantitative convergence analyses are conducted under five sample sizes $N \in \left\{100, 200, 400, 800, 1600\right\}$ and $T=30$ seconds. For each $N$, a neural oscillator is trained, and its $5\times 10^4$ prediction samples are used to estimate the CDF of $\tilde{E}_{X_5}(30)$, which is then compared with the CDF estimated from the corresponding $5\times 10^4$ samples of $E_{X_5}(30)$ obtained by numerically solving the ODE in Eq.~\eqref{eq:4.1}. As shown in Figure~\ref{fig:5}, the CDFs predicted by the trained neural oscillators progressively approach the CDF estimated from the simulated samples as $N$ increases, providing qualitative evidence that the neural oscillator effectively approximates the nonsmooth mapping from $U_{\mathrm{e}}(t)$ to $E_{X_5}(t)$. To further quantify this convergence behavior, the squared Wasserstein-1 distance $W_{1,E_{X_5}(30)}^{2}$ between the probability measures $\rho_{E_{X_5,l}(30)}$ of $E_{X_5,l}(30)$ and $\rho_{\tilde{E}_{X_5,l}(30)}$ of $\tilde{E}_{X_5,l}(30)$ is computed using the empirical CDFs via the sorted-sample formula in Eq.~\eqref{eq:4.5}, and its variation with respect to $N$ is shown in Figure~\ref{fig:6}. The numerical convergence rate of $W_{1,E_{X_5}(30)}^{2}$ is approximately $-1$, which is steeper than the theoretical convergence rate of $-0.5$ predicted by Corollaries 1 and 2. Since the established theoretical results in Corollaries 1 and 2 provide upper generalization bounds, the steeper empirical rate observed in Figure~\ref{fig:6} does not violate these bounds and instead reinforces the qualitative convergence in Figure~\ref{fig:5}.}
\begin{figure}[t]
    \centering
    \begin{minipage}[t]{0.48\linewidth}
        \centering
        \includegraphics[width=\linewidth]{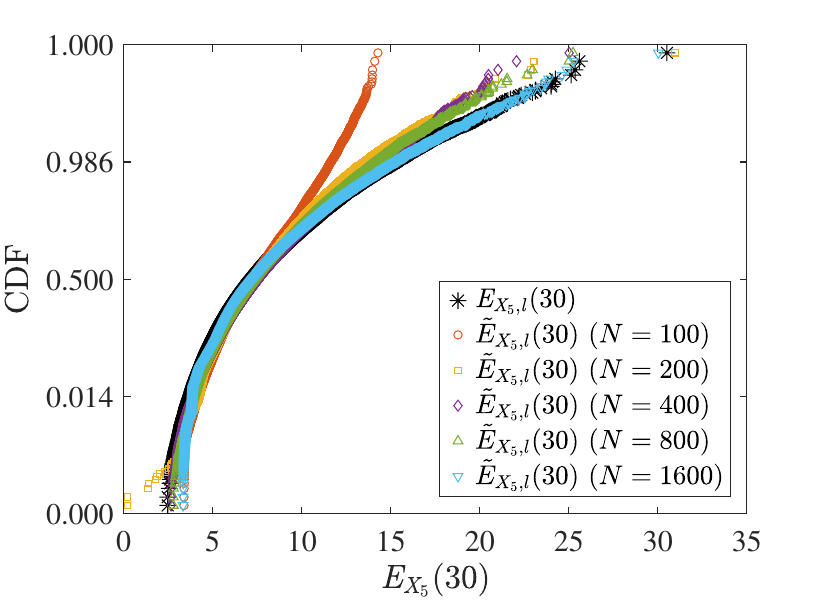}
        \caption{{\color{black} CDFs of $E_{X_5}(30)$ and $\tilde{E}_{X_5}(30)$ from the neural oscillators trained under different sample numbers $N$.}}
        \label{fig:5}
    \end{minipage}
    \hfill
    \begin{minipage}[t]{0.48\linewidth}
        \centering
        \includegraphics[width=\linewidth]{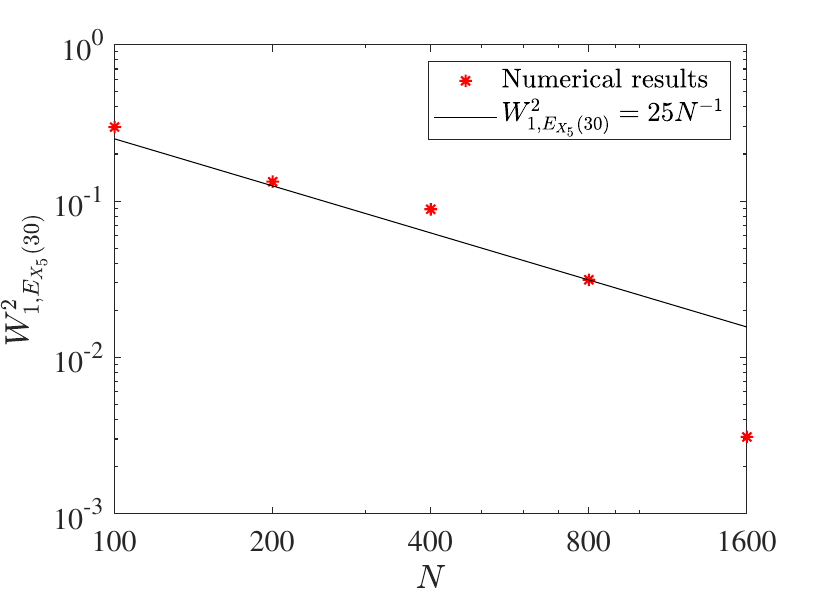}
        \caption{{\color{black} Squared Wasserstein-1 distance $W_{1,E_{X_5}(30)}^{2}$ between the probability measures $\rho_{E_{X_5,l}(30)}$ of $E_{X_5,l}(30)$ and $\rho_{\tilde{E}_{X_5,l}(30)}$ of $\tilde{E}_{X_5,l}(30)$ versus the sample number $N$.}}
        \label{fig:6}
    \end{minipage}
\end{figure}}

\section{Conclusions}
In this study, two upper PAC generalization bounds of the neural oscillator consisting of a second-order ODE followed by an MLP are derived for approximating the causal and uniformly continuous operators between continuous temporal function spaces and for approximating the uniformly asymptotically incrementally stable second-order dynamical systems. {\color{black} These generalization bounds are further extended to the squared Wasserstein-1 distance between the probability measures of Lipschitz continuous quantities of interest calculated from target causal operators and the corresponding learned neural oscillators.} The theoretical results demonstrate that the estimation errors in the generalization bounds grow polynomially with increasing MLP sizes and time length, thereby avoiding the curse of parametric complexity. Moreover, the derived generalization bounds reveal that the generalization capability of the neural oscillator can be enhanced by constraining the Lipschitz constants of the employed MLPs through appropriate regularization in the loss function. {\color{black} Numerical studies based on a Bouc-Wen nonlinear system subjected to stochastic seismic excitation validate} the power laws of the estimation errors {\color{black} in Theorems 1 and 2, and Corollaries 1 and 2,} with respect to the sample size and time length, and confirms the effectiveness of constraining the MLPs' matrix and vector norms in improving the generalization performance of the neural oscillator under limited training data.

{\color{black} The generalization bounds established in Theorems 1 and 2 are existence results. They show that there exist weight configurations of the MLPs $\itGamma(\cdot)$ and $\itPi(\cdot)$ such that the neural oscillator in Eq.~\eqref{eq:2.1} can approximate the target causal operators or dynamical systems within the predicted error bounds when trained with a finite number of samples. However, these theoretical results do not address the training problem, in particular, which optimization algorithms can efficiently identify models that achieve these bounds. In the numerical experiments of this study, some training runs for learning the mapping from $U_{\mathrm{e}}(t)$ to $E_{X_{5}}(t)$ do not converge. This gap between theory and practice is a frontier challenge in neural network research. Developing specialized neural oscillator architectures, together with efficient training strategies, that enable trained models to realize the theoretically predicted error bounds is a promising direction for future research.}

\section*{CRediT Authorship Contribution Statement}
{\bf Zifeng Huang:} Conceptualization, Methodology, Software, Writing-reviewing \& editing, Project administration. {\bf Konstantin M. Zuev:} Methodology, Writing-reviewing \& editing. {\bf Yong Xia:} Writing-reviewing \& editing, Supervision. {\bf Michael Beer:} Supervision, Project administration.

\section*{Declaration of Competing Interest}
The authors declare that they have no known competing financial interests or personal relationships that could have appeared to influence the work reported in this paper.

\section*{Data and Code Availability}
All code used in this study is available at: \url{https://github.com/ZifengH22/Upper-Generalization-Bounds-for-Neural-Oscillators}. All data presented in this study were generated using this code.

\section*{Acknowledgments}
This work is supported by the German Research Foundation (Project No. 554325683 and HU 3418/2-1). The opinions and conclusions presented in this paper are entirely those of the authors.

\appendix
\section{Proofs for Lemmas and Theorems in Section \ref{sec3}}
\label{appA}

\subsection{Proof of Lemma 3}
\label{appA.1}
{\bf Proof:} The empirical Rademacher complexity $\Re_{\mathcal{G}_{\vartheta}}$ in Eq.~\eqref{eq:3.5} can be calculated as
\begin{equation}
    {\Re _{{{\cal G}_\vartheta }}} = {\mathbb{E}_{\boldsymbol{\upsigma }}}\mathop {\sup }\limits_{{\it{\Pi}} \circ {\Phi_{\it{\Gamma}} } \in {{\cal F}_{{\it{\Pi}} \circ {\Phi_{\it{\Gamma}} }}}} \frac{1}{N}\sum\limits_{i = 1}^N {{\sigma _i}\sum\limits_{j = 1}^{{w_{{\it{\Pi}} ,{\text{out}}}}} {\int_0^T {{{\left\{ {{v_{ij}}(t) - {\it{\Pi}}\circ {\Phi_{\it{\Gamma}} }{{\left[ {{{\bf{u}}_i}(\tau )} \right]}_j}(t)} \right\}}^2}{\text{d}}t} } },
    \label{eq:a.1}
\end{equation}
where $v_{ij}(t) = \Phi\left[\mathbf{u}_i(\tau) \right]_j\left(t\right)$ and $\PiPhiGamma\left[\mathbf{u}_i(\tau) \right]_j(t)$ are the $j^\text{th}$ elements of the $q$-dimensional vector functions $\Phi\left[\mathbf{u}_i(\tau) \right]\left(t\right)$ and $\PiPhiGamma\left[\mathbf{u}_i(\tau) \right](t)$, respectively, and $q = \wPiout$. Since $\PiPhiGamma\left[\mathbf{u}_i(\tau) \right]_j(t)$ and $v_{ij}(t)$ are bounded and continuous functions over $[0,T]$, they can be respectively expressed as
\begin{equation}
    {\it{\Pi}}\circ{\Phi_{\it{\Gamma}} }{\left[ {{{\bf{u}}_i}(\tau )} \right]_j}(t) = \sum\limits_{k = 1}^{ + \infty } {{\alpha _{jk}}\left\{ {{\it{\Pi}}\circ{\Phi_{\it{\Gamma}} }{{\left[ {{{\bf{u}}_i}(\tau )} \right]}_j}} \right\}{e_k}(t)}
    \label{eq:a.2}
\end{equation}
and
\begin{equation}
    {v_{ij}}(t) = \sum\limits_{k = 1}^{ + \infty } {{\beta _{jk}}({v_{ij}}){e_k}(t)},
    \label{eq:a.3}
\end{equation}
where $e_k(t)$, $k = 1, 2,...$, denote a countable set of standard orthogonal basis functions over $[0,T]$, $\alpha_{jk}\left\{\PiPhiGamma\left[\mathbf{u}_i(\tau)\right]_j\right\}$ and ${\beta _{jk}}({v_{ij}})$ are the coefficients of $\PiPhiGamma\left[\mathbf{u}_i(\tau)\right]_j(t)$ and $v_{ij}(t)$, respectively.

Substituting Eqs.~\eqref{eq:a.2} and \eqref{eq:a.3} in Eq.~\eqref{eq:a.1} and utilizing the Parseval’s identity, $\Re_{\mathcal{G}_{\vartheta}}$ can be expressed as
\begin{equation}
    \begin{aligned}
        &\Re_{\mathcal{G}_\vartheta} = {\mathbb{E}_{\boldsymbol{\upsigma }}}\mathop {\sup }\limits_{{\it{\Pi}} \circ {\Phi_{\it{\Gamma}} } \in {{\cal F}_{{\it{\Pi}} \circ {\Phi_{\it{\Gamma}} }}}} \frac{1}{N}\sum\limits_{i = 1}^N {{\sigma _i}\sum\limits_{j = 1}^{{w_{{\it{\Pi}} ,{\rm{out}}}}} {\int_0^T {{{\left[ {\sum\limits_{k = 1}^{ + \infty } {\left( {{\beta _{jk}}({v_{ij}}) - {\alpha _{jk}}\left\{ {{\it{\Pi}} \circ {\Phi_{\it{\Gamma}} }{{\left[ {{{\bf{u}}_i}(\tau )} \right]}_j}} \right\}} \right){e_k}(t)} } \right]}^2}{\rm{d}}t} } } \\
        &\;\;\;\;\;\;\;\; = {\mathbb{E}_{\boldsymbol{\upsigma }}}\mathop {\sup }\limits_{{\it{\Pi}} \circ {\Phi_{\it{\Gamma}} } \in {{\cal F}_{{\it{\Pi}} \circ {\Phi_{\it{\Gamma}} }}}} \frac{1}{N}\sum\limits_{i = 1}^N {{\sigma _i}z\left\{ {{\it{\Pi}} \circ {\Phi_{\it{\Gamma}} }\left[ {{{\bf{u}}_i}(\tau )} \right]} \right\}},
    \end{aligned}
    \label{eq:a.4}
\end{equation}
where
\begin{equation}
    z\left\{ {{\it{\Pi}}\circ {\Phi_{\it{\Gamma}} }\left[ {{{\bf{u}}_i}(\tau )} \right]} \right\} = \sum\limits_{j = 1}^{{w_{{\it{\Pi}} ,{\rm{out}}}}} {\sum\limits_{k = 1}^{ + \infty } {{{\left( {{\beta _{jk}}({v_{ij}}) - {\alpha _{jk}}\left\{ {{\it{\Pi}}\circ {\Phi_{\it{\Gamma}} }{{\left[ {{{\bf{u}}_i}(\tau )} \right]}_j}} \right\}} \right)}^2}} }.
    \label{eq:a.5}
\end{equation}

Consider two neural oscillators $\PiPhiGammaone$ and $\PiPhiGammatwo$ $\in \FPiPhiGamma$, the difference between $z\left\{\PiPhiGammaone\left[\mathbf{u}_i(\tau)\right]_j(t)  \right\}$ and $z\left\{\PiPhiGammatwo\left[\mathbf{u}_i(\tau)\right]_j(t)  \right\}$ can be bounded by
\begin{equation}
    \begin{aligned}
    &\left|z\left\{ {{{\it{\Pi}} _1} \circ {\Phi _{{{\it{\Gamma}} _1}}}{{\left[ {{{\bf{u}}_i}(\tau )} \right]}_j}} \right\} - z\left\{ {{{\it{\Pi}} _2} \circ {\Phi _{{{\it{\Gamma}} _2}}}{{\left[ {{{\bf{u}}_i}(\tau )} \right]}_j}} \right\}\right|\\
    &\;\;\; = \left|\sum\limits_{j = 1}^{{w_{{\it{\Pi}} ,{\rm{out}}}}} {\sum\limits_{k = 1}^{ + \infty } {{{\left( {{\beta _{jk}}({v_{ij}}) - {\alpha _{jk}}\left\{ {{{\it{\Pi}} _1} \circ {\Phi _{{{\it{\Gamma}} _1}}}{{\left[ {{{\bf{u}}_i}(\tau )} \right]}_j}} \right\}} \right)}^2}} }- \sum\limits_{j = 1}^{{w_{{\it{\Pi}} ,{\rm{out}}}}} {\sum\limits_{k = 1}^{ + \infty } {{{\left( {{\beta _{jk}}({v_{ij}}) - {\alpha _{jk}}\left\{ {{{\it{\Pi}} _2} \circ {\Phi _{{{\it{\Gamma}} _2}}}{{\left[ {{{\bf{u}}_i}(\tau )} \right]}_j}} \right\}} \right)}^2}} }\right|\\
    &\;\;\; = \left|\sum\limits_{j = 1}^{{w_{{\it{\Pi}} ,{\rm{out}}}}} {\sum\limits_{k = 1}^{ + \infty } {\left( {\alpha _{jk}^2\left\{ {{{\it{\Pi}} _1} \circ {\Phi _{{{\it{\Gamma}} _1}}}{{\left[ {{{\bf{u}}_i}(\tau )} \right]}_j}} \right\} - \alpha _{jk}^2\left\{ {{{\it{\Pi}} _2} \circ {\Phi _{{{\it{\Gamma}} _2}}}{{\left[ {{{\bf{u}}_i}(\tau )} \right]}_j}} \right\}} \right)} }\right. \\
    &\;\;\;\;\;\;\;\;\;\left.- 2\sum\limits_{j = 1}^{{w_{{\it{\Pi}} ,{\rm{out}}}}} {\sum\limits_{k = 1}^{ + \infty } {{\beta _{jk}}({v_{ij}})\left( {{\alpha _{jk}}\left\{ {{{\it{\Pi}} _1} \circ {\Phi _{{{\it{\Gamma}} _1}}}{{\left[ {{{\bf{u}}_i}(\tau )} \right]}_j}} \right\} - {\alpha _{jk}}\left\{ {{{\it{\Pi}} _2} \circ {\Phi _{{{\it{\Gamma}} _2}}}{{\left[ {{{\bf{u}}_i}(\tau )} \right]}_j}} \right\}} \right)} }\right|\\
    &\;\;\; = \left|\sum\limits_{j = 1}^{{w_{{\it{\Pi}} ,{\rm{out}}}}} {\sum\limits_{k = 1}^{ + \infty } {\left( {{\alpha _{jk}}\left\{ {{{\it{\Pi}} _1} \circ {\Phi _{{{\it{\Gamma}} _1}}}{{\left[ {{{\bf{u}}_i}(\tau )} \right]}_j}} \right\} - {\alpha _{jk}}\left\{ {{{\it{\Pi}} _2} \circ {\Phi _{{{\it{\Gamma}} _2}}}{{\left[ {{{\bf{u}}_i}(\tau )} \right]}_j}} \right\}} \right)} }\right. \\
    &\;\;\;\;\;\;\;\;\;\;\;\;\;\;\;\;\;\;\;\;\;\;\;\left. \times \left( {{\alpha _{jk}}\left\{ {{{\it{\Pi}} _1} \circ {\Phi _{{{\it{\Gamma}} _1}}}{{\left[ {{{\bf{u}}_i}(\tau )} \right]}_j}} \right\} + {\alpha _{jk}}\left\{ {{{\it{\Pi}} _2} \circ {\Phi _{{{\it{\Gamma}} _2}}}{{\left[ {{{\bf{u}}_i}(\tau )} \right]}_j}} \right\} - 2{\beta _{jk}}({v_{ij}})} \right)\right|\\
    &\;\;\;\mathop  \le \limits^{(a)} \sqrt {\sum\limits_{j = 1}^{{w_{{\it{\Pi}} ,{\rm{out}}}}} {\sum\limits_{k = 1}^{ + \infty } {{{\left( {{\alpha _{jk}}\left\{ {{{\it{\Pi}} _1} \circ {\Phi _{{{\it{\Gamma}} _1}}}{{\left[ {{{\bf{u}}_i}(\tau )} \right]}_j}} \right\} - {\alpha _{jk}}\left\{ {{{\it{\Pi}} _2} \circ {\Phi _{{{\it{\Gamma}} _2}}}{{\left[ {{{\bf{u}}_i}(\tau )} \right]}_j}} \right\}} \right)}^2}} } }\\
    &\;\;\;\;\;\;\;\; \times \sqrt {\sum\limits_{j = 1}^{{w_{{\it{\Pi}} ,{\rm{out}}}}} {\sum\limits_{k = 1}^{ + \infty } {{{\left( {{\alpha _{jk}}\left\{ {{{\it{\Pi}} _1} \circ {\Phi _{{{\it{\Gamma}} _1}}}{{\left[ {{{\bf{u}}_i}(\tau )} \right]}_j}} \right\} + {\alpha _{jk}}\left\{ {{{\it{\Pi}} _2} \circ {\Phi _{{{\it{\Gamma}} _2}}}{{\left[ {{{\bf{u}}_i}(\tau )} \right]}_j}} \right\} - 2{\beta _{jk}}({v_{ij}})} \right)}^2}} } } \\
    &\;\;\; = \sqrt {\sum\limits_{j = 1}^{{w_{{\it{\Pi}} ,{\rm{out}}}}} {\sum\limits_{k = 1}^{ + \infty } {{{\left( {{\alpha _{jk}}\left\{ {{{\it{\Pi}} _1} \circ {\Phi _{{{\it{\Gamma}} _1}}}{{\left[ {{{\bf{u}}_i}(\tau )} \right]}_j}} \right\} - {\alpha _{jk}}\left\{ {{{\it{\Pi}} _2} \circ {\Phi _{{{\it{\Gamma}} _2}}}{{\left[ {{{\bf{u}}_i}(\tau )} \right]}_j}} \right\}} \right)}^2}} } }\\
    &\;\;\;\;\;\;\;\; \times \sqrt {\sum\limits_{j = 1}^{{w_{{\it{\Pi}} ,{\rm{out}}}}} {\int_0^T {{{\left\{ {{{\it{\Pi}} _1} \circ {\Phi _{{{\it{\Gamma}} _1}}}{{\left[ {{{\bf{u}}_i}(\tau )} \right]}_j}(t) + {{\it{\Pi}} _2} \circ {\Phi _{{{\it{\Gamma}} _2}}}{{\left[ {{{\bf{u}}_i}(\tau )} \right]}_j}(t) - 2{v_{ij}}(t)} \right\}}^2}{\rm{d}}t} } }\\
    &\;\;\;\mathop  \le \limits^{(b)} 2\sqrt {T{w_{{\it{\Pi}} ,{\rm{out}}}}} {B_{{\rm{loss}}}}\sqrt {\sum\limits_{j = 1}^{{w_{{\it{\Pi}} ,{\rm{out}}}}} {\sum\limits_{k = 1}^{ + \infty } {{{\left( {{\alpha _{jk}}\left\{ {{{\it{\Pi}} _1} \circ {\Phi _{{{\it{\Gamma}} _1}}}{{\left[ {{{\bf{u}}_i}(\tau )} \right]}_j}} \right\} - {\alpha _{jk}}\left\{ {{{\it{\Pi}} _2} \circ {\Phi _{{{\it{\Gamma}} _2}}}{{\left[ {{{\bf{u}}_i}(\tau )} \right]}_j}} \right\}} \right)}^2}} } },
    \end{aligned}
    \label{eq:a.6}
\end{equation}
where Step (\textit{a}) follows from the Cauchy-Schwarz inequality and Step (\textit{b}) relies on the Assumption 11 that $\left|\PiPhiGammaone\left[\mathbf{u}_i(\tau)\right]_j + \PiPhiGammatwo\left[\mathbf{u}_i(\tau)\right]_j - 2v_{ij}(t)\right|$ is bounded by $2\Bloss$. Eq.~\eqref{eq:a.6} indicates that $2\sqrt{T\wPiout}\Bloss$ is a Lipschitz constant of $z\left\{\PiPhiGamma\left[\mathbf{u}_i(\tau)\right]_j\right\}$ with respect to ${\alpha}_{jk}\left\{\PiPhiGamma\left[\mathbf{u}_i(\tau)\right]_j\right\}$. Then, from Corollary 4 of \citet{maurer2016vector} and Eq.~\eqref{eq:a.4}, $\Re_{\mathcal{G}_{\vartheta}}$ can be bounded by
\begin{equation}
    {\Re_{{\mathcal{G}_\vartheta }}} \le {\mathbb{E}_{{\boldsymbol{\tilde{\upsigma} }}}}\mathop {\sup }\limits_{{\it{\Pi}}  \circ {\Phi _{\it{\Gamma}} } \in {{\cal F}_{{\it{\Pi}}  \circ {\Phi _{\it{\Gamma}} }}}} \frac{{2\sqrt {2T{w_{{\it{\Pi}} ,{\rm{out}}}}} {B_{{\rm{loss}}}}}}{N}\sum\limits_{i = 1}^N {\sum\limits_{j = 1}^{\wPiout} {\sum\limits_{k = 1}^{ + \infty } {{{\tilde \sigma }_{ijk}}{\alpha _{jk}}\left\{ {{\it{\Pi}}  \circ {\Phi_{\it{\Gamma}} }{{\left[ {{{\bf{u}}_i}(\tau )} \right]}_j}} \right\}} } }.
    \label{eq:a.7}
\end{equation}
This completes the proof. 
\hfill$\blacksquare$

\subsection{Proof of Lemma 4}
\label{appA.2}
{\bf Proof:} Consider a random variable sequence $X_K$, $K = 1, 2,...$
\begin{equation}
    {X_K} = \sum\limits_{i = 1}^N {\sum\limits_{j = 1}^{{w_{{\itPi} ,{\rm{out}}}}} {\sum\limits_{k = 1}^K {{{\tilde \sigma }_{ijk}}{\alpha _{ijk}}} } },
    \label{eq:a.8}
\end{equation}
where ${\alpha _{ijk}} = {\alpha _{jk}}\left\{ {\itPi  \circ {\Phi _\itGamma }{{\left[ {{{\bf{u}}_i}(\tau )} \right]}_j}} \right\}$ for $\forall\PiPhiGamma \in \FPiPhiGamma$, this sequence satisfies
\begin{equation}
    \begin{aligned}
        &\mathop {\lim }\limits_{{K_1},{K_2} \to  + \infty } {\mathbb{E}_{{\tilde{\boldsymbol{\upsigma} }}}}{\left( {{X_{{K_2}}} - {X_{{K_1}}}} \right)^2} = \mathop {\lim }\limits_{{K_1},{K_2} \to  + \infty } {\mathbb{E}_{{\tilde{\boldsymbol{\upsigma} }}}}{\left( {\sum\limits_{i = 1}^N {\sum\limits_{j = 1}^{{w_{\itPi ,{\rm{out}}}}} {\sum\limits_{k = 1}^{{K_2}} {{{\tilde \sigma }_{ijk}}{\alpha _{ijk}}} } }  - \sum\limits_{i = 1}^N {\sum\limits_{j = 1}^{{w_{\itPi ,{\rm{out}}}}} {\sum\limits_{k = 1}^{{K_1}} {{{\tilde \sigma }_{ijk}}{\alpha _{ijk}}} } } } \right)^2}\\
        &\;\;\;\;\;\;\;\;\;\;\;\;\;\;\;\;\;\;\;\;\;\;\;\;\;\;\;\;\;\;\;\;\;\;\;\;\;\mathop  = \limits^{(a)} \mathop {\lim }\limits_{{K_{\min }},{K_{\max }} \to  + \infty } {{\sum\limits_{i = 1}^N {\sum\limits_{j = 1}^{{w_{\itPi ,{\rm{out}}}}} {\sum\limits_{k = {K_{\min }} + 1}^{{K_{\max }}} {\alpha^2_{ijk}} } } }}\mathop  = \limits^{(b)} 0,
    \end{aligned}
    \label{eq:a.9}
\end{equation}
where in Step (\textit{a}) $K_{\text{min}} = \min\left(K_1,K_2\right)$ and $K_{\text{max}} = \max\left(K_1,K_2\right)$, and Step (\textit{b}) is valid since the sequence $\alpha_{ijk}$ is square-summable. Eq.~\eqref{eq:a.9} indicates that the sequence $X_K$ converges in mean-square. The expectation of ${\kappa}\left(\PiPhiGamma, \tilde{\boldsymbol{\upsigma}}\right)$ is 
\begin{equation}
    {\mathbb{E}_{{\tilde{\boldsymbol{\upsigma}}}}}\kappa \left( {\itPi  \circ {\Phi _\itGamma },{\tilde{\boldsymbol{\upsigma}}}} \right) = {\mathbb{E}_{{\tilde{\boldsymbol{\upsigma}}}}}\sum\limits_{i = 1}^N {\sum\limits_{j = 1}^{{w_{\itPi ,{\rm{out}}}}} {\sum\limits_{k = 1}^{ + \infty } {{{\tilde \sigma }_{ijk}}{\alpha _{ijk}}} } } \mathop  = \limits^{(a)} \sum\limits_{i = 1}^N {\sum\limits_{j = 1}^{{w_{\itPi ,{\rm{out}}}}} {\sum\limits_{k = 1}^{ + \infty } {{\mathbb{E}_{{\tilde{\boldsymbol{\upsigma}}}}}{{\tilde \sigma }_{ijk}}{\alpha _{ijk}}} } } = 0,
    \label{eq:a.10}
\end{equation}
where the mean-square convergence property of $X_K$ allows the expectation and limit to be interchanged in Step (a) \citep{huang2022probability}. 

For arbitrary $\PiPhiGammaone,\PiPhiGammatwo \in \FPiPhiGamma$, $\PiPhiGammaone\left[\mathbf{u}_i(\tau)\right]_j(t) - \PiPhiGammatwo\left[\mathbf{u}_i(\tau)\right]_j(t)$ are bounded and continuous functions over $[0,T]$ for all $i = 1, 2,..., N$ and $j = 1, 2,..., \wPiout$. Thus, $\Delta \alpha _{ijk} = \alpha _{ijk,1} - \alpha _{ijk,2}$ is also a square-summable sequence, where $\alpha _{ijk,1} = \alpha_{jk}$ $\left\{ \PiPhiGammaone\left[\mathbf{u}_i(\tau)\right]_j\right\}$ and $\alpha _{ijk,2} = \alpha_{jk}\left\{\PiPhiGammatwo\left[\mathbf{u}_i(\tau)\right]_j\right\}$. ${X_{K,\Delta}} = \sum_{i = 1}^N {\sum_{j = 1}^{{w_{{\itPi} ,{\rm{out}}}}} {\sum_{k = 1}^K{{{\tilde \sigma }_{ijk}}{\Delta \alpha _{ijk}}} } }$ is a sequence converging in mean-square and satisfies
\begin{equation}
    {\mathbb{E}_{{\tilde{\boldsymbol \upsigma }}}}{\left( {\mathop {\lim }\limits_{K \to  + \infty } {X_{K,\Delta }}} \right)^2} = \mathop {\lim }\limits_{K \to  + \infty } {\mathbb{E}_{{\tilde{\boldsymbol \upsigma }}}}X_{K,\Delta }^2.
    \label{eq:a.11}
\end{equation}

The pseudo-metric $d_{\kappa}\left(\PiPhiGammaone,\PiPhiGammatwo\right)$ for arbitrary $\PiPhiGammaone,\PiPhiGammatwo \in \FPiPhiGamma$ is calculated as
\begin{equation}
    \begin{aligned}
        &{d_\kappa }\left( {{\itPi _1} \circ {\Phi _{{\itGamma _1}}},{\itPi _2} \circ {\Phi _{{\itGamma _2}}}} \right)\\
        &\;\;\;\;\; = \sqrt {{\mathbb{E}_{{\tilde{\boldsymbol{\upsigma}}}}}{{\left[ {\kappa \left( {{\itPi _1} \circ {\Phi _{{\itGamma _1}}},{\tilde{\boldsymbol{\upsigma}}}} \right) - \kappa \left( {{\itPi _2} \circ {\Phi _{{\itGamma _2}}},{\tilde{\boldsymbol{\upsigma}}}} \right)} \right]}^2}} = \sqrt {{\mathbb{E}_{{\tilde{\boldsymbol{\upsigma}}}}}{{\left[ {\sum\limits_{i = 1}^N {\sum\limits_{j = 1}^{{w_{\itPi ,{\rm{out}}}}} {\sum\limits_{k = 1}^{ + \infty } {{{\tilde \sigma }_{ijk}}\Delta \alpha_{ijk} } } } } \right]}^2}} \\
        &\;\;\;\; \mathop  = \limits^{(a)} \sqrt {\sum\limits_{i = 1}^N {\sum\limits_{j = 1}^{{w_{\itPi ,{\rm{out}}}}} {\sum\limits_{k = 1}^{ + \infty } {{\mathbb{E}_{{\tilde{\boldsymbol{\upsigma}}}}}\tilde \sigma _{ijk}^2{\Delta \alpha_{ijk}^2}} } } } = \sqrt {\sum\limits_{i = 1}^N {\sum\limits_{j = 1}^{{w_{\itPi ,{\rm{out}}}}} {\sum\limits_{k = 1}^{ + \infty } {{\Delta \alpha_{ijk}^2}} } } } \\ 
        &\;\;\;\;\; = \sqrt {\sum\limits_{i = 1}^N {\sum\limits_{j = 1}^{{w_{\itPi ,{\rm{out}}}}} {\int_0^T {{{\left\{ {{\itPi _1} \circ {\Phi _{{\itGamma _1}}}{{\left[ {{{\bf{u}}_i}(\tau )} \right]}_j}(t) - {\itPi _2} \circ {\Phi _{{\itGamma _2}}}{{\left[ {{{\bf{u}}_i}(\tau )} \right]}_j}(t)} \right\}}^2}{\rm{d}}t} } } } \\
        &\;\;\;\;\; = \sqrt {\sum\limits_{i = 1}^N {\left\| {{\itPi _1} \circ {\Phi _{{\itGamma _1}}}\left[ {{{\bf{u}}_i}(\tau )} \right](t) - {\itPi _2} \circ {\Phi _{{\itGamma _2}}}\left[ {{{\bf{u}}_i}(\tau )} \right](t)} \right\|_{{L^2}}^2} } ,
    \end{aligned}
    \label{eq:a.12}
\end{equation}
where Step (\textit{a}) holds from Eq.~\eqref{eq:a.11}. Given $\forall w \in \mathbb{R}$, it can be obtained
\begin{equation}
    \begin{aligned}
        &{\mathbb{E}_{{\tilde{\boldsymbol{\upsigma}}}}}\exp \left\{ {w\left[ {\kappa \left( {{\itPi _1} \circ {\Phi _{{\itGamma _1}}},{\tilde{\boldsymbol{\upsigma}}}} \right) - \kappa \left( {{\itPi _2} \circ {\Phi _{{\itGamma _2}}},{\tilde{\boldsymbol{\upsigma}}}} \right)} \right]} \right\}\\   
        &\;\;\;\;\; = {\mathbb{E}_{{\tilde{\boldsymbol{\upsigma}}}}}\exp \left( {w\sum\limits_{i = 1}^N {\sum\limits_{j = 1}^{\wPiout} {\sum\limits_{k = 1}^{ + \infty } {{{\tilde \sigma }_{ijk}}\Delta \alpha_{ijk}} } } } \right)\mathop  = \limits^{(a)} \prod\limits_{i = 1}^N {\prod\limits_{j = 1}^{{w_{\itPi ,{\rm{out}}}}} {\prod\limits_{k = 1}^{ + \infty } {{E_{{\tilde \sigma }_{ijk}}}\exp } } } \left( {{{\tilde \sigma }_{ijk}}w\Delta \alpha_{ijk}} \right)\\
        &\;\;\;\;\;\mathop  \le \limits^{(b)} \prod\limits_{i = 1}^N {\prod\limits_{j = 1}^{{w_{\itPi ,{\rm{out}}}}} {\prod\limits_{k = 1}^{ + \infty } {\exp \left({0.5{w^2}{\Delta \alpha_{ijk}^2}} \right)} } } = \exp \left( {0.5{w^2}\sum\limits_{i = 1}^N {\sum\limits_{j = 1}^{{w_{\itPi ,{\rm{out}}}}} {\sum\limits_{k = 1}^{ + \infty } {{\Delta \alpha_{ijk}^2}} } } } \right)\\
         &\;\;\;\;\;\mathop  = \limits^{(c)} \exp \left[ {0.5{w^2}d_\kappa ^2\left( {{\itPi _1} \circ {\Phi _{{\itGamma _1}}},{\itPi _2} \circ {\Phi _{{\itGamma _2}}}} \right)} \right],
    \end{aligned}
    \label{eq:a.13}
\end{equation}
where Step (\textit{b}) is valid because of Lemma A.6 in \citet{shalev2014understanding} that for $\forall a \in \mathbb{R}$, it is satisfied ${{\mathbb{E}}_{{{\tilde \sigma }_{ijk}}}}\exp ({\tilde \sigma _{ijk}}a) = 0.5\left[ {\exp (a) + \exp ( - a)} \right] \le \exp (0.5{a^2})$, Step (\textit{c}) is from Eq.~\eqref{eq:a.12}, and Step (\textit{a}) is demonstrated as follows.

A random variable sequence ${Y_K}$, $K = 1, 2,...,$ is defined as
\begin{equation}
    {Y_K} = \exp \left( {w\sum\limits_{i = 1}^N {\sum\limits_{j = 1}^{{w_{\itPi ,{\rm{out}}}}} {\sum\limits_{k = 1}^K {{{\tilde \sigma }_{ijk}}{{\Delta \alpha }_{ijk}}} } } }\right).
    \label{eq:a.14}
\end{equation}
This sequence satisfies
\begin{equation}
    \begin{aligned}
        &\mathop {\lim }\limits_{{K_1},{K_2} \to  + \infty } {\mathbb{E}_{{ \tilde{\boldsymbol{\upsigma}} }}}{Y_{{K_1}}}{Y_{{K_2}}}\\
        &\;\;\; = \mathop {\lim }\limits_{{K_1},{K_2} \to  + \infty } {\mathbb{E}_{{\tilde{\boldsymbol{\upsigma}}}}}\exp \left( {w\sum\limits_{i = 1}^N {\sum\limits_{j = 1}^{{w_{\itPi ,{\rm{out}}}}} {\sum\limits_{k = 1}^{{K_1}} {{{\tilde \sigma }_{ijk}}{{\Delta \alpha }_{ijk}}} } } } \right)\exp \left( {w\sum\limits_{i = 1}^N {\sum\limits_{j = 1}^{{w_{\itPi ,{\rm{out}}}}} {\sum\limits_{k = 1}^{{K_2}} {{{\tilde \sigma }_{ijk}}{{\Delta \alpha }_{ijk}}} } } } \right)\\
        &\;\;\;\mathop  = \limits^{(a)} \mathop {\lim }\limits_{{K_{\min }},{K_{\max }} \to  + \infty } {\mathbb{E}_{{\tilde{\boldsymbol{\upsigma}}}}}\exp \left[ {w\left( {2\sum\limits_{i = 1}^N {\sum\limits_{j = 1}^{{w_{\itPi ,{\rm{out}}}}} {\sum\limits_{k = 1}^{{K_{\min }}} {{{\tilde \sigma }_{ijk}}{{\Delta \alpha }_{ijk}}} } }  + \sum\limits_{i = 1}^N {\sum\limits_{j = 1}^{{w_{\itPi ,{\rm{out}}}}} {\sum\limits_{k = {K_{\min }} + 1}^{{K_{\max }}} {{{\tilde \sigma }_{ijk}}{{\Delta \alpha }_{ijk}}} } } } \right)} \right]\\
        &\;\;\; = \mathop {\lim }\limits_{{K_{\min }},{K_{\max }} \to  + \infty } {\mathbb{E}_{{\tilde{\boldsymbol{\upsigma}}}}}\exp \left( {2w\sum\limits_{i = 1}^N {\sum\limits_{j = 1}^{{w_{\itPi ,{\rm{out}}}}} {\sum\limits_{k = 1}^{{K_{\min }}} {{{\tilde \sigma }_{ijk}}{{\Delta \alpha}_{ijk}}} } } } \right){\mathbb{E}_{{\tilde{\boldsymbol{\upsigma}}}}}\exp \left( {w\sum\limits_{i = 1}^N {\sum\limits_{j = 1}^{{w_{\itPi ,{\rm{out}}}}} {\sum\limits_{k = {K_{\min }} + 1}^{{K_{\max }}} {{{\tilde \sigma }_{ijk}}{{\Delta \alpha}_{ijk}}} } } } \right)\\
        &\;\;\; = \mathop {\lim }\limits_{{K_{\min }},{K_{\max }} \to  + \infty } {A_{{K_{\min }}}}B_{{K_{\min }}}^{{K_{\max }}},
    \end{aligned}
    \label{eq:a.15}
\end{equation}
where
\begin{equation}
    {A_{{K_{\min }}}} = \prod\limits_{i = 1}^N {\prod\limits_{j = 1}^{{w_{\itPi ,{\rm{out}}}}} {\prod\limits_{k = 1}^{{K_{\min }}} {0.5\left[ {\exp \left( {2w\Delta {\alpha _{ijk}}} \right) + \exp \left( { - 2w\Delta {\alpha _{ijk}}} \right)} \right]} } } \le \exp \left( {2{w^2}\sum\limits_{i = 1}^N {\sum\limits_{j = 1}^{{w_{\itPi ,{\rm{out}}}}} {\sum\limits_{k = 1}^{{K_{\min }}} {\Delta \alpha _{ijk}^2} } } } \right),
    \label{eq:a.16}
\end{equation}
\begin{equation}
    1 \le B_{{K_{\min }}}^{{K_{\max }}} = \prod\limits_{i = 1}^N {\prod\limits_{j = 1}^{{w_{\itPi ,{\rm{out}}}}} {\prod\limits_{k = {K_{\min }} + 1}^{{K_{\max }}} {0.5\left[ {\exp \left( {w\Delta {\alpha _{ijk}}} \right) + \exp \left( { -w\Delta {\alpha _{ijk}}} \right)} \right]} } } \le \exp \left( {0.5{w^2}\sum\limits_{i = 1}^N {\sum\limits_{j = 1}^{{w_{\itPi ,{\rm{out}}}}} {\sum\limits_{k = {K_{\min }} + 1}^{{K_{\max }}} {\Delta \alpha _{ijk}^2} } } } \right),
    \label{eq:a.17}
\end{equation}
and in Step (\textit{a}) of Eq.~\eqref{eq:a.15} $K_{\text{min}} = \min\left(K_1,K_2\right)$ and $K_{\text{max}} = \max\left(K_1,K_2\right)$. Since the sequence $\Delta \alpha_{ijk}$ is square-summable and $0.5\left[ {\exp \left( a\right) + \exp \left(-a\right)} \right] \geq 1$ for all $a \in \mathbb{R}$, $A_{K_{\text{min}}}$ is a monotonically increasing and bounded sequence with respect to $K_{\text{min}}$, and its limit exists. $B_{K_{\text{min}}}^{K_{\text{max}}}$ is also a bounded sequence with respect to ${K_{\text{min}}}$ and ${K_{\text{max}}}$, and its limit is 1 because of $1 = \mathop {\lim }\limits_{{K_{\min }},{K_{\max }} \to  + \infty } \exp \left( {0.5{w^2}\sum_{i = 1}^N {\sum_{j = 1}^{{w_{\itPi ,{\rm{out}}}}} {\sum_{k = {K_{\min }} + 1}^{{K_{\max }}} {\Delta \alpha _{ijk}^2} } } } \right) \ge \mathop {\lim }\limits_{{K_{\min }},{K_{\max }} \to  + \infty } B_{{K_{\min }}}^{{K_{\max }}} \ge 1$. Since the values of $A_{K_{\text{min}}}$ and $B_{K_{\text{min}}}^{K_{\text{max}}}$ are bounded, the limit in Eq.~\eqref{eq:a.15} exists and equals that of $A_{K_{\text{min}}}$, and it is independent of the manner in which $K_{\text{min}}$ and $K_{\text{max}}$ approach infinity. It is also independent of the manner in which $K_1$ and $K_2$ approach infinity. Thus, it can be obtained $\mathop {\lim }\limits_{{K_1},{K_2} \to  + \infty } {\mathbb{E}_{{\tilde{\boldsymbol \upsigma }}}}{\left( {{Y_{{K_2}}} - {Y_{{K_1}}}} \right)^2} = \mathop {\lim }\limits_{{K_1},{K_2} \to  + \infty } {\mathbb{E}_{{\tilde{\boldsymbol \upsigma }}}}\left( Y_{{K_1}}^2 + Y_{{K_2}}^2\right.$ $\left.- {Y_{{K_1}}}{Y_{{K_2}}} - {Y_{{K_2}}}{Y_{{K_1}}}\right) = 0$. This result indicates that the random variable sequence $Y_K$ converges in mean-square, thereby allowing the expectation and limit to be interchanged in Step (a) of Eq.~\eqref{eq:a.13} \citep{huang2022probability}.

Eqs. \eqref{eq:a.10} and \eqref{eq:a.13} prove that $\kappa \left( {{\it{\Pi}}  \circ {\Phi _{\it{\Gamma}} },{\boldsymbol{\tilde \upsigma }}} \right)$ is a zero-mean sub-Gaussian process with respect to $\PiPhiGamma$ over $\FPiPhiGamma$. The diameter of $\kappa \left( {{\it{\Pi}}  \circ {\Phi _{\it{\Gamma}} },{\boldsymbol{\tilde \upsigma }}} \right)$ over $\FPiPhiGamma$ is bounded by
\begin{equation}
    \begin{array}{l}
        {\Delta _\kappa }\left( {{{\cal F}_{\itPi  \circ {\Phi _\itGamma }}}} \right) = \mathop {\sup }\limits_{{\itPi _1} \circ {\Phi _{{\itGamma _1}}},{\itPi _2} \circ {\Phi _{{\itGamma _2}}} \in {{\cal F}_{\itPi  \circ {\Phi _\itGamma }}}} {d_\kappa }\left( {{\itPi _1} \circ {\Phi _{{\itGamma _1}}},{\itPi _2} \circ {\Phi _{{\itGamma _2}}}} \right) \mathop  \leq \limits^{(a)} \mathop 2\sqrt {NT{w_{\itPi ,{\rm{out}}}}} {B_\itPi },
    \end{array}
    \label{eq:a.18}
\end{equation}
where Step (\textit{a}) holds from Eq.~\eqref{eq:a.12}.
\hfill$\blacksquare$

\subsection{Proof of Lemma 5}
\label{appA.3}
{\bf Proof:} For $\forall \itGamma(\cdot) \in \FGamma$, the difference between $\itGamma(\mathbf{x}_1)$ and $\itGamma(\mathbf{x}_2)$ is bounded by
\begin{equation}
    \begin{aligned}
        &\left| {\itGamma \left({{\bf{x}}_1}\right) - \itGamma \left({{\bf{x}}_2}\right)} \right| = \left| {{{\bf{W}}_2}\left[ {{\sigma _{{\rm{ReLU}}}}\left( {{{\bf{W}}_1}{{\bf{x}}_1} + {{\bf{b}}_1}} \right) - {\sigma _{{\rm{ReLU}}}}\left( {{{\bf{W}}_1}{{\bf{x}}_2} + {{\bf{b}}_1}} \right)} \right]} \right|\\
        &\;\;\;\;\;\;\;\;\;\;\;\;\;\;\;\;\;\;\;\;\;\;\; \le {\left| {{{\bf{W}}^{\top}_2}} \right|_{\infty,1}}\left| {\left[ {{\sigma _{{\rm{ReLU}}}}\left( {{{\bf{W}}_1}{{\bf{x}}_1} + {{\bf{b}}_1}} \right) - {\sigma _{{\rm{ReLU}}}}\left( {{{\bf{W}}_1}{{\bf{x}}_2} + {{\bf{b}}_1}} \right)} \right]} \right|\\
        &\;\;\;\;\;\;\;\;\;\;\;\;\;\;\;\;\;\;\;\;\;\;\; \le {w_{\itGamma ,{\rm{out}}}}{\left| {{{\bf{W}}_2}} \right|_{\infty ,\infty }}\left| {\left[ {{{\bf{W}}_1}\left( {{{\bf{x}}_1} - {{\bf{x}}_2}} \right)} \right]} \right|\\
        &\;\;\;\;\;\;\;\;\;\;\;\;\;\;\;\;\;\;\;\;\;\;\; \le {w_{\itGamma ,{\rm{out}}}}{\left| {{{\bf{W}}_2}} \right|_{\infty ,\infty }}{\left| {{{\bf{W}}_1}} \right|_{\infty ,\infty }}{w_\itGamma }\left| {{{\bf{x}}_1} - {{\bf{x}}_2}} \right| \le {w_{\itGamma ,{\rm{out}}}}{w_\itGamma }B_{\bf{W}}^2\left| {{{\bf{x}}_1} - {{\bf{x}}_2}} \right|.
    \end{aligned}
    \label{eq:a.19}
\end{equation}
Thus, $L_{\FGamma} = \wGammaout\wGamma\BW^2$ is an $L^1$-norm Lipschitz constant of all $\itGamma(\cdot) \in \FGamma$. For $\forall \itPi(\cdot) \in \FPi$, the difference between $\itPi(\mathbf{x}_1)$ and $\itPi(\mathbf{x}_2)$ is bounded by
\begin{equation}
    \begin{aligned}
        &\left| {\itPi ({{\bf{x}}_1}) - \itPi ({{\bf{x}}_2})} \right| \le {\left| {{{{\tilde{\bf W}}}^{\top}_{{h_\itPi }}}} \right|_{\infty,1}}\left| {{\sigma _{{\rm{ReLU}}}}\left[ { \ldots {\sigma _{{\rm{ReLU}}}}\left( {{{{\tilde{\bf W}}}_1}{{\bf{x}}_1} + {{{\tilde{\bf b}}}_1}} \right)} \right] - {\sigma _{{\rm{ReLU}}}}\left[ { \ldots {\sigma _{{\rm{ReLU}}}}\left( {{{{\tilde{\bf W}}}_1}{{\bf{x}}_2} + {{{\tilde{\bf b}}}_1}} \right)} \right]} \right|\\
        &\;\;\;\;\;\;\;\;\;\;\;\;\;\;\;\;\;\;\;\;\;\;\;\; \le {w_{\itPi ,{\rm{out}}}}{B_{{\tilde{\bf W}}}}\left| {{\sigma _{{\rm{ReLU}}}}\left[ { \ldots {\sigma _{{\rm{ReLU}}}}\left( {{{{\tilde{\bf W}}}_1}{{\bf{x}}_1} + {{{\tilde{\bf b}}}_1}} \right)} \right] - {\sigma _{{\rm{ReLU}}}}\left[ { \ldots {\sigma _{{\rm{ReLU}}}}\left( {{{{\tilde{\bf W}}}_1}{{\bf{x}}_2} + {{{\tilde{\bf b}}}_1}} \right)} \right]} \right|\\
        &\;\;\;\;\;\;\;\;\;\;\;\;\;\;\;\;\;\;\;\;\;\;\;\; \le {w_{\itPi ,{\rm{out}}}}w_\itPi ^{{h_\Pi } - 1}B_{{\tilde{\bf W}}}^{{h_\itPi }}\left| {{{\bf{x}}_1} - {{\bf{x}}_2}}\right|.
    \end{aligned}
    \label{eq:a.20}
\end{equation}
Thus, $L_{\FPi} = \wPiout\wPi^{\hPi-1}\BtildeW^{\hPi}$ is an $L^1$-norm Lipschitz constant of all $\itPi(\cdot) \in \FPi$.
\hfill$\blacksquare$

\subsection{Proof of Lemma 6}
\label{appA.4}
{\bf Proof:} Since the weight matrices and vectors of all $\itGamma(\cdot) \in \FGamma$ in Assumption 9 are bounded and $\sigma_{\text{ReLU}}(\cdot)$ is Lipschitz continuous, the {\color{black} second-order} ODE in Eq. \eqref{eq:2.1} with the initial conditions $\mathbf{x}(0) = {\mathbf{x}}'(0) = \mathbf{0}$ has unique solutions $\mathbf{x}(t) = \PhiGamma\left[\mathbf{u}(\tau)\right](t)$ for all $\mathbf{u}(t) \in K$. $\mathbf{x}(t)$ and its derivative ${\mathbf{x}}'(t)$ satisfy
\begin{equation}
    \begin{aligned}
        &\left| {{\bf{x}}\left(t\right)} \right| + \left| {{\bf{x'}}\left(t\right)} \right| = \left| {\int_0^t {{\bf{x'}}(\tau ){\rm{d}}\tau } } \right| + \left| {\int_0^t {\itGamma \left[ {{\bf{x}}(\tau ),{\bf{x'}}(\tau ),{\bf{u}}(\tau )} \right]{\rm{d}}\tau } } \right|\\
        &\;\;\;\;\;\;\;\;\;\;\;\;\;\;\;\;\;\;\;\;\;\le \int_0^t {\left| {{\bf{x'}}(\tau )} \right|{\rm{d}}\tau }  + \int_0^t {\left| {\itGamma \left[ {{\bf{x}}(\tau ),{\bf{x'}}(\tau ),{\bf{u}}(\tau )} \right]} \right|{\rm{d}}\tau } \\
        &\;\;\;\;\;\;\;\;\;\;\;\;\;\;\;\;\;\;\;\;\;\mathop \le \limits^{(a)} \int_0^t {\left| {{\bf{x'}}(\tau )} \right|{\rm{d}}\tau }  + L_{\FGamma}\int_0^t {\left[ {\left| {{\bf{x}}(\tau )} \right| + \left| {{\bf{x'}}(\tau )} \right| + \left| {{\bf{u}}(\tau )} \right|} \right]{\rm{d}}\tau }  + \int_0^t {\left| {\itGamma \left( {\bf{0}} \right)} \right|{\rm{d}}\tau } \\
        &\;\;\;\;\;\;\;\;\;\;\;\;\;\;\;\;\;\;\;\;\;\le \left( {L_{\FGamma} + 1} \right)\int_0^t {\left[ {\left| {{\bf{x}}(\tau )} \right| + \left| {{\bf{x'}}(\tau )} \right|} \right]{\rm{d}}\tau } + pT{B_K}L_{\FGamma} + T\left[ {\left| {{{\bf{W}}_2}{\sigma _{\mathrm{ReLU}}}\left( {{{\bf{b}}_1}} \right)} \right| + \left| {{{\bf{b}}_2}} \right|} \right]\\
        &\;\;\;\;\;\;\;\;\;\;\;\;\;\;\;\;\;\;\;\;\;\le \left( {L_{\FGamma} + 1} \right)\int_0^t {\left[ {\left| {{\bf{x}}(\tau )} \right| + \left| {{\bf{x'}}(\tau )} \right|} \right]{\rm{d}}\tau }  + T\left[ {p{B_K}L_{\FGamma} + \wGammaout{B_{\bf{b}}}\left( {{w_\itGamma }{B_{\bf{W}}} + 1} \right)} \right]\\
        &\;\;\;\;\;\;\;\;\;\;\;\;\;\;\;\;\;\;\;\;\;\mathop  \le \limits^{(b)} T\left[ {p{B_K}L_{\FGamma} + \wGammaout{B_{\bf{b}}}\left( {{w_\itGamma }{B_{\bf{W}}} + 1} \right)} \right]{e^{T\left( {L_{\FGamma} + 1} \right)}},
    \end{aligned}
    \label{eq:a.21}
\end{equation}
where in Step (\textit{a}) $L_{\FGamma}$ is an arbitrary $L^1$-norm Lipschitz constant for all $\itGamma(\cdot) \in \FGamma$ and Step (\textit{b}) is from the Grönwall's inequality \citep{ames1997inequalities}.
\hfill$\blacksquare$

\subsection{Proof of Lemma 7}
\label{appA.5}
{\bf Proof:} As explained in the proof of Lemma 6, ${\mathbf{x}}_1(t) = \Phi_{\itGamma_1}\left[\mathbf{u}(\tau)\right](t)$ and ${\mathbf{x}}_2(t) = \Phi_{\itGamma_2}\left[\mathbf{u}(\tau)\right](t)$ are unique solutions of the {\color{black} second-order} ODE in Eq.~\eqref{eq:2.1} for all $\mathbf{u}(t) \in K$. From Appendix D and Lemma 6 in \citet{huang2025upper}, $\left|\mathbf{x}_2(t) - \mathbf{x}_1(t)\right|$ can be bounded by
\begin{equation}
    \left| {{{\bf{x}}_2}(t) - {{\bf{x}}_1}(t)} \right| \le {e^{T\left( {L_{\FGamma} + 1} \right)}}T\mathop {\max }\limits_{\tau  \in [0,T]} \left| {{\itGamma _2}\left[ {{{\bf{x}}_2}(\tau ),{{{\bf{x}}}'_2}(\tau ),{\bf{u}}(\tau )} \right] - {\itGamma _1}\left[ {{{\bf{x}}_2}(\tau ),{{{\bf{x}}}'_2}(\tau ),{\bf{u}}(\tau )} \right]} \right|,
    \label{eq:a.22}
\end{equation}
where $L_{\FGamma}$ is an arbitrary $L^1$-norm Lipschitz constant that simultaneously bounds all $\itGamma(\cdot) \in \FGamma$. $\left| {{\itGamma _2}\left[ {{{\bf{x}}_2}(\tau ),{{{\bf{x}}}'_2}(\tau ),{\bf{u}}(\tau )} \right] - {\itGamma _1}\left[ {{{\bf{x}}_2}(\tau ),{{{\bf{x}}}'_2}(\tau ),{\bf{u}}(\tau )} \right]} \right|$ in Eq. \eqref{eq:a.22} can be bounded by
\begin{equation}
    \begin{aligned}
        &\left| {{\itGamma _2}\left[ {{{\bf{x}}_2}(\tau ),{{{\bf{x}}}'_2}(\tau ),{\bf{u}}(\tau )} \right] - {\itGamma _1}\left[ {{{\bf{x}}_2}(\tau ),{{{\bf{x}}}'_2}(\tau ),{\bf{u}}(\tau )} \right]} \right|\\
        &\; \le \left| {{{\bf{W}}_{2,2}}{\sigma _{{\rm{ReLU}}}}\left\{ {{{\bf{W}}_{2,1}}{{\left[ {{\bf{x}}_2^{\top}(t),{\bf{x}'}_2^{\top}(t),{{\bf{u}}^{\top}}(t)} \right]}^{\top}} + {{\bf{b}}_{2,1}}} \right\} + {{\bf{b}}_{2,2}}} \right.\\
        &\;\;\;\;\;\left. { - {{\bf{W}}_{1,2}}{\sigma _{{\rm{ReLU}}}}\left\{ {{{\bf{W}}_{1,1}}{{\left[ {{\bf{x}}_2^{\top}(t),{\bf{x'}}_2^{\top}(t),{{\bf{u}}^{\top}}(t)} \right]}^{\top}} + {{\bf{b}}_{1,1}}} \right\} - {{\bf{b}}_{1,2}}} \right|\\
        &\; \le \left| {{{\bf{W}}_{2,2}}{\sigma _{{\rm{ReLU}}}}\left\{ {{{\bf{W}}_{2,1}}{{\left[ {{\bf{x}}_2^{\top}(t),{\bf{x'}}_2^{\top}(t),{{\bf{u}}^{\top}}(t)} \right]}^{\top}} + {{\bf{b}}_{2,1}}} \right\}} - {{\bf{W}}_{1,2}}{\sigma _{{\rm{ReLU}}}}\left\{ {{{\bf{W}}_{2,1}}{{\left[ {{\bf{x}}_2^{\top}(t),{\bf{x'}}_2^{\top}(t),{{\bf{u}}^{\top}}(t)} \right]}^{\top}} + {{\bf{b}}_{2,1}}} \right\}\right.\\
        &\;\;\;\;\left. + {{\bf{W}}_{1,2}}{\sigma _{{\rm{ReLU}}}}\left\{ {{{\bf{W}}_{2,1}}{{\left[ {{\bf{x}}_2^{\top}(t),{\bf{x'}}_2^{\top}(t),{{\bf{u}}^{\top}}(t)} \right]}^{\top}} + {{\bf{b}}_{2,1}}} \right\}{- {{\bf{W}}_{1,2}}{\sigma _{{\rm{ReLU}}}}\left\{ {{{\bf{W}}_{1,1}}{{\left[ {{\bf{x}}_2^{\top}(t),{\bf{x'}}_2^{\top}(t),{{\bf{u}}^{\top}}(t)} \right]}^{\top}} + {{\bf{b}}_{1,1}}} \right\}} \right|\\  &\;\;\;\;+\left| {{{\bf{b}}_{2,2}} - {{\bf{b}}_{1,2}}} \right|\\
        &\; \le {w_{\itGamma ,{\rm{out}}}}{\left| {{{\bf{W}}_{2,2}} - {{\bf{W}}_{1,2}}} \right|_{\infty ,\infty }}\left| {{\sigma _{{\rm{ReLU}}}}\left\{ {{{\bf{W}}_{2,1}}{{\left[ {{\bf{x}}_2^{\top}(t),{\bf{x'}}_2^{\top}(t),{{\bf{u}}^{\top}}(t)} \right]}^{\top}} + {{\bf{b}}_{2,1}}} \right\}} \right|\\
        &\;\;\;\; + {w_{\itGamma ,{\rm{out}}}}{\left| {{{\bf{W}}_{1,2}}} \right|_{\infty ,\infty }}\left| {\left\{ {{{\bf{W}}_{2,1}}{{\left[ {{\bf{x}}_2^{\top}(t),{\bf{x'}}_2^{\top}(t),{{\bf{u}}^{\top}}(t)} \right]}^{\top}} + {{\bf{b}}_{2,1}}} \right\}}{-\left\{ {{{\bf{W}}_{1,1}}{{\left[ {{\bf{x}}_2^{\top}(t),{\bf{x'}}_2^{\top}(t),{{\bf{u}}^{\top}}(t)} \right]}^{\top}} + {{\bf{b}}_{1,1}}} \right\}} \right|\\  &\;\;\;\;+ {w_{\itGamma ,{\rm{out}}}}{\left| {{{\bf{b}}_{2,2}} - {{\bf{b}}_{1,2}}} \right|_\infty }\\
        &\; \le \varsigma {w_{\itGamma ,{\rm{out}}}}\left| {{{\bf{W}}_{2,1}}{{\left[ {{\bf{x}}_2^{\top}(t),{\bf{x'}}_2^{\top}(t),{{\bf{u}}^{\top}}(t)} \right]}^{\top}} + {{\bf{b}}_{2,1}}} \right| + \varsigma {w_{\itGamma ,{\rm{out}}}}\\
        &\;\;\;\; + {w_{\itGamma ,{\rm{out}}}}{B_{\bf{W}}}\left\{\left| \left( {{{\bf{W}}_{2,1}} - {{\bf{W}}_{1,1}}} \right){{\left[ {{\bf{x}}_2^{\top}(t),{\bf{x'}}_2^{\top}(t),{{\bf{u}}^{\top}}(t)} \right]}^{\top}}\right| + \left|{{\bf{b}}_{2,1}} - {{\bf{b}}_{1,1}}\right|\right\}\\
        &\;\mathop  \le \limits^{(a)} \varsigma {w_{\itGamma ,{\rm{out}}}}\left( {{w_\itGamma }{B_{\bf{W}}}\left\{ T\left[ {p{B_K}L_{\FGamma} + \wGammaout{B_{\bf{b}}}\left( {{w_\itGamma }{B_{\bf{W}}} + 1} \right)} \right]{e^{T\left( {L_{\FGamma} + 1} \right)}} +pB_{K} \right\} + {w_\itGamma }{B_{\bf{b}}}} \right)\\
        &\;\;\;\; + {w_{\itGamma ,{\rm{out}}}}{B_{\bf{W}}}\left( {{w_\itGamma }\varsigma \left\{ T\left[ {p{B_K}L_{\FGamma} + \wGammaout{B_{\bf{b}}}\left( {{w_\itGamma }{B_{\bf{W}}} + 1} \right)} \right]{e^{T\left( {L_{\FGamma} + 1} \right)}} +pB_{K} \right\} + {w_\itGamma }\varsigma } \right) + \varsigma {w_{\itGamma ,{\rm{out}}}}\\
        &\;\mathop  \le \limits^{(b)} \varsigma {w_{\itGamma ,{\rm{out}}}}\wGamma\left( {{B_{\bf{W}}}\left\{ T\left[ {p{B_K}\left(L_{\FGamma}+1\right) + \wGammaout{B_{\bf{b}}}\left( {{w_\itGamma }{B_{\bf{W}}} + 1} \right)} \right]{e^{T\left( {L_{\FGamma} + 1} \right)}} \right\}+{B_{\bf{b}}}} \right)\\
        &\;\;\;\; + {w_{\itGamma ,{\rm{out}}}}{B_{\bf{W}}}{\wGamma}\varsigma\left( {\left\{ T\left[ {p{B_K}\left(L_{\FGamma}+1\right) + \wGammaout{B_{\bf{b}}}\left( {{w_\itGamma }{B_{\bf{W}}} + 1} \right)} \right]{e^{T\left( {L_{\FGamma} + 1} \right)}} \right\}+1} \right) + \varsigma {w_{\itGamma ,{\rm{out}}}}\\
        &\;\leq 2\varsigma {w_{\itGamma ,{\rm{out}}}}\wGamma \max\left(B_{\bf{W}},B_{\bf{b}}\right)\left\{{ T\left[ {p{B_K}\left(L_{\FGamma}+1\right) + \wGammaout{B_{\bf{b}}}\left( {{w_\itGamma }{B_{\bf{W}}} + 1} \right)} \right]{e^{T\left( {L_{\FGamma} + 1} \right)}} +1} \right\}+ \varsigma {w_{\itGamma ,{\rm{out}}}}\\
        &\;\mathop  \le \limits^{(c)} 2\varsigma {\wGammaout}\wGamma{pT}{e^{T\left( {L_{\FGamma} + 1} \right)}} \left\{\max\left(B_{\bf{W}},B_{\bf{b}}\right)\left[{{B_K}\left(L_{\FGamma}+1\right) + \wGammaout{B_{\bf{b}}}\left( {{w_\itGamma }{B_{\bf{W}}} + 1} \right)}+1\right] +1\right\},
    \end{aligned}
    \label{eq:a.23}
\end{equation}
where Step (\textit{a}) utilizes Eq.~\eqref{eq:3.13} in Lemma 6 and Steps (\textit{b}) and (\textit{c}) hold from $T \geq 1$ in Assumption 1. Substituting Eq.~\eqref{eq:a.23} into Eq.~\eqref{eq:a.22}, Eq.~\eqref{eq:3.14} is proven.
\hfill$\blacksquare$

\subsection{Proof of Lemma 8}
\label{appA.6}
{\bf Proof:} $\left|\itPi_1(\mathbf{x}) - \itPi_2(\mathbf{x}) \right|$ is bounded by
\begin{equation}
    \begin{aligned}
        &\left| {{\itPi _1}({\bf{x}}) - {\itPi _2}({\bf{x}})} \right| \le \left| {{{{\tilde{\bf W}}}_{1,{h_\itPi }}}{\sigma _{{\rm{ReLU}}}}\left[ { \ldots {\sigma _{{\rm{ReLU}}}}\left( {{{{\tilde{\bf W}}}_{1,1}}{\bf{x}} + {{{\tilde{\bf b}}}_{1,1}}} \right)} \right]} \right.\\
        &\;\;\;\;\;\;\;\;\;\;\;\;\;\;\;\;\;\;\;\;\;\;\;\;\;\;\;\left. { - {{{\tilde{\bf W}}}_{2,{h_\itPi }}}{\sigma _{{\rm{ReLU}}}}\left[ { \ldots {\sigma _{{\rm{ReLU}}}}\left( {{{{\tilde{\bf W}}}_{2,1}}{\bf{x}} + {{{\tilde{\bf b}}}_{2,1}}} \right)} \right]} \right| + \left| {{{{\tilde{\bf b}}}_{1,{h_\itPi }}} - {{{\tilde{\bf b}}}_{2,{h_\itPi }}}} \right|\\
        &\;\;\;\;\;\;\;\;\;\;\;\;\;\;\;\;\;\;\;\;\;\;\; \le \left| {{{{\tilde{\bf W}}}_{1,{h_\itPi }}}{\sigma _{{\rm{ReLU}}}}\left[ { \ldots {\sigma _{{\rm{ReLU}}}}\left( {{{{\tilde{\bf W}}}_{1,1}}{\bf{x}} + {{{\tilde{\bf b}}}_{1,1}}} \right)} \right]} \right. - {{{\tilde{\bf W}}}_{2,{h_\itPi }}}{\sigma _{{\rm{ReLU}}}}\left[ { \ldots {\sigma _{{\rm{ReLU}}}}\left( {{{{\tilde{\bf W}}}_{1,1}}{\bf{x}} + {{{\tilde{\bf b}}}_{1,1}}} \right)} \right]\\
        &\;\;\;\;\;\;\;\;\;\;\;\;\;\;\;\;\;\;\;\;\;\;\;\;\;\;\; + {{{\tilde{\bf W}}}_{2,{h_\itPi }}}{\sigma _{{\rm{ReLU}}}}\left[ { \ldots {\sigma _{{\rm{ReLU}}}}\left( {{{{\tilde{\bf W}}}_{1,1}}{\bf{x}} + {{{\tilde{\bf b}}}_{1,1}}} \right)} \right]\\
        &\;\;\;\;\;\;\;\;\;\;\;\;\;\;\;\;\;\;\;\;\;\;\;\;\;\;\;\left. { - {{{\tilde{\bf W}}}_{2,{h_\itPi }}}{\sigma _{{\rm{ReLU}}}}\left[ { \ldots {\sigma _{{\rm{ReLU}}}}\left( {{{{\tilde{\bf W}}}_{2,1}}{\bf{x}} + {{{\tilde{\bf b}}}_{2,1}}} \right)} \right]} \right| + {w_{\itPi ,{\rm{out}}}}\varsigma \\
        &\;\;\;\;\;\;\;\;\;\;\;\;\;\;\;\;\;\;\;\;\;\;\;\le {\left| {{{{\tilde{\bf W}}}_{1,{h_\itPi }}}^{\top} - {{{\tilde{\bf W}}}_{2,{h_\itPi }}}^{\top}} \right|_{\infty ,1}}\left| {{\sigma _{{\rm{ReLU}}}}\left[ { \ldots {\sigma _{{\rm{ReLU}}}}\left( {{{{\tilde{\bf W}}}_{1,1}}{\bf{x}} + {{{\tilde{\bf b}}}_{1,1}}} \right)} \right]} \right| + {w_{\itPi ,{\rm{out}}}}\varsigma\\
        &\;\;\;\;\;\;\;\;\;\;\;\;\;\;\;\;\;\;\;\;\;\;\;\;\;\;\;+ L_{\FPi,\mathrm{layer}}\left| {{\sigma _{{\rm{ReLU}}}}\left[ { \ldots {\sigma _{{\rm{ReLU}}}}\left( {{{{\tilde{\bf W}}}_{1,1}}{\bf{x}} + {{{\tilde{\bf b}}}_{1,1}}} \right)} \right]}{ - {\sigma _{{\rm{ReLU}}}}\left[ { \ldots {\sigma _{{\rm{ReLU}}}}\left( {{{{\tilde{\bf W}}}_{2,1}}{\bf{x}} + {{{\tilde{\bf b}}}_{2,1}}} \right)} \right]} \right|,
    \end{aligned}
    \label{eq:a.24}
\end{equation}
where $L_{\FPi,\mathrm{layer}} = \max_{j \in \left[1,2,\dots,h_{\itPi}\right]}{L_{\FPi,j}}$ and $L_{\FPi,j}$ is an arbitrary $L^1$-norm Lipschitz constant that bounds the mappings between the ${\left(j-1\right)}^{\mathrm{th}}$ and $j^{\mathrm{th}}$ layers of all $\itPi(\cdot) \in \FPi$, for $j = 1, 2,\dots,h_{\itPi}$. $\left|{{\sigma _{{\rm{ReLU}}}}\left[ { \ldots {\sigma _{{\rm{ReLU}}}}\left( {{{{\tilde{\bf W}}}_{1,1}}{\bf{x}} + {{{\tilde{\bf b}}}_{1,1}}} \right)} \right]}\right|$ in Eq.~\eqref{eq:a.24} is bounded by
\begin{equation}
    \begin{aligned}
        &\left| {{\sigma _{{\rm{ReLU}}}}\left[ { \ldots {\sigma _{{\rm{ReLU}}}}\left( {{{{\tilde{\bf W}}}_{1,1}}{\bf{x}} + {{{\tilde{\bf b}}}_{1,1}}} \right)} \right]} \right|\le L_{\FPi,\mathrm{layer}}\left| {{\sigma _{{\rm{ReLU}}}}\left[ { \ldots {\sigma _{{\rm{ReLU}}}}\left( {{{{\tilde{\bf W}}}_{1,1}}{\bf{x}} + {{{\tilde{\bf b}}}_{1,1}}} \right)} \right]} \right| + {w_\itPi }{B_{{\tilde{\bf b}}}}\\
        &\;\;\;\;\;\;\;\;\;\;\;\;\;\;\;\;\;\;\;\;\;\;\;\;\;\;\;\;\;\;\;\;\;\;\;\;\;\;\;\;\;\;\;\;\;\;\;\;\; \le L_{\FPi,\mathrm{layer}}^2\left| {{\sigma _{{\rm{ReLU}}}}\left[ { \ldots {\sigma _{{\rm{ReLU}}}}\left( {{{{\tilde{\bf W}}}_{1,1}}{\bf{x}} + {{{\tilde{\bf b}}}_{1,1}}} \right)} \right]} \right| + {w_\itPi }{B_{{\tilde{\bf b}}}}\left( {1 + L_{\FPi,\mathrm{layer}}} \right)\\
        &\;\;\;\;\;\;\;\;\;\;\;\;\;\;\;\;\;\;\;\;\;\;\;\;\;\;\;\;\;\;\;\;\;\;\;\;\;\;\;\;\;\;\;\;\;\;\;\;\; \le L_{\FPi,\mathrm{layer}}^{h_{\itPi}-1}\left| {\bf{x}} \right| + {w_\itPi }{B_{{\tilde{\bf b}}}}\sum\limits_{i = 0}^{{h_\itPi } - 2}L_{\FPi,\mathrm{layer}}^i.
    \end{aligned}
    \label{eq:a.25}
\end{equation}
Substituting Eq.~\eqref{eq:a.25} into Eq.~\eqref{eq:a.24}, $\left|\itPi_1(\mathbf{x}) - \itPi_2(\mathbf{x}) \right|$ is bounded by
\begin{equation}
    \begin{aligned}
        &\left| {{\itPi_1}({\bf{x}}) - {\itPi_2}({\bf{x}})} \right|\\
        &\;\;\;\le {w_{\itPi ,{\rm{out}}}}\varsigma \left(L_{\FPi,\mathrm{layer}}^{h_{\itPi}-1}\left| {\bf{x}} \right| + {w_\itPi }{B_{{\tilde{\bf b}}}}\sum\limits_{i = 0}^{{h_\itPi } - 2}L_{\FPi,\mathrm{layer}}^i + 1 \right)\\
        &\;\;\;\;\;\;+ L_{\FPi,\mathrm{layer}}\left| {{{{\tilde{\bf W}}}_{1,{h_\itPi } - 1}}{\sigma _{{\rm{ReLU}}}}\left[ { \ldots {\sigma _{{\rm{ReLU}}}}\left( {{{{\tilde{\bf W}}}_{1,1}}{\bf{x}} + {{{\tilde{\bf b}}}_{1,1}}} \right)} \right] + {{{\tilde{\bf b}}}_{1,{h_\itPi } - 1}}}{- {{{\tilde{\bf W}}}_{2,{h_\itPi } - 1}}{\sigma _{{\rm{ReLU}}}}\left[ { \ldots {\sigma _{{\rm{ReLU}}}}\left( {{{{\tilde{\bf W}}}_{2,1}}{\bf{x}} + {{{\tilde{\bf b}}}_{2,1}}} \right)} \right] - {{{\tilde{\bf b}}}_{2,{h_\itPi } - 1}}} \right|.
    \end{aligned}
    \label{eq:a.26}
\end{equation}
Similarly, the last term of Eq.~\eqref{eq:a.26} is bounded by
\begin{equation}
    \begin{aligned}
        &\left| {{{{\tilde{\bf W}}}_{1,{h_\itPi } - 1}}{\sigma _{{\rm{ReLU}}}}\left[ { \ldots {\sigma _{{\rm{ReLU}}}}\left( {{{{\tilde{\bf W}}}_{1,1}}{\bf{x}} + {{{\tilde{\bf b}}}_{1,1}}} \right)} \right] + {{{\tilde{\bf b}}}_{1,{h_\itPi } - 1}}}- {{{\tilde{\bf W}}}_{2,{h_\itPi } - 1}}{\sigma _{{\rm{ReLU}}}}\left[ { \ldots {\sigma _{{\rm{ReLU}}}}\left( {{{{\tilde{\bf W}}}_{2,1}}{\bf{x}} + {{{\tilde{\bf b}}}_{2,1}}} \right)} \right] - {{{\tilde{\bf b}}}_{2,{h_\itPi } - 1}} \right|\\
        &\;\;\; \le {w_\itPi }\varsigma \left(L_{\FPi,\mathrm{layer}}^{h_{\itPi}-2}\left| {\bf{x}} \right| + {w_\itPi }{B_{{\tilde{\bf b}}}}\sum\limits_{i = 0}^{{h_\itPi } - 3}L_{\FPi,\mathrm{layer}}^i + 1 \right)\\
        &\;\;\;\;\;\; + L_{\FPi,\mathrm{layer}}\left| {{{{\tilde{\bf W}}}_{1,{h_\itPi } - 2}}{\sigma _{{\rm{ReLU}}}}\left[ { \ldots {\sigma _{{\rm{ReLU}}}}\left( {{{{\tilde{\bf W}}}_{1,1}}{\bf{x}} + {{{\tilde{\bf b}}}_{1,1}}} \right)} \right] + {{{\tilde{\bf b}}}_{1,{h_\itPi } - 2}}} - {{{\tilde{\bf W}}}_{2,{h_\itPi } - 2}}{\sigma _{{\rm{ReLU}}}}\left[ { \ldots {\sigma _{{\rm{ReLU}}}}\left( {{{{\tilde{\bf W}}}_{2,1}}{\bf{x}} + {{{\tilde{\bf b}}}_{2,1}}} \right)} \right] - {{{\tilde{\bf b}}}_{2,{h_\itPi } - 2}} \right|.
    \end{aligned}
    \label{eq:a.27}
\end{equation}
By solving the iteration in Eq.~\eqref{eq:a.27}, the last term of Eq.~\eqref{eq:a.26} is bounded by
\begin{equation}
    \begin{aligned}
        &\left| {{{{\tilde{\bf W}}}_{1,{h_\itPi } - 1}}{\sigma _{{\rm{ReLU}}}}\left[ { \ldots {\sigma _{{\rm{ReLU}}}}\left( {{{{\tilde{\bf W}}}_{1,1}}{\bf{x}} + {{{\tilde{\bf b}}}_{1,1}}} \right)} \right] + {{{\tilde{\bf b}}}_{1,{h_\itPi } - 1}}} - {{{\tilde{\bf W}}}_{2,{h_\itPi } - 1}}{\sigma _{{\rm{ReLU}}}}\left[ { \ldots {\sigma _{{\rm{ReLU}}}}\left( {{{{\tilde{\bf W}}}_{2,1}}{\bf{x}} + {{{\tilde{\bf b}}}_{2,1}}} \right)} \right] - {{{\tilde{\bf b}}}_{2,{h_\itPi } - 1}} \right|\\
        &\;\;\; \le {w_\itPi }\varsigma \left(L_{\FPi,\mathrm{layer}}^{h_{\itPi}-2}\left| {\bf{x}} \right| + {w_\itPi }{B_{{\tilde{\bf b}}}}\sum\limits_{i = 0}^{{h_\itPi } - 3}L_{\FPi,\mathrm{layer}}^i + 1 \right) + {w_\itPi }\varsigma L_{\FPi,\mathrm{layer}} \left(L_{\FPi,\mathrm{layer}}^{h_{\itPi}-3}\left| {\bf{x}} \right| + {w_\itPi }{B_{{\tilde{\bf b}}}}\sum\limits_{i = 0}^{{h_\itPi } - 4}L_{\FPi,\mathrm{layer}}^i + 1 \right)\\
        &\;\;\;\;\;\; + L_{\FPi,\mathrm{layer}}^2\left| {{{{\tilde{\bf W}}}_{1,{h_\itPi } - 3}}{\sigma _{{\rm{ReLU}}}}\left[ { \ldots {\sigma _{{\rm{ReLU}}}}\left( {{{{\tilde{\bf W}}}_{1,1}}{\bf{x}} + {{{\tilde{\bf b}}}_{1,1}}} \right)} \right] + {{{\tilde{\bf b}}}_{1,{h_\itPi } - 3}}}- {{{\tilde{\bf W}}}_{2,{h_\itPi } - 3}}{\sigma _{{\rm{ReLU}}}}\left[ { \ldots {\sigma _{{\rm{ReLU}}}}\left( {{{{\tilde{\bf W}}}_{2,1}}{\bf{x}} + {{{\tilde{\bf b}}}_{2,1}}} \right)} \right] - {{{\tilde{\bf b}}}_{2,{h_\itPi } - 3}} \right|\\
        &\;\;\; \le {w_\itPi }\varsigma \sum\limits_{j = 0}^{{h_\itPi } - 3}{L_{\FPi,\mathrm{layer}}^j\left(L_{\FPi,\mathrm{layer}}^{h_{\itPi}-2-j}\left| {\bf{x}} \right| + {w_\itPi }{B_{{\tilde{\bf b}}}}\sum\limits_{i = 0}^{{h_\itPi } - 3-j}L_{\FPi,\mathrm{layer}}^i + 1 \right)} + L_{\FPi,\mathrm{layer}}^{{h_\itPi} - 2}\left| {\left( {{{{\tilde{\bf W}}}_{1,1}}{\bf{x}} + {{{\tilde{\bf b}}}_{1,1}}} \right) - \left( {{{{\tilde{\bf W}}}_{2,1}}{\bf{x}} + {{{\tilde{\bf b}}}_{2,1}}} \right)} \right|\\
        &\;\;\; \le {w_\itPi }\varsigma \sum\limits_{j = 0}^{{h_\itPi } - 3}{L_{\FPi,\mathrm{layer}}^j\left(L_{\FPi,\mathrm{layer}}^{h_{\itPi}-2-j}\left| {\bf{x}} \right| + {w_\itPi }{B_{{\tilde{\bf b}}}}\frac{1 - L_{\FPi,\mathrm{layer}}^{{h_\itPi } - 2 - j}}{1 -L_{\FPi,\mathrm{layer}}} + 1 \right)} + L_{\FPi,\mathrm{layer}}^{{h_\itPi} - 2}{w_\itPi }\varsigma \left( {\left| {\bf{x}} \right| + 1} \right)\\
        &\;\;\; = {w_\itPi }\varsigma \sum\limits_{i = 0}^{{h_\itPi } - 2}{L_{\FPi,\mathrm{layer}}^i\left(L_{\FPi,\mathrm{layer}}^{h_{\itPi}-2-i}\left| {\bf{x}} \right| + {w_\itPi }{B_{{\tilde{\bf b}}}}\frac{1 - L_{\FPi,\mathrm{layer}}^{{h_\itPi } - 2 - i}}{1 -L_{\FPi,\mathrm{layer}}} + 1 \right)}.
    \end{aligned}
    \label{eq:a.28}
\end{equation}
Substituting Eq.~\eqref{eq:a.28} into Eq.~\eqref{eq:a.26}, $\left|\itPi_1(\mathbf{x}) - \itPi_2(\mathbf{x}) \right|$ is bounded by
\begin{equation}
    \begin{aligned}
        &\left|{{\itPi _1}({\bf{x}}) - {\itPi _2}({\bf{x}})} \right| \le {w_{\itPi ,{\rm{out}}}}\varsigma \left(L_{\FPi,\mathrm{layer}}^{h_{\itPi}-1}\left| {\bf{x}} \right| + {w_\itPi }{B_{{\tilde{\bf b}}}}\sum\limits_{i = 0}^{{h_\itPi } - 2}L_{\FPi,\mathrm{layer}}^i + 1 \right)\\
        &\;\;\;\;\;\;\;\;\;\;\;\;\;\;\;\;\;\;\;\;\;\;\;\;\;\;\;+ L_{\FPi,\mathrm{layer}}{w_\itPi }\varsigma \sum\limits_{i = 0}^{{h_\itPi } - 2}{L_{\FPi,\mathrm{layer}}^i\left(L_{\FPi,\mathrm{layer}}^{h_{\itPi}-2-i}\left| {\bf{x}} \right| + {w_\itPi }{B_{{\tilde{\bf b}}}}\frac{1 - L_{\FPi,\mathrm{layer}}^{{h_\itPi } - 2 - i}}{1 -L_{\FPi,\mathrm{layer}}} + 1 \right)}\\
        &\;\;\;\;\;\;\;\;\;\;\;\;\;\;\;\;\;\;\;\;\;\;\;= {w_{\itPi ,{\rm{out}}}}\varsigma \left(L_{\FPi,\mathrm{layer}}^{h_{\itPi}-1}\left| {\bf{x}} \right| + {w_\itPi }{B_{{\tilde{\bf b}}}}\sum\limits_{i = 0}^{{h_\itPi } - 2}L_{\FPi,\mathrm{layer}}^i + 1 \right)\\
        &\;\;\;\;\;\;\;\;\;\;\;\;\;\;\;\;\;\;\;\;\;\;\;\;\;\;\;+ {w_\itPi }\varsigma \sum\limits_{i = 1}^{{h_\itPi } - 1}{L_{\FPi,\mathrm{layer}}^i\left(L_{\FPi,\mathrm{layer}}^{h_{\itPi}-1-i}\left| {\bf{x}} \right| + {w_\itPi }{B_{{\tilde{\bf b}}}}\frac{1 - L_{\FPi,\mathrm{layer}}^{{h_\itPi } - 1 - i}}{1 -L_{\FPi,\mathrm{layer}}} + 1 \right)}\\
        &\;\;\;\;\;\;\;\;\;\;\;\;\;\;\;\;\;\;\;\;\;\;\; \le \max\left(\wGammaout,\wGamma\right)\varsigma \sum\limits_{i = 0}^{{h_\itPi } - 1}{L_{\FPi,\mathrm{layer}}^i\left(L_{\FPi,\mathrm{layer}}^{h_{\itPi}-1-i}\left| {\bf{x}} \right| + {w_\itPi }{B_{{\tilde{\bf b}}}}\frac{1 - L_{\FPi,\mathrm{layer}}^{{h_\itPi } - 1 - i}}{1 -L_{\FPi,\mathrm{layer}}} + 1 \right)}\\
        &\;\;\;\;\;\;\;\;\;\;\;\;\;\;\;\;\;\;\;\;\;\;\; = \max\left(\wGammaout,\wGamma\right)\varsigma\left( {h_\itPi }{L_{\FPi,\mathrm{layer}}^{{h_\itPi } - 1}}\left| {\bf{x}} \right| + \sum\limits_{i = 0}^{{h_\itPi } - 1} {L_{\FPi,\mathrm{layer}}^i} + {w_\itPi }{B_{{\tilde{\bf b}}}}\sum\limits_{i = 0}^{{h_\itPi } - 2} {\sum\limits_{j = i}^{{h_\itPi } - 2}{L_{\FPi,\mathrm{layer}}^j}}\right)\\
        &\;\;\;\;\;\;\;\;\;\;\;\;\;\;\;\;\;\;\;\;\;\;\; \leq {\varsigma}\max{\left(\wGammaout,\wGamma\right)}{h^2_{\itPi}}\left(L_{\FPi,\mathrm{layer}}^{h_\itPi - 1}+1\right)\left(\left| {\bf{x}} \right| + \wPi\Btildeb + 1\right).
    \end{aligned}
    \label{eq:a.29}
\end{equation}
This completes the proof.
\hfill$\blacksquare$

\subsection{Proof of Lemma 9}
\label{appA.7}
{\bf Proof:} $\left| {{\itPi _1} \circ {\Phi _{{\itGamma _1}}}\left[ {{\bf{u}}(\tau )} \right](t) - {\itPi _2} \circ {\Phi _{{\itGamma _2}}}\left[ {{\bf{u}}(\tau )} \right](t)} \right|$ is bounded by
\begin{equation}
    \begin{aligned}
        &\left| {{\itPi _1} \circ {\Phi _{{\itGamma _1}}}\left[ {{\bf{u}}(\tau )} \right](t) - {\itPi _2} \circ {\Phi _{{\itGamma _2}}}\left[ {{\bf{u}}(\tau )} \right](t)} \right|\\ 
        &\;\;\;\;\; \le \left| {{\itPi _1} \circ {\Phi _{{\itGamma _1}}}\left[ {{\bf{u}}(\tau )} \right](t) - {\itPi _1} \circ {\Phi _{{\itGamma _2}}}\left[ {{\bf{u}}(\tau )} \right](t)} \right| + \left| {{\itPi _1} \circ {\Phi _{{\itGamma _2}}}\left[ {{\bf{u}}(\tau )} \right](t) - {\itPi _2} \circ {\Phi _{{\itGamma _2}}}\left[ {{\bf{u}}(\tau )} \right](t)} \right|.
    \end{aligned}
    \label{eq:a.30}
\end{equation}
From Lemma 7, $\left| {{\itPi _1} \circ {\Phi _{{\itGamma _1}}}\left[ {{\bf{u}}(\tau )} \right](t) - {\itPi _1} \circ {\Phi _{{\itGamma _2}}}\left[ {{\bf{u}}(\tau )} \right](t)} \right|$ in Eq.~\eqref{eq:a.30} is bounded by
\begin{equation}
    \begin{aligned}
        &\left| {{\itPi _1} \circ {\Phi _{{\itGamma _1}}}\left[ {{\bf{u}}(\tau )} \right](t) - {\itPi _1} \circ {\Phi _{{\itGamma _2}}}\left[ {{\bf{u}}(\tau )} \right](t)} \right|\\ 
        &\;\;\; \le 2L_{\FPi}\varsigma {\wGammaout}\wGamma{pT^2}{e^{2T\left( {L_{\FGamma} + 1} \right)}} \left\{{\max\left(B_{\bf{W}},B_{\bf{b}}\right)\left[{{B_K}\left(L_{\FGamma}+1\right) + \wGammaout{B_{\bf{b}}}\left( {{w_\itGamma }{B_{\bf{W}}} + 1} \right)}+1\right]} +1\right\},
    \end{aligned}
    \label{eq:a.31}
\end{equation}
where $L_{\FPi}$ is an arbitrary $L^1$-norm Lipschitz constant for all $\itPi(\cdot) \in \FPi$.
From Eq.~\eqref{eq:2.1}, Lemma 6, and Lemma 8, $\left| {{\itPi _1} \circ {\Phi _{{\itGamma _2}}}\left[ {{\bf{u}}(\tau )} \right](t) - {\itPi _2} \circ {\Phi _{{\itGamma _2}}}\left[ {{\bf{u}}(\tau )} \right](t)} \right|$ in Eq.~\eqref{eq:a.30} with $\mathbf{x}(t) = {\Phi _{{\itGamma _2}}}\left[ {{\bf{u}}(\tau )} \right](t)$ is bounded by
\begin{equation}
    \begin{aligned}
        &\left| {{\itPi _1} \circ {\Phi _{{\itGamma _2}}}\left[ {{\bf{u}}(\tau )} \right](t) - {\itPi _2} \circ {\Phi _{{\itGamma _2}}}\left[ {{\bf{u}}(\tau )} \right](t)} \right|\\ 
        &\le \varsigma\max\left(\wGammaout,\wGamma\right)h_{\itPi}^2\left(L_{\FPi,\mathrm{layer}}^{h_\itPi - 1}+1\right)\left({\left| \mathbf{x}(t) \right| + \left| {{{\bf{u}}_0}} \right| + t} + \wPi\Btildeb + 1\right) \\
        &\le \varsigma\max\left(\wGammaout,\wGamma\right)h_{\itPi}^2\left(L_{\FPi,\mathrm{layer}}^{h_\itPi - 1}+1\right)\left\{T\left[ {p{B_K}L_{\FGamma} + \wGammaout{B_{\bf{b}}}\left( {{w_\itGamma }{B_{\bf{W}}} + 1} \right)} \right]{e^{T\left( {L_{\FGamma} + 1} \right)}} + pB_K + T + \wPi\Btildeb + 1\right\} \\
        &\le \varsigma\max\left(\wGammaout,\wGamma\right)h_{\itPi}^2\left(L_{\FPi,\mathrm{layer}}^{h_\itPi - 1}+1\right)T\left[ {p{B_K}\left(L_{\FGamma} + 1\right)+ \wGammaout{B_{\bf{b}}}\left( {{w_\itGamma }{B_{\bf{W}}} + 1} \right)}+\wPi\Btildeb + 2\right]{e^{T\left( {L_{\FGamma} + 1} \right)}} \\
        &\le \varsigma\max\left(\wGammaout,\wGamma\right)h_{\itPi}^2\left(L_{\FPi,\mathrm{layer}}^{h_\itPi - 1}+1\right)pT{e^{T\left( {L_{\FGamma} + 1} \right)}}\left[ {{B_K}\left(L_{\FGamma} + 1\right)+ \wGammaout{B_{\bf{b}}}\left( {{w_\itGamma }{B_{\bf{W}}} + 1} \right)}+\wPi\Btildeb + 2\right].
    \end{aligned}
    \label{eq:a.32}
\end{equation}
Substituting Eqs.~\eqref{eq:a.31} and \eqref{eq:a.32} into Eq.~\eqref{eq:a.30}, $\left| {{\itPi _1} \circ {\Phi _{{\itGamma _1}}}\left[ {{\bf{u}}(\tau )} \right](t) - {\itPi _2} \circ {\Phi _{{\itGamma _2}}}\left[ {{\bf{u}}(\tau )} \right](t)} \right|$ can be bounded by
\begin{equation}
    \begin{aligned}
        &\left| {{\itPi _1} \circ {\Phi _{{\itGamma _1}}}\left[ {{\bf{u}}(\tau )} \right](t) - {\itPi _2} \circ {\Phi _{{\itGamma _2}}}\left[ {{\bf{u}}(\tau )} \right](t)} \right|\\        
        &\;\; \le 2L_{\FPi}\varsigma {\wGammaout}\wGamma{pT^2}{e^{2T\left( {L_{\FGamma} + 1} \right)}} \left\{{\max\left(B_{\bf{W}},B_{\bf{b}}\right)\left[{{B_K}\left(L_{\FGamma}+1\right) + \wGammaout{B_{\bf{b}}}\left( {{w_\itGamma }{B_{\bf{W}}} + 1} \right)}+1\right]} +1\right\} \\        &\;\;\;\;\;+\varsigma\max\left(\wGammaout,\wGamma\right)h_{\itPi}^2\left(L_{\FPi,\mathrm{layer}}^{h_\itPi - 1}+1\right)pT{e^{T\left( {L_{\FGamma} + 1} \right)}}\left[ {{B_K}\left(L_{\FGamma} + 1\right)+ \wGammaout{B_{\bf{b}}}\left( {{w_\itGamma }{B_{\bf{W}}} + 1} \right)}+\wPi\Btildeb + 2\right]\\    
        &\;\; \le 3\varsigma\left(L_{\FPi,\mathrm{layer}}^{h_\itPi}+1\right)\wPi{\wGammaout^2}\wGamma^2{h_{\itPi}^2}{pT^2}{e^{2T\left( {L_{\FGamma} + 1} \right)}}\left(L_{\FGamma} + 1\right){B_K}\left\{\left[\max\left(B_{\bf{W}},B_{\bf{b}}\right)+1\right]\left[B_{\bf{b}}\left(B_{\bf{W}} + 1 \right)+2\right]+\Btildeb +2\right\}.
    \end{aligned}
    \label{eq:a.33}
\end{equation}
This completes the proof.
\hfill$\blacksquare$

\subsection{Proof of Lemma 10}
\label{appA.8}
{\bf Proof:} Under Assumptions 1, 2, 8, 9, and 10, given MLPs $\itGamma_1(\cdot)$ and $\itGamma_2(\cdot)$ $\in \FGamma$ and MPLs $\itPi_1(\cdot)$ and $\itPi_2(\cdot)$ $\in \FPi$, the differences between the weight matrices and vectors of $\itGamma_1(\cdot)$ and $\itGamma_2(\cdot)$ and those of $\itPi_1(\cdot)$ and $\itPi_2(\cdot)$ are bounded by $\varsigma$, then, from Eq.~\eqref{eq:3.9} in Lemma 4 and Eq.~\eqref{eq:3.16} in Lemma 9, the pseudo-metric $d_{\kappa}\left(\PiPhiGammaone,\PiPhiGammatwo\right)$ between $\PiPhiGammaone$ and $\PiPhiGammatwo$ $\in \FPiPhiGamma$, is bounded by
\begin{equation}
    \begin{aligned}
        &{d_\kappa }\left( {{\itPi _1} \circ {\Phi _{{\itGamma _1}}},{\itPi _2} \circ {\Phi _{{\itGamma _2}}}} \right) = \sqrt {\sum\limits_{i = 1}^N {\left\| {{\itPi _1} \circ {\Phi _{{\itGamma _1}}}\left[ {{{\bf{u}}_i}(\tau )} \right](t) - {\itPi _2} \circ {\Phi _{{\itGamma _2}}}\left[ {{{\bf{u}}_i}(\tau )} \right](t)} \right\|_{{L^2}}^2} } \\
        &\;\;\;\;\;\;\;\;\;\;\;\;\;\;\;\;\;\;\;\;\;\;\;\;\;\;\;\;\;\;\;\;\; \le \sqrt {TN} \mathop {\max }\limits_{t \in \left[ {0,T} \right]} \left| {{\itPi _1} \circ {\Phi _{{\itGamma _1}}}\left[ {{\bf{u}}(\tau )} \right](t) - {\itPi _2} \circ {\Phi _{{\itGamma _2}}}\left[ {{\bf{u}}(\tau )} \right](t)} \right| \le \varsigma \sqrt {NT{w_{\itPi ,{\rm{out}}}}} {B_\itPi }{\Delta _{\itPi  \circ {\Phi _\itGamma }}},
    \end{aligned}
    \label{eq:a.34}
\end{equation}
where ${\Delta _{\itPi  \circ {\Phi _\itGamma }}}$ is in Eq.~\eqref{eq:3.19}.

For all neural oscillators  $\PiPhiGamma \in \FPiPhiGamma$ in Assumption 10, the number $N_{\mathbf{W}}$ of the total {\color{black} trainable} parameters in the weight matrices of an arbitrary $\itGamma(\cdot) \in \FGamma$ is ${N_{\bf{W}}} = {w_\itGamma }\left( {3{w_{\itGamma ,{\rm{out}}}} + p} \right)$, the number $N_{\mathbf{b}}$ of the total {\color{black} trainable} parameters in the weight vectors of $\itGamma(\cdot) \in \FGamma$ is ${N_{\bf{b}}} = {w_{\itGamma ,{\rm{out}}}} + {w_\itGamma }$, the number $N_{\tilde{\mathbf{W}}}$ of the total {\color{black} trainable} parameters in the weight matrices of an arbitrary $\itPi(\cdot) \in \FPi$ is ${N_{{\tilde{\bf W}}}} = w_\itPi ^2\left( {{h_\itPi } - 2} \right) + {w_\itPi }\left( {{w_{\itGamma ,{\rm{out}}}} + p + {w_{\itPi ,{\rm{out}}}} + 1} \right)$, and the number $N_{\tilde{\mathbf{b}}}$ of the total {\color{black} trainable} parameters in the weight vectors of $\itPi(\cdot) \in \FPi$ is ${N_{{\tilde{\bf b}}}} = {w_{\itPi ,{\rm{out}}}} + \left( {{h_\itPi } - 1} \right){w_\itPi}$. By discretizing the range of values for each weight parameter using a uniform grid with a grid size of $\varsigma$, the number of the obtained neural oscillators is
\begin{equation}
    {\left( {\frac{{2{B_{\bf{W}}}}}{\varsigma }} \right)^{{N_{\bf{W}}}}}{\left( {\frac{{2{B_{\bf{b}}}}}{\varsigma }} \right)^{{N_{\bf{b}}}}}{\left( {\frac{{2{B_{{\tilde{\bf W}}}}}}{\varsigma }} \right)^{{N_{{\tilde{\bf W}}}}}}{\left( {\frac{{2{B_{\bf{b}}}}}{\varsigma }} \right)^{{N_{{\tilde{\bf b}}}}}}.
    \label{eq:a.35}
\end{equation}
Based on Eq.~\eqref{eq:a.34}, these neural oscillators cover the neural oscillator class $\FPiPhiGamma$ by a distance $\varepsilon  = \varsigma \sqrt {NT{w_{\itPi ,{\rm{out}}}}} {B_\itPi }{\Delta _{\itPi  \circ {\Phi _\itGamma }}}$. Thus, the covering number $N\left(\FPiPhiGamma,d_{\kappa},\varepsilon\right)$ can bounded by
\begin{equation}
    \begin{aligned}
        &N\left( {{{\cal F}_{H \circ {\Phi _G}}},{d_\kappa },\varepsilon } \right) \le {\left( {\frac{{{B_{\bf{W}}}\sqrt {NT{w_{\itPi ,{\rm{out}}}}} {B_\itPi }{\Delta _{\itPi  \circ {\Phi _\itGamma }}}}}{{0.5\varepsilon }}} \right)^{{N_{\bf{W}}}}}{\left( {\frac{{{B_{\bf{b}}}\sqrt {NT{w_{\itPi ,{\rm{out}}}}} {B_\itPi }{\Delta _{\itPi  \circ {\Phi _\itGamma }}}}}{{0.5\varepsilon }}} \right)^{{N_{\bf{b}}}}}\\
        &\;\;\;\;\;\;\;\;\;\;\;\;\;\;\;\;\;\;\;\;\;\;\;\;\;\;\;\times{\left( {\frac{{{B_{{\tilde{\bf W}}}}\sqrt {NT{w_{\itPi ,{\rm{out}}}}} {B_\itPi }{\Delta _{\itPi  \circ {\Phi _\itGamma }}}}}{{0.5\varepsilon }}} \right)^{{N_{{\tilde{\bf W}}}}}}{\left( {\frac{{{B_{{\tilde{\bf b}}}}\sqrt {NT{w_{\itPi ,{\rm{out}}}}} {B_\itPi }{\Delta _{\itPi  \circ {\Phi _\itGamma }}}}}{{0.5\varepsilon }}} \right)^{{N_{{\tilde{\bf b}}}}}}.
    \end{aligned}
    \label{eq:a.36}
\end{equation}

Given an inequality
\begin{equation}
    \int_0^b {\sqrt {\ln \left( {\frac{a}{\varepsilon }} \right)} {\rm{d}}\varepsilon }  \le \int_0^b {\sqrt {\ln \left( {1 + \frac{a}{\varepsilon }} \right)} {\rm{d}}\varepsilon }  = a\int_0^{{b \mathord{\left/
    {\vphantom {b a}} \right.
    \kern-\nulldelimiterspace} a}} {\sqrt {\ln \left( {1 + \frac{1}{\varepsilon }} \right)} {\rm{d}}\varepsilon } \mathop  \le \limits^{(a)} b\sqrt {\ln \left[ {3\left( {1 + \frac{a}{b}} \right)} \right]},
    \label{eq:a.37}
\end{equation}
where Step (\textit{a}) is from Lemma C. 9 of \citet{foucart2013mathematical}. The integral in Eq.~\eqref{eq:3.12} is bounded by
\begin{equation}
    \begin{aligned}
        &\int_0^{\sqrt {NT{w_{\itPi ,{\rm{out}}}}} {B_\itPi }} {\sqrt {\ln N\left( {{{\cal F}_{H \circ {\Phi _G}}},{d_\kappa },\varepsilon } \right)} {\rm{d}}\varepsilon } \\
        &\; \le \int_0^{\sqrt {NT{w_{\itPi ,{\rm{out}}}}} {B_\itPi }} {\sqrt {{N_{\bf{W}}}\ln \left( {\frac{{{B_{\bf{W}}}\sqrt {NT{w_{\itPi ,{\rm{out}}}}} {B_\itPi }{\Delta _{\itPi  \circ {\Phi _\itGamma }}}}}{{0.5\varepsilon }}} \right)} {\rm{d}}\varepsilon }+ \int_0^{\sqrt {NT{w_{\itPi ,{\rm{out}}}}} {B_\itPi }} {\sqrt {{N_{\bf{b}}}\ln \left( {\frac{{{B_{\bf{b}}}\sqrt {NT{w_{\itPi ,{\rm{out}}}}} {B_\itPi }{\Delta _{\itPi  \circ {\Phi _\itGamma }}}}}{{0.5\varepsilon }}} \right)} {\rm{d}}\varepsilon } \\
        &\;\;\;\; + \int_0^{\sqrt {NT{w_{\itPi ,{\rm{out}}}}} {B_\itPi }} {\sqrt {{N_{{\tilde{\bf W}}}}\ln \left( {\frac{{{B_{{\tilde{\bf W}}}}\sqrt {NT{w_{\itPi ,{\rm{out}}}}} {B_\itPi }{\Delta _{\itPi  \circ {\Phi _\itGamma }}}}}{{0.5\varepsilon }}} \right)} {\rm{d}}\varepsilon }+ \int_0^{\sqrt {NT{w_{\itPi ,{\rm{out}}}}} {B_\itPi }} {\sqrt {{N_{{\tilde{\bf b}}}}\ln \left( {\frac{{{B_{{\tilde{\bf b}}}}\sqrt {NT{w_{\itPi ,{\rm{out}}}}} {B_\itPi }{\Delta _{\itPi  \circ {\Phi _\itGamma }}}}}{{0.5\varepsilon }}} \right)} {\rm{d}}\varepsilon } \\
        &\; \le \sqrt {NT{w_{\itPi ,{\rm{out}}}}} {B_\itPi }\sqrt {{w_\itGamma }\left( {3{w_{\itGamma ,{\rm{out}}}} + p} \right)\ln \left( {3 + 6{B_{\bf{W}}}{\Delta _{\itPi  \circ {\Phi _\itGamma }}}} \right)} \\
        &\;\;\;\; + \sqrt {NT{w_{\itPi ,{\rm{out}}}}} {B_\itPi }\sqrt {\left( {{w_{\itGamma ,{\rm{out}}}} + {w_\itGamma }} \right)\ln \left( {3 + 6{B_{\bf{b}}}{\Delta _{\itPi  \circ {\Phi _\itGamma }}}} \right)} \\
        &\;\;\;\; + \sqrt {NT{w_{\itPi ,{\rm{out}}}}} {B_\itPi }\sqrt {\left[ {w_\itPi ^2\left( {{h_\itPi } - 2} \right) + {w_\itPi }\left( {{w_{\itGamma ,{\rm{out}}}} + p + {w_{\itPi ,{\rm{out}}}} + 1} \right)} \right]\ln \left( {3 + 6{B_{{\tilde{\bf W}}}}{\Delta _{\itPi  \circ {\Phi _\itGamma }}}} \right)}\\
        &\;\;\;\; + \sqrt {NT{w_{\itPi ,{\rm{out}}}}} {B_\itPi }\sqrt {\left[ {{w_{\itPi ,{\rm{out}}}} + \left( {{h_\itPi } - 1} \right){w_\itPi }} \right]\ln \left( {3 + 6{B_{{\tilde{\bf b}}}}{\Delta _{\itPi  \circ {\Phi _\itGamma }}}} \right)}.
    \end{aligned}
    \label{eq:a.38}
\end{equation}
Substituting Eq.~\eqref{eq:a.38} into Eq.~\eqref{eq:3.12}, Eq.~\eqref{eq:3.18} is proven.
\hfill$\blacksquare$

\subsection{Proof of Theorem 1}
\label{appA.9}
{\bf Proof:} Under Assumptions 1-11, together with the conditions and results in Lemma 1, for the target causal and uniformly continuous operator $\Phi$, with the specified sizes of $\itGamma(\cdot)$ and $\itPi(\cdot)$ from Lemma 1, the generalization error $\loss\left(\hatPiPhiGamma\right)$ of a neural oscillator $\hatPiPhiGamma$, learned by Eq.~\eqref{eq:3.2} with \textit{N} i.i.d. sample pairs $\mathbf{U}_N(t)$ and $\mathbf{V}_N(t)$ from $\mu(\cdot)$, is bounded by Eq.~\eqref{eq:3.11} with probability at least $1-\delta$ for $\forall\delta \in \left(0,1\right)$. The empirical loss $\hatloss\left(\hatPiPhiGamma\right)$ in Eq.~\eqref{eq:3.11} can be bounded by $T\varepsilon_{\mathbf{y}}^2$ with $\varepsilon_{\mathbf{y}}$ in Eq.~\eqref{eq:2.6} of Lemma 1. The empirical Rademacher complexity $\Re_{\mathcal{G}_{\vartheta}}$ in Eq.~\eqref{eq:3.11} can be bounded by the expected supremum $\kappa\left(\PiPhiGamma, {\tilde{\boldsymbol{\upsigma}}}\right)$, as demonstrated in Lemma 3. By bounding the expected supremum of $\kappa\left(\PiPhiGamma, {\tilde{\boldsymbol{\upsigma}}}\right)$ using Eq.~\eqref{eq:3.18} in Lemma 10, together with $L_{\FGamma} = w_{\mathrm{max}}^2B_{\mathrm{max}}^2$ and $L_{\FPi,\mathrm{layer}}^{\hPi} = w_{\mathrm{max}}^{\hPi}B_{\mathrm{max}}^{\hPi}$ from Lemma 5, {\color{black} $h_{\itPi} = H_{\itPi}+1$ and $w_{\mathrm{max}} = \max{\left\{p\left(2M_{\itGamma} + 1\right),2pM_{\itGamma}, pM_{\itGamma}, p\left(M_{\itGamma} + 1\right)+1,q\left[p\left(M_{\itGamma} + 1\right) + 4\right],q\right\}} = \max{\left[p\left(2M_{\itGamma}+1\right),q\left(pM_{\itGamma}+p+4 \right)\right]}$ from Lemma 1,} Eq.~\eqref{eq:3.20} is proven.
\hfill$\blacksquare$

\subsection{Proof of Theorem 2}
\label{appA.10}
{\bf Proof:} Under Assumptions 1-11 and the conditions in Lemma 2, for the target second-order differential system governed by Eq.~\eqref{eq:2.7}, with the specified sizes of $\itGamma(\cdot)$ and $\itPi(\cdot)$ from Lemma 2, by replacing the $\varepsilon_{\mathbf{y}}$ in Eq.~\eqref{eq:3.20} with the bound ${\color{black} L_{\mathbf{h}}B_{\beta_{\mathbf{g}}}}\sqrt{r}\varepsilon_1+q\varepsilon_2$ in Eq.~\eqref{eq:2.8} of Lemma 2, the generalization error $\ell\left(\hatPiPhiGamma\right)$ of $\hatPiPhiGamma$ learned by Eq.~\eqref{eq:3.2} with \textit{N} i.i.d. sample pairs $\mathbf{U}_N(t)$ and $\mathbf{V}_N(t)$ from $\mu(\cdot)$ can be bounded by
\begin{equation}
    \begin{aligned}
        &\ell \left( {\hat \itPi  \circ {\Phi _{\hat \itGamma }}} \right) \le T{\left( {{\color{black} L_{\mathbf{h}}B_{\beta_{\mathbf{g}}}}\sqrt r {\varepsilon _1} + q{\varepsilon _2}} \right)^2}+3\frac{TqB_{{\rm{loss}}}^2}{{\sqrt N }}\left[172w_{\mathrm{max}}^{1.5}\sqrt{\ln \left( {3 + 6B_{\mathrm{max}}{\Delta _{\itPi  \circ {\Phi _\itGamma }}}} \right)}+\sqrt{0.5{{\color{black} \ln} \left( {2{\delta ^{ - 1}}} \right)}}\right]\\
        &\;\;\;\;\;\;\;\;\;\;\;\;\;\;\;\;\; \le 2T{\left({\color{black} L^2_{\mathbf{h}}B^2_{\beta_{\mathbf{g}}}}r {{\varepsilon}^2_1} + q^2{{\varepsilon}^2_2} \right)}+3\frac{TqB_{{\rm{loss}}}^2}{{\sqrt N }}\left[172w_{\mathrm{max}}^{1.5}\sqrt{\ln \left( {3 + 6B_{\mathrm{max}}{\Delta _{\itPi  \circ {\Phi _\itGamma }}}} \right)}+\sqrt{0.5{{\color{black} \ln} \left( {2{\delta ^{ - 1}}} \right)}}\right]
    \end{aligned}
    \label{eq:a.39}
\end{equation}
with probability at least $1-\delta$ for $\forall\delta \in \left(0,1\right)$, $\forall 0 < \varepsilon_1 < \alpha/r$, and $\forall \varepsilon_2 > 0$, where $\Delta_{\PiPhiGamma}$ is in Eq.~\eqref{eq:3.19} with $\hPi = 2$, $L_{\FGamma} = w_{\mathrm{max}}^2B_{\mathrm{max}}^2$, and {\color{black} $L_{\FPi,\mathrm{layer}}^{\hPi} = w_{\mathrm{max}}^{\hPi}B_{\mathrm{max}}^{\hPi}$}. From $\wGamma = 8r\ceil{C_{\itGamma} \varepsilon_{1}^{-2}}$, $\wPi = 8q\ceil{C_{\itPi} \varepsilon_{2}^{-2}}$, $\wGammaout = r$, and $\wPiout = q$ in Lemma 2, $\varepsilon^2_1$ and $\varepsilon^2_2$ can be replaced using $\wGamma$, and $\wPi$, respectively, thus obtaining Eq.~\eqref{eq:3.21}. {\color{black} $w_{\mathrm{max}} = \max\left(\wPiin,\wPi,\wPiout,\wGammain,\wGamma,\wGammaout\right)$ is simplified as
$w_{\mathrm{max}} = \max\left(2r+p,\wPi,r,r+p+1,\wGamma,q\right) =  \max\left(2r+p,\wPi,\wGamma\right)$, where $\wPi = 8q\ceil{C_{\itPi} \varepsilon_{2}^{-2}} > q$ is employed.}
\hfill$\blacksquare$

{\color{black} \subsection{Proof of Corollary 1}
\label{appA.11}
{\bf Proof:} Under the assumptions and conditions in Theorem 1, for the target causal and uniformly continuous operator $\Phi$, the neural oscillator $\hatPiPhiGamma$ learned by Eq.~\eqref{eq:3.2} with $N$ i.i.d. sample pairs $\mathbf{U}_N(t)$ and $\mathbf{V}_N(t)$ from $\mu(\cdot)$, and any Lipschitz continuous mapping $\Psi: \Czero{q} \to \mathbb{R}^s$ with an $L^2$-norm Lipschitz constant $L_{\Psi}$, let $\rho_{\Phi}$ and $\rho_{\hatPiPhiGamma}$ denote the probability measures on $\mathbb{R}^s$ of $\Psi\left\{\Phi\left[\mathbf{u}(\tau)\right](t)\right\}$ and $\Psi\left\{\hatPiPhiGamma\left[\mathbf{u}(\tau)\right](t)\right\}$, respectively, with $\mathbf{u}(t) \sim \mu(\cdot)$. Since $\Phi$ is uniformly continuous from Assumption 3, $\hatPiPhiGamma$ is the composition of the continuous solution operator of the second-order ODE in Eq.~\eqref{eq:2.1} with the Lipschitz continuous MLP $\itPi(\cdot)$ from Lemma 5, and $\Psi$ is Lipschitz continuous, both $\Psi\circ\Phi$ and $\Psi\circ\hatPiPhiGamma$ are Borel measurable. Together with Assumption 7 that $\mu(\cdot)$ is a Borel probability measure on the compact set $K$, the pushforward measures $\rho_{\Phi}$ and $\rho_{\hatPiPhiGamma}$ are well-defined Borel probability measures on the Polish space $\mathbb{R}^s$ equipped with the Euclidean topology.

By the Kantorovich-Rubinstein duality (Theorem 11.8.2 of \citealt{dudley2018real}), the Wasserstein-1 distance $W_1\left(\rho_{\Phi},\rho_{\hatPiPhiGamma}\right)$ admits the representation
\begin{equation}
    W_1\left(\rho_{\Phi},\rho_{\hatPiPhiGamma}\right) = \mathop{\sup}\limits_{\left\|f\right\|_{\mathrm{Lip}}\le 1} \left[\mathbb{E}_{\mathbf{u}\sim\mu}f\left(\Psi\left\{\Phi\left[\mathbf{u}(\tau)\right](t)\right\}\right) - \mathbb{E}_{\mathbf{u}\sim\mu}f\left(\Psi\left\{\hatPiPhiGamma\left[\mathbf{u}(\tau)\right](t)\right\}\right)\right],
    \label{eq:a.40}
\end{equation}
where the supremum is taken over all $1$-Lipschitz functions $f(\cdot): \mathbb{R}^s \to \mathbb{R}$ and $\left\|f\right\|_{\mathrm{Lip}}$ represents the smallest Lipschitz constant of $f(\cdot)$ under the $L^2$-norm. Coupling the two random vectors through the common input $\mathbf{u}(t)\sim\mu(\cdot)$ yields
\begin{equation}
    \begin{aligned}
        W_1\left(\rho_{\Phi},\rho_{\hatPiPhiGamma}\right) &\le \mathbb{E}_{\mathbf{u}\sim\mu}\left|\Psi\left\{\Phi\left[\mathbf{u}(\tau)\right](t)\right\} - \Psi\left\{\hatPiPhiGamma\left[\mathbf{u}(\tau)\right](t)\right\}\right|_2\\
        &\le L_{\Psi}\,\mathbb{E}_{\mathbf{u}\sim\mu}\normtwo{\Phi\left[\mathbf{u}(\tau)\right](t) - \hatPiPhiGamma\left[\mathbf{u}(\tau)\right](t)},
    \end{aligned}
    \label{eq:a.41}
\end{equation}
where the second inequality follows from the $L^2$-norm Lipschitz continuity of $\Psi$. Squaring both sides of Eq.~\eqref{eq:a.41} and applying Jensen's inequality produce
\begin{equation}
    \begin{aligned}
    &W_1^2\left(\rho_{\Phi},\rho_{\hatPiPhiGamma}\right) \le L_{\Psi}^2\left(\mathbb{E}_{\mathbf{u}\sim\mu}\normtwo{\Phi\left[\mathbf{u}(\tau)\right](t) - \hatPiPhiGamma\left[\mathbf{u}(\tau)\right](t)}\right)^2\\
    &\;\;\;\;\;\;\;\;\;\;\;\;\;\;\;\;\;\;\;\;\;\;\le L_{\Psi}^2\,\mathbb{E}_{\mathbf{u}\sim\mu}\normtwo{\Phi\left[\mathbf{u}(\tau)\right](t) - \hatPiPhiGamma\left[\mathbf{u}(\tau)\right](t)}^2 = L_{\Psi}^2\,\loss\left(\hatPiPhiGamma\right),
    \end{aligned}
    \label{eq:a.42}
\end{equation}
where the last equality follows from the definition of $\loss\left(\hatPiPhiGamma\right)$ in Eqs.~\eqref{eq:3.3} and \eqref{eq:3.4}. Substituting the upper bound of $\loss\left(\hatPiPhiGamma\right)$ established in Eq.~\eqref{eq:3.20} of Theorem 1 into Eq.~\eqref{eq:a.42}, Eq.~\eqref{eq:3.22} is proven.
\hfill$\blacksquare$}

{\color{black}
\subsection{Proof of Corollary 2}
\label{appA.12}
{\bf Proof:} Under the assumptions and conditions in Theorem 2, for the target second-order differential system governed by Eq.~\eqref{eq:2.7} denoted as $\hat{\mathbf{y}}(t)=\Phi\left[\mathbf{u}(\tau)\right](t)$, the neural oscillator $\hatPiPhiGamma$ learned by Eq.~\eqref{eq:3.2} with $N$ i.i.d. sample pairs $\mathbf{U}_N(t)$ and $\mathbf{V}_N(t)$ from $\mu(\cdot)$, and any Lipschitz continuous functional mapping $\Psi: \Czero{q} \to \mathbb{R}^s$ with an $L^2$-norm Lipschitz constant $L_{\Psi}$, let $\rho_{\Phi}$ and $\rho_{\hatPiPhiGamma}$ denote the probability measures on $\mathbb{R}^s$ of $\Psi\left\{\Phi\left[\mathbf{u}(\tau)\right](t)\right\}$ and $\Psi\left\{\hatPiPhiGamma\left[\mathbf{u}(\tau)\right](t)\right\}$, respectively, with $\mathbf{u}(t) \sim \mu(\cdot)$. Under the conditions in Lemma 2, the input-output mapping $\Phi$ of the target second-order differential system in Eq.~\eqref{eq:2.7} is continuous from $K \subset \Czero{p}$ into $\Czero{q}$. Following the same derivation as in Eqs.~\eqref{eq:a.40}-\eqref{eq:a.42} of Corollary 1, $\rho_{\Phi}$ and $\rho_{\hatPiPhiGamma}$ are well-defined Borel probability measures on the Polish space $\mathbb{R}^s$, and the squared Wasserstein-1 distance $W_1^2\left(\rho_{\Phi},\rho_{\hatPiPhiGamma}\right)$ between $\rho_{\Phi}$ and $\rho_{\hatPiPhiGamma}$ is bounded by
\begin{equation}
    W_1^2\left(\rho_{\Phi},\rho_{\hatPiPhiGamma}\right) \le L_{\Psi}^2\,\loss\left(\hatPiPhiGamma\right).
    \label{eq:a.43}
\end{equation}
Substituting the upper bound of $\loss\left(\hatPiPhiGamma\right)$ in Eq.~\eqref{eq:3.21} of Theorem 2 into Eq.~\eqref{eq:a.43}, Eq.~\eqref{eq:3.23} is proven.
\hfill$\blacksquare$}

\section{Proof of Eq.~\eqref{eq:3.12}}
\label{appB}
Given an arbitrary integer $M$, let $\mathcal{B}_m$, $m = 1, 2,\dots, M$, be the subsets of $\FPiPhiGamma$, that each $\mathcal{B}_m$ corresponds to a minimal $2\sqrt{NT\wPiout}\BPi2^{-m}$-cover of $\FPiPhiGamma$, with the covering number $N\left(\FPiPhiGamma,d_{\kappa},2\sqrt{NT\wPiout}\BPi2^{-m}\right)$, where $2\sqrt{NT\wPiout}\BPi$ is the bound of the diameter of $\FPiPhiGamma$, as shown in Eq.~\eqref{eq:3.10}. Then, given $\forall \PiPhiGamma \in \FPiPhiGamma$, it can be expressed as
\begin{equation}
    \kappa \left( {\PiPhiGamma,{\tilde{\boldsymbol \upsigma}}} \right) = \kappa \left( {{\it{\Pi}}\circ{\Phi_{\it{\Gamma}}},{\tilde{\boldsymbol \upsigma}}} \right) - \kappa \left( {{{\it{\Pi}}_M} \circ {\Phi_{{{\it{\Gamma}}_M}}},{\tilde{\boldsymbol \upsigma}}} \right) + \sum\limits_{m = 1}^M {\kappa \left( {{{\it{\Pi}}_m} \circ {\Phi_{{{\it{\Gamma}}_m}}},{\tilde{\boldsymbol \upsigma}}} \right) - \kappa \left( {{{\it{\Pi}}_{m-1}} \circ {\Phi_{{{\it{\Gamma}}_{m-1}}}},{\tilde{\boldsymbol \upsigma}}} \right)},
    \label{eq:b.1}
\end{equation}
where ${{\it{\Pi}}_{m-1}} \circ {\Phi_{{{\it{\Gamma}}_{m-1}}}} \in \mathcal{B}_{m-1}$ is closet to ${{\it{\Pi}}_{m}} \circ {\Phi_{{{\it{\Gamma}}_{m}}}}$, $m = 2, 3,\dots,M$, ${{\it{\Pi}}_{M}} \circ {\Phi_{{{\it{\Gamma}}_{M}}}} \in \mathcal{B}_{M}$ is closet to $\PiPhiGamma$, $\kappa \left( {{{\it{\Pi}}_{0}} \circ {\Phi_{{{\it{\Gamma}}_{0}}}},{\tilde{\boldsymbol \upsigma}}} \right) = 0$, and ${{\it{\Pi}}_{0}} \circ {\Phi_{{{\it{\Gamma}}_{0}}}}$ represents case where the parameter of the MLPs $\itGamma(\cdot)$ and $\itPi(\cdot)$ are all zeros. $\kappa \left( {\PiPhiGamma,{\tilde{\boldsymbol \upsigma}}} \right)$ can be bounded by
\begin{equation}
    \begin{aligned}
        &\kappa \left( {\PiPhiGamma,{\tilde{\boldsymbol \upsigma}}} \right) \le \mathop {\sup }\limits_{\scriptstyle{d_\kappa }\left( {{{\it{\Pi}}_1} \circ {\Phi_{{{\it{\Gamma}}_1}}},{{\it{\Pi}}_2} \circ {\Phi_{{{\it{\Gamma}}_2}}}} \right) \le {2\sqrt{NT\wPiout}{\BPi}2^{ - M}}\hfill\atop\scriptstyle{{{\it{\Pi}}_1} \circ {\Phi_{{{\it{\Gamma}}_1}}},{{\it{\Pi}}_2} \circ {\Phi_{{{\it{\Gamma}}_2}}}} \in {\FPiPhiGamma}\hfill} {\color{black} \left[\kappa \left( {{{\it{\Pi}}_1} \circ {\Phi_{{{\it{\Gamma}}_1}}},{\tilde{\boldsymbol \upsigma}}} \right) - \kappa \left( {{{\it{\Pi}}_2} \circ {\Phi_{{{\it{\Gamma}}_2}}},{\tilde{\boldsymbol \upsigma}}} \right)\right]}\\
        &\;\;\;\;\;\;\;\;\;\;\;\;\;\;\;\;\;\;\;\;\;\;\;\; + \sum\limits_{m = 1}^M {\mathop {\sup }\limits_{{{\it{\Pi}}_m} \circ {\Phi_{{{\it{\Gamma}}_m}}} \in {{\cal B}_m}} \left\{ {\kappa \left( {{{\it{\Pi}}_m} \circ {\Phi_{{{\it{\Gamma}}_m}}},{\tilde{\boldsymbol \upsigma}}} \right) - \kappa \left[ {{{\it{\Pi}}_{m-1}} \circ {\Phi_{{{\it{\Gamma}}_{m-1}}}}\left( {{\it{\Pi}}_m} \circ {\Phi_{{{\it{\Gamma}}_m}}} \right),{\tilde{\boldsymbol \upsigma}}} \right]} \right\}} 
    \end{aligned}
    \label{eq:b.2}
\end{equation}
and its expected supremum is bounded by
\begin{equation}
        \begin{aligned}    &\mathbb{E}_{{\tilde{\boldsymbol \upsigma }}}\kappa \left( {\PiPhiGamma,{\tilde{\boldsymbol \upsigma}}} \right) \le \mathbb{E}_{{\tilde{\boldsymbol \upsigma }}}\mathop {\sup }\limits_{\scriptstyle{d_\kappa }\left( {{{\it{\Pi}}_1} \circ {\Phi_{{{\it{\Gamma}}_1}}},{{\it{\Pi}}_2} \circ {\Phi_{{{\it{\Gamma}}_2}}}} \right) \le {2\sqrt{NT\wPiout}{\BPi}2^{ - M}}\hfill\atop\scriptstyle{{{\it{\Pi}}_1} \circ {\Phi_{{{\it{\Gamma}}_1}}},{{\it{\Pi}}_2} \circ {\Phi_{{{\it{\Gamma}}_2}}}} \in {\FPiPhiGamma}\hfill} {\color{black} \left[\kappa \left( {{{\it{\Pi}}_1} \circ {\Phi_{{{\it{\Gamma}}_1}}},{\tilde{\boldsymbol \upsigma}}} \right) - \kappa \left( {{{\it{\Pi}}_2} \circ {\Phi_{{{\it{\Gamma}}_2}}},{\tilde{\boldsymbol \upsigma}}} \right)\right]}\\
        &\;\;\;\;\;\;\;\;\;\;\;\;\;\;\;\;\;\;\;\;\;\;\;\;\;\;\;\;\; + \sum\limits_{m = 1}^M \mathbb{E}_{{\tilde{\boldsymbol \upsigma }}}{\mathop {\sup }\limits_{{{\it{\Pi}}_m} \circ {\Phi_{{{\it{\Gamma}}_m}}} \in {{\cal B}_m}} \left\{ {\kappa \left( {{{\it{\Pi}}_m} \circ {\Phi_{{{\it{\Gamma}}_m}}},{\tilde{\boldsymbol \upsigma}}} \right) - \kappa \left[ {{{\it{\Pi}}_{m-1}} \circ {\Phi_{{{\it{\Gamma}}_{m-1}}}}\left( {{\it{\Pi}}_m} \circ {\Phi_{{{\it{\Gamma}}_m}}} \right),{\tilde{\boldsymbol \upsigma}}} \right]} \right\}}. 
    \end{aligned}
    \label{eq:b.3}
\end{equation}
Since the cardinalities of the subsets $\mathcal{B}_m$  are finite, and the expected supremum of a sub-Gaussian process’ increment over a finite index set can be bounded by the index's finite cardinality, as shown in {\color{black} Bartlett’s finite-class lemma} on Page 2 of \citet{bartlett2013theoretical}, the last term in Eq.~\eqref{eq:b.3} can be bounded by
\begin{equation}
    \begin{aligned}
        &\sum\limits_{m = 1}^M \mathbb{E}_{{\tilde{\boldsymbol \upsigma }}}{\mathop {\sup }\limits_{{{\it{\Pi}}_m} \circ {\Phi_{{{\it{\Gamma}}_m}}} \in {{\cal B}_m}} \left\{ {\kappa \left( {{{\it{\Pi}}_m} \circ {\Phi_{{{\it{\Gamma}}_m}}},{\tilde{\boldsymbol \upsigma}}} \right) - \kappa \left[ {{{\it{\Pi}}_{m-1}} \circ {\Phi_{{{\it{\Gamma}}_{m-1}}}}\left( {{\it{\Pi}}_m} \circ {\Phi_{{{\it{\Gamma}}_m}}} \right),{\tilde{\boldsymbol \upsigma}}} \right]} \right\}}\\
        &\;\;\;\mathop  \le \sum\limits_{m = 1}^M {2\sqrt {NT{\wPiout}} {\BPi}{2^{ - (m - 1)}}\sqrt {2\ln N\left( {{\FPiPhiGamma},{d_\kappa },2\sqrt {NT{\wPiout}} {\BPi}{2^{ - m}}} \right)}} \\
        &\;\;\; = 4\sum\limits_{m = 1}^M {2\sqrt {NT{\wPiout}} {\BPi}{2^{ - (m + 1)}}\sqrt {2\ln N\left( {{\FPiPhiGamma},{d_\kappa },2\sqrt {NT{\wPiout}} {\BPi}{2^{ - m}}} \right)}} \\
        &\;\;\; \le 4\sum\limits_{m = 1}^M {\int_{2\sqrt {NT{\wPiout}} {\BPi}{2^{ - (m + 1)}}}^{2\sqrt {NT{\wPiout}} {\BPi}{2^{ - m}}} {\sqrt {2\ln N\left( {\FPiPhiGamma,{d_\kappa },\varepsilon } \right)} {\rm{d}}\varepsilon } } \\
        &\;\;\; = 4\sqrt 2 \int_{\sqrt {NT{\wPiout}} {B_\Pi }{2^{ - M}}}^{\sqrt {NT{\wPiout}} {B_\Pi }} {\sqrt {\ln N\left( {\FPiPhiGamma,{d_\kappa },\varepsilon } \right)} {\rm{d}}\varepsilon }.
    \end{aligned}
    \label{eq:b.4}
\end{equation}
Substituting Eq.~\eqref{eq:b.4} into Eq.~\eqref{eq:b.3}, it can be obtained
\begin{equation}
    \begin{aligned}
        &\mathbb{E}_{{\tilde{\boldsymbol \upsigma }}}\kappa \left( {\PiPhiGamma,{\tilde{\boldsymbol \upsigma}}} \right) \le \mathbb{E}_{{\tilde{\boldsymbol \upsigma }}}\mathop {\sup }\limits_{\scriptstyle{d_\kappa }\left( {{{\it{\Pi}}_1} \circ {\Phi_{{{\it{\Gamma}}_1}}},{{\it{\Pi}}_2} \circ {\Phi_{{{\it{\Gamma}}_2}}}} \right) \le {2\sqrt{NT\wPiout}{\BPi}2^{ - M}}\hfill\atop\scriptstyle{{{\it{\Pi}}_1} \circ {\Phi_{{{\it{\Gamma}}_1}}},{{\it{\Pi}}_2} \circ {\Phi_{{{\it{\Gamma}}_2}}}} \in {\FPiPhiGamma}\hfill} \left[\kappa \left( {{{\it{\Pi}}_1} \circ {\Phi_{{{\it{\Gamma}}_1}}},{\tilde{\boldsymbol \upsigma}}} \right) - \kappa \left( {{{\it{\Pi}}_2} \circ {\Phi_{{{\it{\Gamma}}_2}}},{\tilde{\boldsymbol \upsigma}}} \right)\right]\\
        &\;\;\;\;\;\;\;\;\;\;\;\;\;\;\;\;\;\;\;\;\;\;\;\;\;\;\;\; + 4\sqrt 2 \int_{\sqrt {NT{\wPiout}} {B_\Pi }{2^{ - M}}}^{\sqrt {NT{\wPiout}} {B_\Pi }} {\sqrt {\ln N\left( {\FPiPhiGamma,{d_\kappa },\varepsilon } \right)} {\rm{d}}\varepsilon}\\
        &{\color{black} \;\;\;\;\;\;\;\;\;\;\;\;\;\;\;\;\;\;\;\;\;\;\;\;\; = E_1(M) + 4\sqrt 2 \int_{\sqrt {NT{\wPiout}} {B_\Pi }{2^{ - M}}}^{\sqrt {NT{\wPiout}} {B_\Pi }} {\sqrt {\ln N\left( {\FPiPhiGamma,{d_\kappa },\varepsilon } \right)} {\rm{d}}\varepsilon}.} 
    \end{aligned}
    \label{eq:b.5}
\end{equation}
{\color{black} The first term $E_1(M)$ on the right-hand side of Eq.~\eqref{eq:b.5} can also be bounded by a tail of a similar entropy integral and it consequently tends to zero as $M\to+\infty$. This is demonstrated as follows.

For each
integer $m\ge M$, $\mathcal{B}_m$ is a minimal $2\sqrt{NT\wPiout}\BPi2^{-m}$-cover of
$\FPiPhiGamma$ with a covering number
$N\left(\FPiPhiGamma,d_\kappa,2\sqrt{NT\wPiout}\BPi2^{-m}\right)$. Given any
${\itPi_i}\circ\Phi_{\itGamma_i}\in\FPiPhiGamma$, $i = 1$ and $2$, with
$d_\kappa\left({\itPi_1}\circ\Phi_{\itGamma_1},
{\itPi_2}\circ\Phi_{\itGamma_2}\right)\le 2\sqrt{NT\wPiout}\BPi2^{-M}$, neural oscillator sequences ${\itPi^{(i)}_M}\circ\Phi_{\itGamma^{(i)}_M},
{\itPi^{(i)}_{M+1}}\circ\Phi_{\itGamma^{(i)}_{M+1}},
{\itPi^{(i)}_{M+2}}\circ\Phi_{\itGamma^{(i)}_{M+2}},\ldots$ are defined by setting ${\itPi^{(i)}_m}\circ\Phi_{\itGamma^{(i)}_m}\in\mathcal{B}_m$
to be the closest element of $\mathcal{B}_m$ to
${\itPi_i}\circ\Phi_{\itGamma_i}$ for $m\ge M$. Then, $\kappa\left({\itPi_i}\circ\Phi_{\itGamma_i},\tilde{\boldsymbol\upsigma}\right)$ can be calculated as
\begin{equation}
    \kappa\left({\itPi_i}\circ\Phi_{\itGamma_i},\tilde{\boldsymbol\upsigma}\right)
    = \kappa\left({\itPi^{(i)}_M}\circ\Phi_{\itGamma^{(i)}_M},\tilde{\boldsymbol\upsigma}\right) + \sum_{m=M}^{\infty}\left[\kappa\left({\itPi^{(i)}_{m+1}}\circ\Phi_{\itGamma^{(i)}_{m+1}},\tilde{\boldsymbol\upsigma}\right)
    -\kappa\left({\itPi^{(i)}_m}\circ\Phi_{\itGamma^{(i)}_m},\tilde{\boldsymbol\upsigma}\right)\right],
    \label{eq:b.6}
\end{equation}
and $\kappa\left({\itPi_1}\circ\Phi_{\itGamma_1},\tilde{\boldsymbol\upsigma}\right) - \kappa\left({\itPi_2}\circ\Phi_{\itGamma_2},\tilde{\boldsymbol\upsigma}\right)$ is
\begin{equation}
    \begin{aligned}
    &\kappa\left({\itPi_1}\circ\Phi_{\itGamma_1},\tilde{\boldsymbol\upsigma}\right) - \kappa\left({\itPi_2}\circ\Phi_{\itGamma_2},\tilde{\boldsymbol\upsigma}\right)\\
    &\;\;\; = \kappa\left({\itPi^{(1)}_M}\circ\Phi_{\itGamma^{(1)}_M},\tilde{\boldsymbol\upsigma}\right) - \kappa\left({\itPi^{(2)}_M}\circ\Phi_{\itGamma^{(2)}_M},\tilde{\boldsymbol\upsigma}\right) + \sum_{m=M}^{\infty}\left[\kappa\left({\itPi^{(1)}_{m+1}}\circ\Phi_{\itGamma^{(1)}_{m+1}},\tilde{\boldsymbol\upsigma}\right) - \kappa\left({\itPi^{(1)}_m}\circ\Phi_{\itGamma^{(1)}_m},\tilde{\boldsymbol\upsigma}\right)\right]\\
    &\;\;\;\;\;\; - \sum_{m=M}^{\infty}\left[\kappa\left({\itPi^{(2)}_{m+1}}\circ\Phi_{\itGamma^{(2)}_{m+1}},\tilde{\boldsymbol\upsigma}\right) - \kappa\left({\itPi^{(2)}_m}\circ\Phi_{\itGamma^{(2)}_m},\tilde{\boldsymbol\upsigma}\right)\right].
    \end{aligned}
    \label{eq:b.7}
\end{equation}
Taking the supremum over $\left\{{\itPi_i}\circ\Phi_{\itGamma_i}\in\FPiPhiGamma,\,i=1,2\,:\,d_\kappa\left({\itPi_1}\circ\Phi_{\itGamma_1},{\itPi_2}\circ\Phi_{\itGamma_2}\right)\le 2\sqrt{NT\wPiout}\BPi 2^{-M}\right\}$
on both sides of Eq.~(\ref{eq:b.7}) and then taking the expectation
$\mathbb{E}_{\tilde{\boldsymbol\upsigma}}$, the first term $E_1(M)$ on the right-hand side of Eq.~\eqref{eq:b.5} can be bounded by
\begin{equation}
    \begin{aligned}
    &E_1(M)\le \mathbb{E}_{\tilde{\boldsymbol\upsigma}}\sup_{{\itPi^{(1)}_M}\circ\Phi_{\itGamma^{(1)}_M},{\itPi^{(2)}_M}\circ\Phi_{\itGamma^{(2)}_M}\in\mathcal{B}_M}\left[\kappa\left({\itPi^{(1)}_M}\circ\Phi_{\itGamma^{(1)}_M},\tilde{\boldsymbol\upsigma}\right) - \kappa\left({\itPi^{(2)}_M}\circ\Phi_{\itGamma^{(2)}_M},\tilde{\boldsymbol\upsigma}\right)\right]\\
    &\;\;\;\;\;\;\;\;\;\;\;\;\; + \sum_{i=1}^{2}\sum_{m=M}^{\infty}\mathbb{E}_{\tilde{\boldsymbol\upsigma}}\sup_{{\itPi^{(i)}_{m+1}}\circ\Phi_{\itGamma^{(i)}_{m+1}}\in\mathcal{B}_{m+1}}\left[\kappa\left({\itPi^{(i)}_{m+1}}\circ\Phi_{\itGamma^{(i)}_{m+1}},\tilde{\boldsymbol\upsigma}\right) - \kappa\left({\itPi^{(i)}_m}\circ\Phi_{\itGamma^{(i)}_m},\tilde{\boldsymbol\upsigma}\right)\right].
    \end{aligned}
    \label{eq:b.8}
\end{equation}

The cardinalities of $\mathcal{B}_M$ and $\mathcal{B}_{m+1}$
are finite. By the triangle inequality,
the pair of ${\itPi^{(1)}_M}\circ\Phi_{\itGamma^{(1)}_M}$ and ${\itPi^{(2)}_M}\circ\Phi_{\itGamma^{(2)}_M}$ satisfies
$d_\kappa\left({\itPi^{(1)}_M}\circ\Phi_{\itGamma^{(1)}_M},
{\itPi^{(2)}_M}\circ\Phi_{\itGamma^{(2)}_M}\right)\le
6\sqrt{NT\wPiout}\BPi 2^{-M}$ and takes at most
$N^2\left(\FPiPhiGamma,d_\kappa,2\sqrt{NT\wPiout}\BPi 2^{-M}\right)$
distinct values. Similarly, the pair
${\itPi^{(i)}_{m+1}}\circ\Phi_{\itGamma^{(i)}_{m+1}}$ and ${\itPi^{(i)}_m}\circ\Phi_{\itGamma^{(i)}_m}$ takes at most
$N^2\left(\FPiPhiGamma,d_\kappa,2\sqrt{NT\wPiout}\BPi 2^{-m-1}\right)$
distinct values with $d_\kappa\left({\itPi^{(i)}_{m+1}}\circ\Phi_{\itGamma^{(i)}_{m+1}},
{\itPi^{(i)}_m}\circ\Phi_{\itGamma^{(i)}_m}\right)\le 6\sqrt{NT\wPiout}\BPi2^{-m-1}$ for $i = 1$ and $2$. Since $\kappa \left(\PiPhiGamma,\tilde{\boldsymbol \upsigma}\right)$
is a zero-mean sub-Gaussian process with respect to $\PiPhiGamma \in \FPiPhiGamma$ (Lemma~4), Bartlett's finite-class lemma
\citep{bartlett2013theoretical} gives
\begin{equation}
    \begin{aligned}
    &\mathbb{E}_{\tilde{\boldsymbol\upsigma}}\sup_{{\itPi^{(1)}_M}\circ\Phi_{\itGamma^{(1)}_M},{\itPi^{(2)}_M}\circ\Phi_{\itGamma^{(2)}_M}\in\mathcal{B}_M}\left[\kappa\left({\itPi^{(1)}_M}\circ\Phi_{\itGamma^{(1)}_M},\tilde{\boldsymbol\upsigma}\right) - \kappa\left({\itPi^{(2)}_M}\circ\Phi_{\itGamma^{(2)}_M},\tilde{\boldsymbol\upsigma}\right)\right]\\
    &\;\;\;\le 12\sqrt{NT\wPiout}\BPi 2^{-M}\sqrt{\ln N\left(\FPiPhiGamma,d_\kappa,2\sqrt{NT\wPiout}\BPi 2^{-M}\right)},
    \end{aligned}
    \label{eq:b.9}
\end{equation}
and
\begin{equation}
    \begin{aligned}
    &\mathbb{E}_{\tilde{\boldsymbol\upsigma}}\sup_{{\itPi^{(i)}_{m+1}}\circ\Phi_{\itGamma^{(i)}_{m+1}}\in\mathcal{B}_{m+1}}\left[\kappa\left({\itPi^{(i)}_{m+1}}\circ\Phi_{\itGamma^{(i)}_{m+1}},\tilde{\boldsymbol\upsigma}\right) - \kappa\left({\itPi^{(i)}_m}\circ\Phi_{\itGamma^{(i)}_m},\tilde{\boldsymbol\upsigma}\right)\right]\\
    &\;\;\;\le 12\sqrt{NT\wPiout}\BPi 2^{-m-1}\sqrt{\ln N\left(\FPiPhiGamma,d_\kappa,2\sqrt{NT\wPiout}\BPi 2^{-m-1}\right)}.
    \end{aligned}
    \label{eq:b.10}
\end{equation}

Substituting Eqs.~\eqref{eq:b.9} and \eqref{eq:b.10} into
Eq.~\eqref{eq:b.8} and converting the dyadic summation into an integral using the same method in Eq.~\eqref{eq:b.4},
$E_1(M)$ in Eq.~\eqref{eq:b.5} can be bounded by
\begin{equation}
    E_1(M)\le 48\int_0^{2\sqrt{NT\wPiout}\BPi 2^{-M}}\sqrt{\ln N\left(\FPiPhiGamma,d_\kappa,\varepsilon\right)}\,{\rm{d}}\varepsilon.
    \label{eq:b.11}
\end{equation}
Under $M$ approaching the positive infinity, $E_1(M)$ in Eq.~\eqref{eq:b.5} converges to zero from Eq.~\eqref{eq:b.11}, yielding
\begin{equation}
    {\mathbb{E}_{{\tilde{\boldsymbol \upsigma }}}}\mathop {\sup }\limits_{\itPi  \circ {\Phi _\itGamma } \in {{\cal F}_{\itPi  \circ {\Phi _\itGamma }}}} \kappa \left( {\itPi  \circ {\Phi _\itGamma },{\tilde{\boldsymbol \upsigma }}} \right) \le 4\sqrt 2 \int_0^{\sqrt {NT{w_{\itPi ,{\rm{out}}}}} {B_\itPi }} {\sqrt {\ln N\left( {{{\cal F}_{\itPi  \circ {\Phi _\itGamma }}},{d_\kappa },\varepsilon } \right)} {\rm{d}}\varepsilon }.
    \label{eq:b.12}
\end{equation}
This completes the proof.
\hfill$\blacksquare$}

\section{Approach for Training the Neural Oscillator}
\label{appC}
By introducing the state space vector $\mathbf{z}(t) = \left[\mathbf{z}_1^\top(t),\mathbf{z}_2^\top(t)\right]^\top = \left[\mathbf{x}^\top(t),\mathbf{x}'^\top(t)\right]^\top$, Eq.~\eqref{eq:2.1} over the time interval $\left[0,T\right]$ can be rewritten in the state function form
\begin{equation}
    \left\{
    \begin{aligned}
        &\mathbf{z}_1'(t) = \mathbf{z}_2(t)\\
        &\mathbf{z}_2'(t) = \itGamma\left[\mathbf{z}_1(t),\mathbf{z}_2(t),\mathbf{u}(t)\right]\\
        &\mathbf{y}(t) = \itPi\left[\mathbf{z}_1(t),\mathbf{u}(0),t\right]
    \end{aligned}.\right.
    \label{eq:c1}
\end{equation}
Utilizing a second-order Runge-Kutta scheme \citep{griffiths2010numerical}, a time-discretization form of Eq.~\eqref{eq:c1} is
\begin{equation}
    \left\{
    \begin{aligned}
        &\mathbf{y}(t_{i+1}) = \itPi\left[\mathbf{z}_1(t_{i+1}),\mathbf{u}(0),t_{i+1}\right]\\
        &\mathbf{z}(t_{i+1}) = 0.5\Delta{t_i}\left(\mathbf{k}_2+\mathbf{k}_1\right)\\
        &\mathbf{k}_2 =
        \begin{Bmatrix}
            \mathbf{z}_2(t_i)+\Delta{t_i}\mathbf{k}_{12}\\
            \itGamma\left[\mathbf{z}_1(t_i)+\Delta{t_i}\mathbf{k}_{11},\mathbf{z}_2(t_i)+\Delta{t_i}\mathbf{k}_{12},\mathbf{u}(t_{i+1})\right]
        \end{Bmatrix}\\
        &\mathbf{k}_1=
        \begin{bmatrix}
            \mathbf{k}_{11}\\
            \mathbf{k}_{12}
        \end{bmatrix}=
        \begin{Bmatrix}
            \mathbf{z}_2(t_i)\\
            \itGamma\left[\mathbf{z}_1(t_i),\mathbf{z}_2(t_i),\mathbf{u}(t_i)\right]
        \end{Bmatrix}
    \end{aligned},\right.
    \label{eq:c2}
\end{equation}
where $\mathbf{z}(0) = \mathbf{0}$, $\mathbf{y}(0)=\itPi\left[\mathbf{z}_1(0),\mathbf{u}(0),t\right]$, $\Delta{t_i} = t_{i+1} - t_i$, $0 = t_0 < t_1 < t_2 < \dots < t_I = T$. Given available output function samples $\hat{\mathbf{y}}_l(t_i) = \left[\hat{y}_{l,1}(t_i),\hat{y}_{l,2}(t_i),\dots,\hat{y}_{l,q}(t_i)\right]^\top$ caused by the input samples $\mathbf{u}_l(t_i)$, $l = 1, 2, \dots, L$, the {\color{black} following} loss function $\hat\ell_{\lambda_L}$ is used to learn the parameters of the MLPs $\itGamma(\cdot)$ and $\itPi(\cdot)$ in Eq.~\eqref{eq:c1},
\begin{equation}
    \hat\ell_{\lambda_L} = \frac{1}{LIq}{\sum\limits_{l = 1}^L {\sum\limits_{i = 0}^{I-1}\sum\limits_{j = 1}^q {{\left|{y_{l,j}}(t_i) - \hat{y}_{l,j}(t_i)\right|}^2}}} + \frac{\lambda_L}{\sqrt{N}}\left[\sum_{i=1}^{2}\left(\left|\mathbf{W}_i\right|_{1,1}+\left|\mathbf{b}_i\right|_{1} \right)+ \sum_{j=1}^{\hPi}\left(\left|\tilde{\mathbf{W}}_j\right|_{1,1}+\left|\tilde{\mathbf{b}}_j\right|_{1} \right)\right],
    \label{eq:c3}
\end{equation}
where $\mathbf{y}_l(t_i) = \left[y_{l,1}(t_i),y_{l,2}(t_i),\dots,y_{l,q}(t_i)\right]^\top$ represents the output computed from the neural oscillator driven by $\mathbf{u}_l(t_i)$ using Eq.~\eqref{eq:c2}. In this numerical study, the Adam method \citep{kingma2014adam} implemented using the PyTorch \citep{paszke2019pytorch} is utilized to train the neural oscillator using the Runge-Kutta scheme in Eq.~\eqref{eq:c2} and the loss function $\hat\ell_{\lambda_L}$ in Eq.~\eqref{eq:c3}.

\bibliographystyle{elsarticle-harv} 
\bibliography{references}






\end{document}